\documentclass[journal]{IEEEtran}
\usepackage{amsmath}
\DeclareMathOperator{\Tr}{Tr}

\usepackage{silence}
\usepackage{graphicx}
\usepackage[utf8]{inputenc}
\usepackage{csquotes}
\usepackage{xcolor}

\usepackage{colortbl}

\usepackage{mathtools}
\usepackage{bbold}

\usepackage{pifont}
\newcommand{\xmark}{\ding{55}}%

\usepackage{subcaption}
\usepackage{booktabs}   

\usepackage{comment}
\usepackage{CJKutf8}
\include{macro.tex}

\usepackage{hyperref}
\hypersetup{colorlinks=true,                
    breaklinks=true,                
    urlcolor= black,                
    linkcolor= blue,                
    bookmarksopen=false,
    filecolor=black,
    citecolor=blue,
    linkbordercolor=blue}

\AtBeginDocument{\hypersetup{pdfborder={0 0 1}}}

\begin{document}
\title{A Review on Object Pose Recovery: from $3$D Bounding Box Detectors to Full $6$D Pose Estimators}

\author{Caner Sahin, Guillermo Garcia-Hernando, Juil Sock, and Tae-Kyun Kim \\
Imperial College London, UK
\thanks{Electrical and Electronic Engineering Department, Imperial Computer Vision and Learning Lab (ICVL), Imperial College London, SW72AZ, UK, canersahin130@gmail.com, \{g.garcia-hernando, ju-il.sock08, tk.kim\}@imperial.ac.uk}}


\maketitle
\begin{abstract}
Object pose recovery has gained increasing attention in the computer vision field as it has become an important problem in rapidly evolving technological areas related to autonomous driving, robotics, and augmented reality. Existing review-related studies have addressed the problem at visual level in $2$D, going through the methods which produce $2$D bounding boxes of objects of interest in RGB images. The $2$D search space is enlarged either using the geometry information available in the $3$D space along with RGB (Mono/Stereo) images, or utilizing depth data from LIDAR sensors and/or RGB-D cameras. $3$D bounding box detectors, producing category-level amodal $3$D bounding boxes, are evaluated on gravity aligned images, while full $6$D object pose estimators are mostly tested at instance-level on the images where the alignment constraint is removed. Recently, $6$D object pose estimation is tackled at the level of categories. In this paper, we present the first comprehensive and most recent review of the methods on object pose recovery, from $3$D bounding box detectors to full $6$D pose estimators. The methods mathematically model the problem as a classification, regression, classification \& regression, template matching, and point-pair feature matching task. Based on this, a mathematical-model-based categorization of the methods is established. Datasets used for evaluating the methods are investigated with respect to the challenges, and evaluation metrics are studied. Quantitative results of experiments in the literature are analyzed to show which category of methods best performs across what types of challenges. The analyses are further extended comparing two methods, which are our own implementations, so that the outcomes from the public results are further solidified. Current position of the field is summarized regarding object pose recovery, and possible research directions are identified.
\end{abstract}
\IEEEpeerreviewmaketitle

\section{Introduction}
\IEEEPARstart{O}bject pose recovery is an important problem in the computer vision field, which, to the full extent, determines the $3$D position and $3$D rotation of an object in camera-centered coordinates. It has extensively been studied in the past decade as it is of great importance to many rapidly evolving technological areas such as autonomous driving, robotics, and augmented reality.\\
\indent Autonomous driving, as being a focus of attention of both industry and research community in recent years, fundamentally requires accurate object detection and pose estimation in order for a vehicle to avoid collisions with pedestrians, cyclists, and cars. To this end, autonomous vehicles are equipped with active LIDAR sensors \cite{e150,e156}, passive Mono/Stereo (Mo/St) RGB/D/RGB-D cameras \cite{88,91}, and their fused systems \cite{86,124}.\\
\indent Robotics has various sub-fields that vastly benefit from accurate object detection and pose estimation. One prominent field is robotic manipulation, in which industrial parts are grasped and are placed to a desired location \cite{65, 56, 57, 62, 101, b212, b210}. Amazon Picking Challenge (APC) \cite{1} is an important example demonstrating how object detection and $6$D pose estimation, when successfully performed, improves the autonomy of the manipulation, regarding the automated handling of parts by robots \cite{55, 58, 60}. Household robotics is another field where the ability of recognizing objects and accurately estimating their poses is a key element \cite{75, 113, 114, 115}. This capability is needed for such robots, since they should be able to navigate in unconstrained human environments calculating grasping and avoidance strategies. Autonomous flight performance of Unmanned Aerial Vehicles (UAVs) is improved integrating object pose recovery into their control systems. Successful results are particularly observed in environments where GPS information is denied or is inaccurate \cite{54, 64, 53}.\\
\indent Simultaneous Localization and Mapping (SLAM) is the process of building the map of an unknown environment and determining the location of a robot concurrently using this map. Recently introduced SLAM algorithms have object-oriented design \cite{66, 67, 68, 69, 70, 71, 72, 73, 74}. Accurate object pose parameters provide camera-object constraints and lead to better performance on localization and mapping.\\
\indent Augmented Reality (AR) technology transforms the real world into an interactive and digitally manipulated space, superimposing computer generated content onto the real world. Augmentation of reality requires the followings: (i) a region to augment must be detected in the image (object detection) and (ii) the rotation for the augmented content must be estimated (pose estimation) \cite{45}. Virtual Reality (VR) transports users into a number of real-world and imagined environments \cite{46, 49, 50, 52}, and Mixed Reality (MR) \cite{47, 48, 51} merges both AR and VR technologies. Despite slight differences, their success lies on accurate detection and pose estimation of objects.\\
\indent In the progress of our research on object pose recovery, we have come across several review-related publications \cite{8, 9, 10, 11, 12, 13, 125, b209} handling object pose recovery at visual level in $2$D. These publications discuss the effects of challenges, such as occlusion, clutter, texture, \textit{etc.}, on the performances of the methods, which are mainly evaluated on large-scale datasets, \textit{e.g.}, ImageNet \cite{6}, PASCAL \cite{7}. In \cite{8}, the effect of different context sources, such as geographic context, object spatial support, on $2$D object detection is examined. Hoiem et al. \cite{9} evaluate the performances of several baselines on the PASCAL dataset particularly analyzing the reasons why false positives are hypothesized. Torralba et al. \cite{11} compare several datasets regarding the involved samples, cross-dataset generalization, and relative data bias. Since there are less number of object categories in PASCAL, Russakovsky et al. \cite{10} use ImageNet in order to do meta-analysis and to examine the influences of color, texture, \textit{etc.}, on the performances of object detectors. The retrospective evaluation \cite{12} and benchmark \cite{13} studies perform the most comprehensive analyses on $2$D object localization and category detection, by examining the PASCAL Visual Object Classes (VOC) Challenge, and the ImageNet Large Scale Visual Recognition Challenge, respectively. Recently proposed study \cite{125} systematically reviews deep learning based object detection frameworks including several specific tasks, such as salient object detection, face detection, and pedestrian detection. These studies introduce important implications for generalized object detection, however, the reviewed methods and discussions are restricted to visual level in $2$D.\\
\indent $2$D-driven $3$D bounding box (BB) detection methods enlarge the $2$D search space using the available appearance and geometry information in the $3$D space along with RGB images \cite{94, e162, e164, e165a, e152}. The methods presented in \cite{88, e165b} directly detect $3$D BBs of the objects in a monocular RGB image exploiting contextual models as well as semantics. Apart from \cite{88, e165b}, the methods which directly detect $3$D BBs use stereo images \cite{102}, RGB-D cameras \cite{90, 104, 105, 106, e163}, and LIDAR sensors \cite{e163} as input. There are also several methods fusing multiple sensor inputs, and generating $3$D BB hypotheses \cite{123}. Without depending on the input (RGB (Mo/St), RGB-D, or LIDAR), $3$D BB detection methods produce \textbf{oriented} $3$D BBs, which are parameterized with center $ \mathbf{x} = (x, y, z)$, size $ \mathbf{d} = (d_w, d_h, d_l)$, and orientation $(\theta_y)$ around the gravity direction. Note that, any $3$D BB detector can be extended to the $6$D space, however, available methods in the literature are mainly evaluated on the KITTI \cite{126}, NYU-Depth v2 \cite{127}, and SUN RGB-D \cite{128} datasets, where the objects of interest are aligned with the gravity direction. As required by the datasets, these methods work at the level of categories \textit{e.g.}, cars, bicycles, pedestrians, chairs, beds, producing \textbf{amodal} $3$D BBs, \textit{i.e.}, the minimum $3$D BB enclosing the object of interest.\\
\indent The search space of the methods engineered for object pose recovery is further enlarged to $6$D \cite{b175,b181,b185,b193}, \textit{i.e.}, $3$D translation $\mathbf{x} = (x, y, z)$ and $3$D rotation $\mathbf{\Theta}=(\theta_r, \theta_p, \theta_y)$. Some of the methods estimate $6$D poses of objects of interest in RGB-D images. The holistic template matching approach, Linemod \cite{2}, estimates cluttered object's $6$D pose using color gradients and surface normals. It is improved by discriminative learning in \cite{24}. FDCM \cite{23} is used in robotics applications. Drost et al. \cite{17} create a global model description based on oriented point pair features and match that model locally using a fast voting scheme. It is further improved in \cite{43} making the method more robust across clutter and sensor noise. Latent-class Hough forests (LCHF) \cite{4, 26}, employing one-class learning, utilize surface normal and color gradient features in a part-based approach in order to provide robustness across occlusion. Occlusion-aware features \cite{27} are further formulated. The studies in \cite{28,29} cope with texture-less objects. More recently, feature representations are learnt in an unsupervised fashion using deep convolutional (conv) networks (net) \cite{35, 36}. While these methods fuse data coming from RGB and depth channels, a local belief propagation based approach \cite{76} and an iterative refinement architecture \cite{31, 32} are proposed in depth modality \cite{77}. $6$D object pose estimation is recently achieved from RGB only \cite{b174, b180, b182, b183, b184, 30, 37, 38, 40}, and the current paradigm is to adopt CNNs \cite{b167,b168,b179}. BB8 \cite{40} and Tekin et al. \cite{38} perform corner-point regression followed by PnP. Typically employed is a computationally expensive post processing step such as iterative closest point (ICP) or a verification net \cite{37}. As mainly being evaluated on the LINEMOD \cite{2}, OCCLUSION \cite{28}, LCHF \cite{4}, and T-LESS \cite{42} datasets, full $6$D object pose estimation methods typically work at the level of \textbf{instances}. However, recently proposed methods in \cite{129, 130} address the $6$D object pose estimation problem at \textbf{category level}, handling the challenges, \textit{e.g.}, distribution shifts, intra-class variations \cite{b216}.\\
\indent In this study, we present a comprehensive review on object pose recovery, reviewing the methods from $3$D BB detectors to full $6$D pose estimators. The reviewed methods mathematically model object pose recovery as a classification, regression, classification \& regression, template matching, and point-pair feature matching task. Based on this, a mathematical-model-based categorization of the methods is established, and $5$ different categories are formed. Advances and drawbacks are further studied in order to eliminate ambiguities in between the categories. Each individual dataset in the literature involves some kinds of challenges, across which one can measure the performance by testing a method on it. The challenges of the problem, such as viewpoint variability, occlusion, clutter, intra-class variation, and distribution shift are identified by investigating the datasets created for object pose recovery. The protocols used to evaluate the methods' performance are additionally examined. Once we introduce the categorization of the reviewed methods and identify the challenges of the problem, we next reveal the performance of the methods across the challenges. To this end, publicly available recall values, which are computed under uniform scoring criteria of the Average Distance (AD) metric \cite{2}, are analyzed. The analyses are further extended comparing two more methods \cite{2, 4}, which are our own implementations, using the Visible Surface Discrepancy (VSD) protocol \cite{3} in addition to AD. This extension mainly aims to leverage the outcomes obtained from the public results and to complete the discussions on the problem linking all its aspects. The current position of the field is lastly summarized regarding object pose recovery, and possible research directions are identified.\\
\indent Benchmark for $6$D object pose estimation (BOP) \cite{5}, contributes a unified format of datasets, an on-line evaluation system based on the VSD metric, and a comprehensive evaluation of $15$ different methods \cite{b215}. The benchmark addresses object pose recovery only considering $6$D object pose estimation methods, which work at instance-level. The analyses on methods' performance are relatively limited to the technical background of the methods evaluated. In this study, we contribute the most comprehensive review of the methods on object pose recovery, from $3$D BB detectors to full $6$D pose estimators, which work both at instance- and category-level. Our analyses on quantitative results interpret the reasons behind methods' strength and weakness with respect to the challenges.\\
\noindent \textbf{Contributions.} To the best of our knowledge, this is the first comprehensive and most recent study reviewing the methods on object pose recovery. Our contributions are as follows:
\begin{itemize}
\item A mathematical-model-based categorization of object pose estimation methods is established. The methods concerned range from $3$D BB detectors to full $6$D pose estimators.
\item Datasets for object pose recovery are investigated in order to identify its challenges.
\item Publicly available results are analyzed to measure the performance of object pose estimation methods across the challenges. The analyses are further solidified by comparing in-house implemented methods. 
\item Current position of the field on object pose recovery is summarized to unveil existing issues. Furthermore, open problems are discussed to identify potential future research directions.
\end{itemize}

\noindent \textbf{Organization of the Review.} The paper is organized as follows: Section \ref{sect_2} formulates the problem of object pose recovery at instance- and category-level. Section \ref{sect_3} presents the categorization of the reviewed methods. In the categorization, there are five categories of the methods: classification, regression, classification \& regression, template matching, and point-pair feature matching. Section \ref{ch4_datasets} depicts the investigations on the datasets and examines the evaluation protocols, and Section \ref{ch4_Exp_Res} shows the analyses on public results and the results of in-house implemented methods. Section \ref{Summary} summarizes the current position of the field and concludes the study.\\
\noindent \textbf{Denotation.} Throughout the paper, we denote scalar variables by lowercase letters (\textit{e.g.}, $x$) and vector variables bold lowercase letters (\textit{e.g.}, $\mathbf{x}$). Capital letters (\textit{e.g.}, $P$) are adopted for specific functions or variables. We use bold capital letters (\textit{e.g.}, $\mathbf{I}$) to denote a matrix or a set of vectors. Table \ref{table_denot} lists the symbols utilized throughout this review. Except the symbols in the table, there may be some symbols for a specific reference. As these symbols are not commonly employed, they are not listed in the table but will be rather defined in the context.
\begin{table}[t]
\fontsize{7}{8.2}\selectfont
\caption{Denotations employed in this review}
\centering
\setlength\tabcolsep{3pt}
\renewcommand{\arraystretch}{1.2}
\begin{tabular}{ l l | l l}
    \toprule
    \textbf{Symbol} &\textbf{Description} &\textbf{Symbol} &\textbf{Description} \\ 
    \midrule
    $\mathbf{I}$            &input image           &$\mathbf{x}=(x,y,z)$                             &$3$D translation \\ 
    $P$                     &probability           &$\mathbf{\Theta}=(\theta_r,\theta_p,\theta_y)$   &$3$D rotation \\ 
    $O$                     &object                &$\mathbf{d} = (d_w,d_h,d_l)$                     &$3$D BB dimension \\ 
    $S$                     &seen instance         &$\mathbf{p}_{3D}=(p_x,p_y,p_z)$                  &$3$D BB corner position \\ 
    $\mathcal{S}$           &set (seen inst)       &$\mathbf{p}_{3D}^{proj}=(p_x,p_y)$               &projection of $\mathbf{p}_{3D}$ \\ 
    $U$                     &unseen instance       &$\mathbf{p}_{2D}=(p_x,p_y)$                      &$2$D BB corner position \\ 
    $\mathcal{U}$           &set (unseen inst)     &$\theta_r, \theta_p, \theta_y$                   &roll, pitch, yaw \\ 
    $C$                     &category              &$d_w,d_h,d_l$                                    &width, height, length \\
    $M$                     &$3$D model            &$i,j,k,n,c$                                      &general index \\
    $IG$                    &information gain      &$\mathbf{p}_{2D_c}$                              &$2$D BB center \\
    $IoU_{3D}$              &$3$D IoU loss         &$\theta_{cvp}$                                   &camera viewpoint \\
    $L_{hinge}$             &hinge loss            &$\theta_{ip}$                                    &in-plane rotation \\
    $L_{pair_d}$            &dynamic pair loss     &$L_{log}$                                        &log-loss\\
    $L_{pose}$              &pose loss             &$L_{1_s}$                                        &smooth $L1$ loss\\
    $L_{pair}$              &pair loss             &$L_{ce}$                                         &cross entropy\\
    $L_{tri_d}$             &dynamic triplet loss  &$L_{il}$                                         &insensitive loss  \\
    $L_{0/1}$               &$0/1$ loss            &$L_2$                                            &$L2$ loss \\
    $L_{tri}$               &triplet loss          &$L_1$                                            &$L1$ loss\\
    \bottomrule
\end{tabular}
\label{table_denot}
\end{table}
\section{Object Pose Recovery}
\label{sect_2}
In this study, we present a review of the methods on object pose recovery, from $3$D BB detectors to full $6$D pose estimators. The methods deal with \textit{seen} objects at instance-level and \textit{unseen} objects at the level of categories. In this section, we firstly formulate object pose recovery for instance-level. Given an RGB/D/RGB-D image $\mathbf{I}$ where a seen instance $S$ of the interested object $O$ exists, object pose recovery is cast as a joint probability estimation problem and is formulated as given below:
\begin{equation}
\small
{(\mathbf{x}, \mathbf{\Theta})}^* = \arg \max_{\mathbf{x}, \mathbf{\Theta}}  P(\mathbf{x}, \mathbf{\Theta} \vert \mathbf{I}, S, O)
\label{eq1}
\end{equation}
where $\mathbf{x} = (x, y, z)$ is the $3$D translation and $\mathbf{\Theta} = (\theta_r, \theta_p, \theta_y)$ is the $3$D rotation of the instance $S$. $(\theta_r, \theta_p, \theta_y)$ depicts the Euler angles, roll, pitch, and yaw, respectively. According to Eq. \ref{eq1}, instance-level methods of object pose recovery target to maximize the joint posterior density of the $3$D translation $\mathbf{x}$ and $3$D rotation $\mathbf{\Theta}$. When the image $\mathbf{I}$ involves multiple instances $\mathcal{S} = \{ S_i \vert i = 1, ... ,n \}$ of the object of interest, the problem formulation becomes:
\begin{equation}
\small
{(\mathbf{x}_i, \mathbf{\Theta}_i)}^* = \arg \max_{\mathbf{x}_i, \mathbf{\Theta}_i}  P(\mathbf{x}_i, \mathbf{\Theta}_i \vert \mathbf{I}, \mathcal{S}, O), \quad i = 1, ..., n.
\label{ch1_eq1}
\end{equation}
Given an unseen instance $U$ of a category of interest $C$, the object pose recovery problem is formulated at the level of categories by transforming Eq. \ref{eq1} into the following form:
\begin{equation}
\small
{(\mathbf{x}, \mathbf{\Theta})}^* = \arg \max_{\mathbf{x}, \mathbf{\Theta}}  P(\mathbf{x}, \mathbf{\Theta} \vert \mathbf{I}, U, C).
\label{eq3c}
\end{equation}
When the image $\mathbf{I}$ involves multiple unseen instances $\mathcal{U} = \{ U_i \vert i = 1, ... ,m \}$ of the category of interest $C$, Eq. \ref{ch1_eq1} takes the following form:
\begin{equation}
\small
{(\mathbf{x}_i, \mathbf{\Theta}_i)}^* = \arg \max_{\mathbf{x}_i, \mathbf{\Theta}_i}  P(\mathbf{x}_i, \mathbf{\Theta}_i \vert \mathbf{I}, \mathcal{U}, C), \quad i = 1, ..., m.
\label{ch1_eq2}
\end{equation}
Based on this formulation, any $3$D BB detection method evaluated on any dataset of categories, \textit{e.g.}, KITTI \cite{126}, NYU-Depth v2 \cite{127}, SUN RGB-D \cite{128}, searches for the accurate $3$D translation $\mathbf{x}$ and the rotation angle around the gravity direction $\theta_y$ of $3$D BBs, which are also the ground truth annotations. Unlike $3$D BB detectors, full $6$D object pose estimators, as being evaluated on the LINEMOD \cite{2}, OCCLUSION \cite{28}, LCHF \cite{4}, and T-LESS \cite{42} datasets, search for the accurate full $6$D poses, $3$D translation $\mathbf{x} = (x, y, z)$ and $3$D rotation $\mathbf{\Theta} = (\theta_r, \theta_p, \theta_y)$, of the objects of interest.
\section{Methods on Object Pose Recovery}
\label{sect_3}
Any method for the problem of object pose recovery estimates the object pose parameters, $3$D translation $\mathbf{x} = (x,y,z)$ \textbf{and/or} $3$D rotation $\mathbf{\Theta} = (\theta_r,\theta_p,\theta_y)$. We categorize the methods reviewed in this paper based on their problem modeling approaches: The methods which estimate the pose parameters formulating the problem as a classification task are involved within the \enquote{\textit{classification}}, while the ones regressing the parameters are included in the \enquote{\textit{regression}}. \enquote{\textit{Classification \& regression}} combines both classification and regression tasks to estimate the objects' $3$D translation and/or $3$D rotation. \enquote{\textit{Template matching}} methods estimate the objects' pose parameters matching the annotated and represented templates in the feature space. \enquote{\textit{Point-pair feature matching}} methods use the relationships between any two points, such as distance and angle between normals to represent point-pairs. Together with a hash table and an efficient voting scheme, the pose parameters of the target objects are predicted. These $5$ different modeling approaches, classification, regression, classification \& regression, template matching, point-pair feature matching, form a \textbf{discriminative} categorization allowing us to review the state-of-the-art methods on the problem of object pose recovery, from $3$D to $6$D.\\
\indent When we review the methods, we encounter several $3$D detection methods, which are $2$D-driven. $2$D-driven $3$D methods utilize any off-the-shelf $2$D detectors \textit{e.g.}, R-CNN \cite{131}, Fast R-CNN \cite{132}, Faster R-CNN \cite{133}, R-FCN \cite{134}, FPN \cite{135}, YOLO \cite{136}, SSD \cite{137}, GOP \cite{138}, or MS-CNN \cite{139}, to firstly detect the $2$D BBs of the objects of interest \cite{e166}, which are then lifted to $3$D space, and hence making their performance dependent on the $2$D detectors. Besides, several $3$D detectors and full $6$D pose estimators, which directly generate objects' $3$D translation and/or $3$D rotation parameters, are built on top of those. Decision forests \cite{x17} are important for the problem of object pose recovery. Before delving into our categorization of the methods, we first briefly mention several $2$D detectors and decision forests.\\
\indent \textbf{R-CNN} \cite{131}, from an input RGB image, generates a bunch of arbitrary size of region proposals employing a selective search scheme \cite{140}, which relies on bottom-up grouping and saliency cues. Region proposals are warped into a fixed resolution and are fed into the CNN architecture in \cite{141} to be represented with CNN feature vectors. For each feature vector, category-specific SVMs generate a score whose region is further adjusted with BB regression, and non-maximum suppression (NMS) filtering. The $3$ stages of process allow R-CNN to achieve approximately $30 \%$ improvement over previous best $2$D methods, \textit{e.g.}, DPM HSC \cite{142}.\\
\indent \textbf{Fast R-CNN} \cite{132} takes an RGB image and a set of region proposals as input. The whole image is processed to produce a conv feature map, from which a fixed-length feature vector is generated for each region proposal. Each feature vector is fed into a set of Fully Connected (FC) layers, which eventually branching into softmax classification and BB regression outlets. Each training Region of Interest (RoI) is labeled with a ground-truth class and a ground-truth regression target. Fast R-CNN introduces a multi-task loss on each RoI to jointly train for classification and BB regression.\\
\indent \textbf{Faster R-CNN} \cite{133} presents Region Proposal Networks (RPN), which takes an RGB image as input and predicts rectangular object proposals along with their \enquote{objectness} score. As being modeled with a fully conv net, RPN shares computation with a Fast R-CNN. Sliding over the conv feature map, it operates on the last conv layer, with the preceding layers shared with the object detection net. It is fully connected to a $n \times n$ spatial window of the conv feature map. In each sliding window, a low dimensional feature vector is generated for each of the $k$ anchors, which is then fed into both box classification and box regression layers.\\
\begin{figure}[!t]
\centering
\includegraphics[width=3.4in]{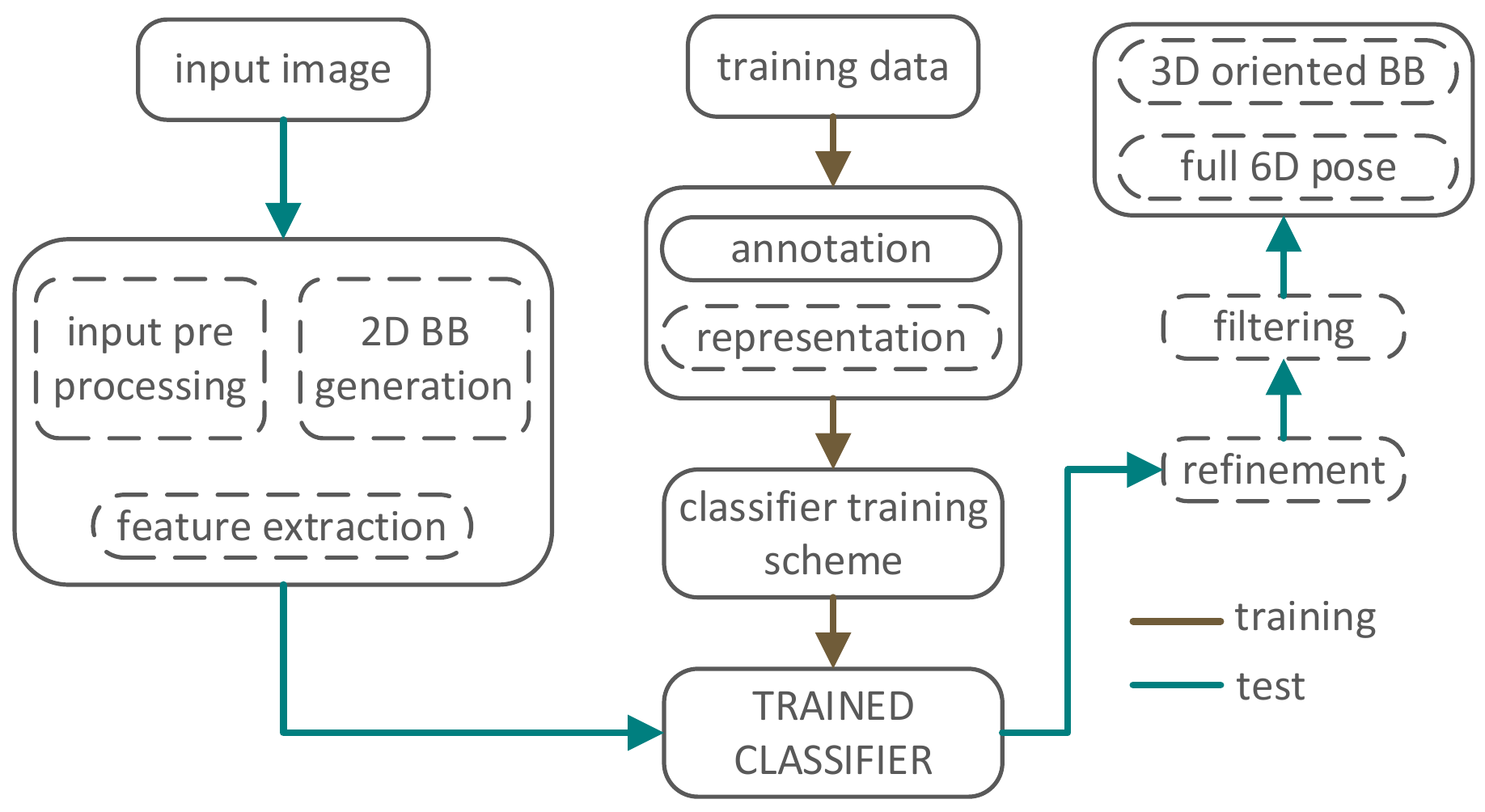}
\caption{Overall schematic representation of classification-based methods: blocks drawn with continuous lines are employed by all methods while dashed-line blocks are operated by clusters of specific methods (see the text and Table \ref{table_cla} for details).}
\label{fig_cla}
\vspace{-1em}
\end{figure}
\indent \textbf{YOLO} \cite{136}, formulating the detection problem as a regression task, predicts BBs and class probabilities directly from full images. The whole process is conducted in a single net architecture, which can be trained in an end-to-end fashion. The architecture divides the input image into an $n \times n$ grid, each cell of which is responsible for predicting the object centered in that grid cell. Each cell predicts $k$ BBs along with their confidence scores. The confidence is defined as the multiplication of $P(O)$ and $IOU_\mathrm{pred}^\mathrm{truth}$, resulting $P(O)*IOU_\mathrm{pred}^\mathrm{truth}$, which indicates how likely there object exists ($P(O) \geq 0$). Each grid cell also calculates $P(C|O)$, conditional class probabilities conditioned on the grid cell containing an object. Class specific confidence scores are obtained during testing, multiplying the conditional class probabilities and the individual box confidence predictions.\\
\indent \textbf{SSD} \cite{137} architecture, taking an RGB image as input, produces $2$D BBs and confidences of the objects of a given category, which are then filtered by the NMS filtering to produce the final detections. The architecture discretizes the output space of the BBs, utilizing a set of default anchor boxes which have different aspect ratios and scales.\\
\begin{table*}[t]
\tiny
\fontsize{10}{12.2}\selectfont
\caption{Classification-based methods ({\small CNN: \textbf{C}onvolutional \textbf{N}eural \textbf{N}etwork}, {\small CRF: \textbf{C}onditional \textbf{R}andom \textbf{F}ield}, {\small ICP: \textbf{I}terative \textbf{C}losest \textbf{P}oint}, {\small NMS: \textbf{N}on-\textbf{M}aximum \textbf{S}uppression}, {\small R: \textbf{R}eal data}, {\small RF: \textbf{R}andom \textbf{F}orest}, {\small S: \textbf{S}ynthetic data}, {\small s-SVM: \textbf{s}tructured-\textbf{S}upport \textbf{V}ector \textbf{M}achine}, for symbols see Table \ref{table_denot})}
\centering
\setlength\tabcolsep{14pt}
\renewcommand{\arraystretch}{1.2}
\begin{minipage}{0.999\textwidth}
\centering
\resizebox{0.999\columnwidth}{!}
{
  \begin{tabular}{l c c c c c c c c c}
    \arrayrulecolor{black}\toprule
    {\small\textbf{method}} &{\small\textbf{input}} &{\small\textbf{input}}          &{\small\textbf{training}}      &{\small\textbf{classification}}    &{\small\textbf{classifier}}                              &{\small\textbf{trained}}     &{\small\textbf{refinement}} &{\small\textbf{filtering}} &{\small\textbf{level}}\\
                    &               &{\small\textbf{pre-processing}}  &{\small\textbf{data}}          &{\small\textbf{parameters}}        &{\small\textbf{training}}                                &{\small\textbf{classifier}}  &{\small\textbf{step}}       &                   &\\
    \arrayrulecolor{black}\bottomrule
    & & & & {\small \textbf{$2$D-DRIVEN $3$D} }\\
    \arrayrulecolor{black}\toprule
    GS$3$D \cite{120}                &{\small RGB}      &\xmark                      &{\small R}         &$\theta_y$                           &$L_{ce}$               &{\small CNN}   &{\small CNN}                 &\xmark       &category\\
    \textbf{refinement step}         &{\small RGB}      &\xmark                      &{\small R}         &$\mathbf{x}, \mathbf{d}, \theta_y$   &$L_{ce}$               &{\small CNN}   &\xmark                       &\xmark       &\\ 
    \arrayrulecolor{gray}\midrule
    Papon et al. \cite{92}           &{\small RGB-D}    &{\small intensity \& normal}  &{\small R \& S}    &$\theta_y, z$                        &$L_{ce}$               &{\small CNN}   &\xmark                       &{\small NMS} &category\\
    Gupta et al. \cite{85}           &{\small RGB-D}    &{\small normal}             &{\small S}         &$\theta_y$                           &$L_{ce}$               &{\small CNN}   &{\small ICP}                 &\xmark       &category\\
    \arrayrulecolor{black}\bottomrule
    & & & & {\small \textbf{$3$D} }\\
    \arrayrulecolor{black}\toprule 
    Sliding Shapes \cite{93}         &{\small Depth}    &{\small $3$D grid}          &{\small R \& S}    &$\mathbf{x}, \mathbf{d}, \theta_y$   &$L_{hinge}$             &{\small SVM}   &\xmark                      &{\small NMS} &category\\
    \arrayrulecolor{gray}\midrule
    Ren et al. \cite{e160}           &{\small RGB-D}    &{\small $3$D grid}          &{\small R}         &$\mathbf{x}, \mathbf{d}, \theta_y$   &$IoU_{3D}$              &{\small s-SVM} &s-SVM                       &{\small NMS} &category\\
    \textbf{refinement step}         &{\small RGB-D}    &{\small $3$D grid}          &{\small R}         &$\mathbf{x}, \mathbf{d}, \theta_y$   &$IoU_{3D}$              &{\small s-SVM} &\xmark                      &\xmark       &\\
    \arrayrulecolor{gray}\midrule
    Wang et al. \cite{111}           &{\small LIDAR}    &{\small $3$D grid}          &{\small R}         &$\mathbf{x}$                         &$L_{hinge}$             &{\small SVM}   &\xmark                      &{\small NMS} &category\\
    Vote$3$Deep \cite{80}            &{\small LIDAR}    &{\small $3$D grid}          &{\small R}         &$\mathbf{x}, \theta_y$               &$L_{hinge}$             &{\small CNN}   &\xmark                      &{\small NMS} &category\\
    \arrayrulecolor{black}\bottomrule
    & & & & {\small \textbf{$6$D} }\\
    \arrayrulecolor{black}\toprule
    Bonde et al. \cite{27}           &{\small Depth}    &{\small $3$D grid}          &{\small S}         &$\mathbf{\Theta}$                    &$IG$                    &{\small RF}    &\xmark                      &\xmark       &instance\\
    Brachmann et al. \cite{28}       &{\small RGB-D}    &\xmark                      &{\small R \& S}    &$\mathbf{x}$                         &$IG$                    &{\small RF}    &{\small ICP}        &\xmark       &instance\\
    \arrayrulecolor{gray}\midrule
    Krull et al. \cite{29}           &{\small RGB-D}    &\xmark                      &{\small R \& S}    &$\mathbf{x}$                         &$IG$                    &{\small RF}    &{\small CNN}                &\xmark       &instance\\
    \textbf{refinement step}         &{\small Depth}        &\xmark                      &{\small R \& S}    &$\mathbf{x}, \mathbf{\Theta}$        &log-like                &{\small CNN}   &\xmark                      &\xmark       &\\
    \arrayrulecolor{gray}\midrule
    Michel et al. \cite{33}          &{\small RGB-D}    &\xmark                      &{\small R \& S}    &$\mathbf{x}$                         &$IG$                    &{\small RF}    &{\small CRF \& ICP} &\xmark       &instance\\
    \textbf{refinement step}         &{\small RGB-D}    &\xmark                      &{\small R \& S}    &$\mathbf{x}, \mathbf{\Theta}$        &\xmark                  &{\small CRF}   &{\small ICP}        &\xmark       &\\
    \arrayrulecolor{black}\bottomrule
  \end{tabular}
}
\end{minipage}%
\label{table_cla}
\end{table*}
\indent \textbf{Decision Forests} \cite{x17, x24} are ensemble of randomized decision trees. Each leaf node of each decision tree of a trained decision forest stores a prediction function. A random decision tree is trained independently from the other trees, and it can be regarded as a weak classifier. In the test, any input ending up at a leaf node, is associated with the stored prediction function. The outcome coming from each tree is averaged, since many weak classifiers, if combined, produce more accurate results \cite{x18, x19, x20, x21, x22}. In classification forests, information gain is often used as the quality function $Q_{cla}$ and in regression tasks, the training objective $Q_{reg}$ is to minimize the variance in translation offset vectors and rotation parameters. For pose regression problems, Hough voting process \cite{x23} is usually employed. 
\subsection{Classification} 
Overall schematic representation of the classification-based methods is shown in Fig. \ref{fig_cla}. In the figure, the blocks drawn with continuous lines are employed by all methods, and depending on the architecture design, dashed-line blocks are additionally operated by the clusters of specific methods.\\
\noindent \textbf{Training Phase.} During an off-line stage, classifiers are trained based on synthetic or real data. Synthetic data are generated using the $3$D model $M$ of an interested object $O$, and a set of RGB/D/RGB-D images are rendered from different camera viewpoints. The $3$D model $M$ can either be a CAD or a reconstructed model, and the following factors are considered when deciding the size of the data:
\begin{itemize}
\item Reasonable viewpoint coverage. In order to capture reasonable viewpoint coverage of the target object, synthetic images are rendered by placing a virtual camera at each vertex of a subdivided icosahedron of a fixed radius. The hemisphere or full sphere of icosahedron can be used regarding the scenario.
\item Object distance. Synthetic images are rendered at different scales depending on the range in which the target object is located.
\end{itemize}
Computer graphic systems provide precise data annotation, and hence synthetic data generated by these systems are used by the classification-based methods \cite{92,85,93,27,28,29,33}. It is hard to get accurate object pose annotations for real images, however, there are classification-based methods using real training data \cite{120,92,93,e160,80,111,28,29,33}. Training data are annotated with pose parameters $\textit{i.e.}$, $3$D translation $\mathbf{x} = (x,y,z)$, $3$D rotation $\mathbf{\Theta} = (\theta_r, \theta_p, \theta_y)$, or both. Once the training data are generated, the classifiers are trained using related loss functions.\\
\noindent \textbf{Testing Phase.} A real test image, during an on-line stage, is taken as input by the classifiers. $2$D-driven $3$D methods \cite{120,92,85} first extract a $2$D BB around the object of interest ($2$D BB generation block), which is then lifted to $3$D. Depending on the input, the methods in \cite{92,85,93,e160,111,80,27} employ a pre-processing step on the input image and then generate $3$D hypotheses (input pre-processing block). $6$D object pose estimators \cite{27,28,29,33} extract features from the input images (feature extraction block), and using the trained classifiers, estimate objects' $6$D pose. Several methods further refine the output of the trained classifiers \cite{120,85,e160,28,29,33} (refinement block), and finally hypothesize the object pose after filtering.\\
\indent Table \ref{table_cla} details the classification-based methods. GS$3$D \cite{120} concentrates on extracting the $3$D information hidden in a $2$D image to generate accurate $3$D BB hypotheses. It modifies Faster R-CNN \cite{133} to classify the rotation $\theta_y$ in RGB images in addition to the $2$D BB parameters. Utilizing another CNN architecture, it refines the object's pose parameters further classifying $\mathbf{x}, \mathbf{d}$, and $\theta_y$. Papon et al. \cite{92} estimate semantic poses of common furniture classes in complex cluttered scenes. The input is converted into the intensity (RGB) \& surface normal (D) image. $2$D BB proposals are generated using the $2$D GOP detector \cite{138}, which is then lifted to $3$D space further classifying $\theta_y$ and $z$ using the bin-based cross entropy loss $L_{ce}$. Gupta et al. \cite{85} start with the segmented region output from \cite{79}, the depth channel of which is further processed to acquire surface normal image. The rotation $\theta_y$ of the segmented region is classified by the $L_{ce}$ loss. The method further refines the estimated coarse pose employing ICP over objects' CAD models. The input depth image of the SVM-based Sliding Shapes (SS) \cite{93} is pre-processed to obtain voxelized point cloud. The classifier is trained using the hinge loss $L_{hinge}$ to classify the parameters $\mathbf{x}, \mathbf{d}$, and $\theta_y$. Ren et al. \cite{e160} propose Cloud of Oriented Gradients (COG) descriptors which link $2$D appearance to $3$D point cloud. The method produces $3$D box proposals, the parameters $\mathbf{x}$, $\mathbf{d}$, and $\theta_y$ of which are after NMS filtering further refined by another classifier (s-SVM) to obtain the final detections. The sparse nature of point clouds is leveraged in Wang et al. \cite{111} employing a voting scheme which enables a search over all putative object locations. It estimates the parameter $\mathbf{x}$ by using an s-SVM after converting the LIDAR image into $3$D grid. As conducted in Wang et al. \cite{111}, the input LIDAR image of Vote$3$Deep \cite{80} is pre-processed to acquire sparse $3$D grid, which is then fed into the CNN architecture to classify $\mathbf{x}$ and $\theta_y$ parameters.\\
\begin{figure}[!t]
\centering
\includegraphics[width=3.4in]{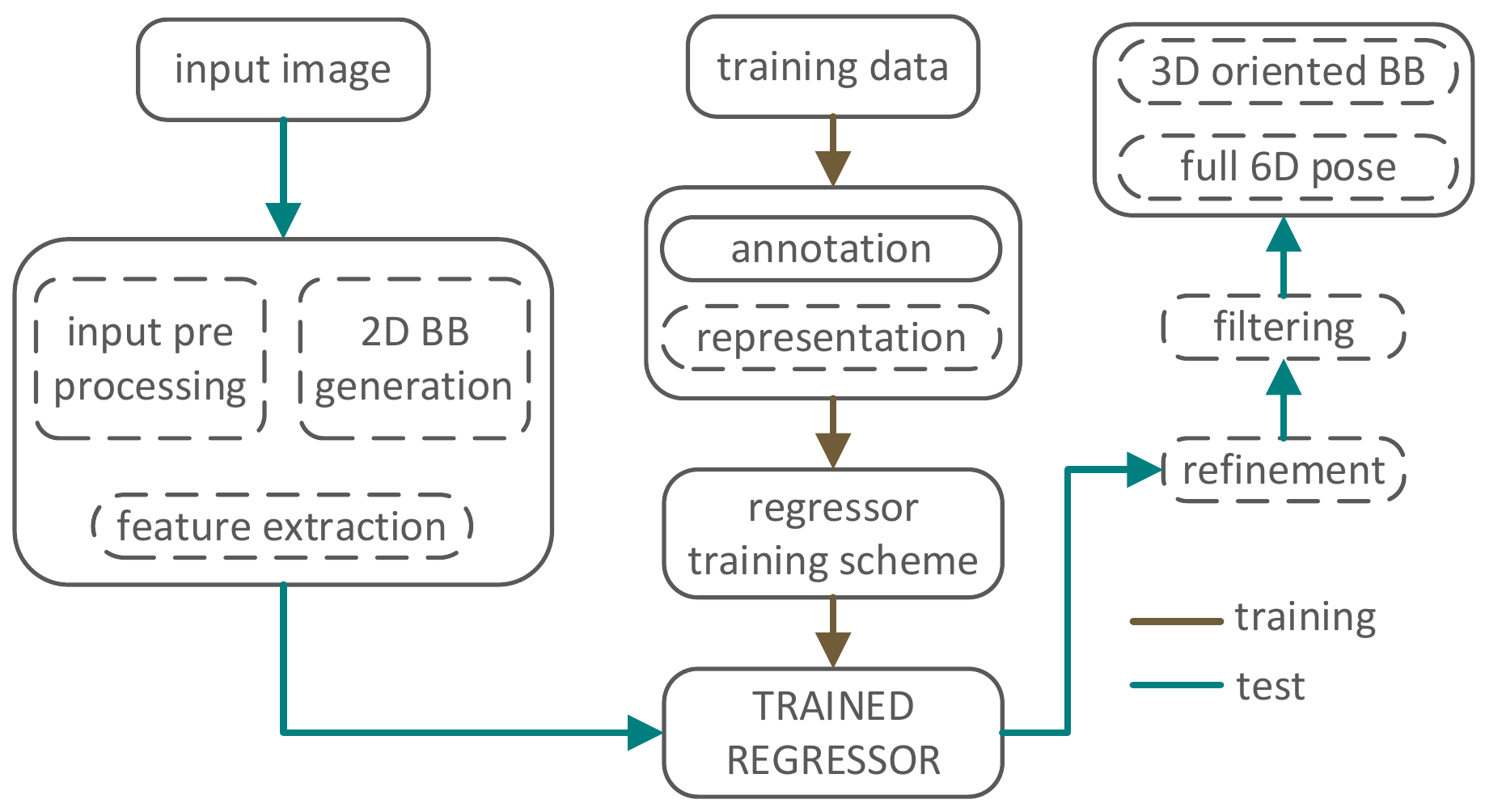}
\caption{Overall schematic representation of regression-based methods: blocks drawn with continuous lines are employed by all methods while dashed-line blocks are operated by clusters of specific methods (see the text and Table \ref{table_reg} for details).}
\label{fig_reg}
\vspace{-1em}
\end{figure}
\indent The search space of the $3$D BB detectors is further enlarged to $6$D \cite{27, 28, 29, 33}. The random forest-based method of Bonde et al. \cite{27} estimates objects' $6$D pose in depth images in the simultaneous presence of occlusion, clutter, and similar-looking distractors. It is trained based on the information gain $IG$ to classify $\mathbf{\Theta}$. Despite the fact that this method classifies $3$D rotation $\mathbf{\Theta}$, after extracting \textit{edgelet} features, it voxelizes the scene and selects the center of the voxel, which has the minimum distance between edgelet features, as the center of the detected object. Brachmann et al. \cite{28} introduce a random forest-based method for the $6$D object pose estimation problem.  Unlike Bonde et al. \cite{27}, this method classifies $3$D translation parameters $\mathbf{x}$ of the objects of interest, the exact $6$D poses of which are obtained using ICP-variant algorithm. Based upon Brachmann et al. \cite{28}, Krull et al. \cite{29} and Michel et al. \cite{33} present novel RGB-D-based methods for $6$D object pose recovery. The main contributions of \cite{29} and \cite{33} are on the refinement step. Krull et al. \cite{29} train a novel CNN architecture which learns to compare observation and renderings, and Michel et al. \cite{33} engineer a novel Conditional Random Field (CRF) for generating a pool of pose hypotheses.\\
\indent The training phase of the methods \cite{120,e160,111,80} are based on real data, while the ones in \cite{85,27} are trained using synthetic images. Papon et al. \cite{92} train the method based on synthetic images, however they additionally utilize real positive data to fine-tune the classifier, a heuristic that improves the accuracy of the method across unseen instances. Sliding Shapes \cite{93} uses positive synthetic and negative real images to train the SVMs. Brachmann et al. \cite{28}, and accordingly Krull et al. \cite{29} and Michel et al. \cite{33} train the forests based on positive and negative real data, and positive synthetic data. Amongst the methods, the hypotheses of the ones presented in \cite{92,93,e160,111,80} are NMS-filtered. All $2$D-driven $3$D methods and $3$D methods work at the level of categories. Full $6$D pose estimators are designed for instance-level object pose estimation.
\begin{table*}[t]
\tiny
\fontsize{10}{12.2}\selectfont
\caption{Regression-based methods ({\small BEV: \textbf{B}ird's-\textbf{E}ye \textbf{V}iew}, {\small CNN: \textbf{C}onvolutional \textbf{N}eural \textbf{N}etwork}, {\small FV: \textbf{F}ront \textbf{V}iew}, {\small HF: \textbf{H}ough \textbf{F}orest}, {\small LMA: \textbf{L}evenberg-\textbf{M}arquardt \textbf{A}lgorithm}, {\small MLP: \textbf{M}ulti-\textbf{L}ayer \textbf{P}erceptron}, {\small Mo/St: \textbf{Mo}no/\textbf{St}ereo}, {\small NMS: \textbf{N}on-\textbf{M}aximum \textbf{S}uppression}, {\small ORN: \textbf{O}bject \textbf{R}ecognition \textbf{N}etwork}, {\small p-LIDAR: \textbf{p}seudo-\textbf{LIDAR}}, {\small EPnP: \textbf{E}fficient \textbf{P}erspective-\textbf{n}-\textbf{P}oint}, {\small R: \textbf{R}eal data}, {\small RPN: \textbf{R}egion \textbf{P}roposal \textbf{N}etwork}, {\small S: \textbf{S}ynthetic data}, for symbols see Table \ref{table_denot})}
\centering
\setlength\tabcolsep{9pt}
\renewcommand{\arraystretch}{1.2}
\begin{minipage}{0.999\textwidth}
\centering
\resizebox{0.999\columnwidth}{!}
{
  \begin{tabular}{l c c c c c c c c c}
    \arrayrulecolor{black}\toprule
    {\small\textbf{method}} &{\small\textbf{input}} &{\small\textbf{input}}         &{\small\textbf{training}}  &{\small\textbf{regression}}       &{\small\textbf{regressor}}     &{\small\textbf{trained}}    &{\small\textbf{refinement}}  &{\small\textbf{filtering}} &{\small\textbf{level}}\\
                    &               &{\small\textbf{pre-processing}} &{\small\textbf{data}}      &{\small\textbf{parameters}}       &\textbf{training}      &{\small\textbf{regressor}}  &{\small\textbf{step}}        &                   &\\
    \arrayrulecolor{black}\bottomrule
    & & & & {\small \textbf{$2$D-DRIVEN $3$D} }\\
    \arrayrulecolor{black}\toprule
    Wang et al. \cite{117}      &{\small RGB(Mo/St)} &{\small p-LIDAR \& BEV} &{\small R}  &$\mathbf{x}, \mathbf{d}$                    &$L_{1_s}$         &{\small RPN}        &{\small CNN scoring}       &{\small NMS}     &category\\
    \textbf{refinement step}    &{\small RGB(Mo/St)} &{\small p-LIDAR \& BEV} &{\small R}  &$\{\mathbf{p}_{2D_i}\}_{i=1}^{4}, h_1, h_2, \theta_y$ &$L_{1_s}$         &{\small CNN}         &\xmark          &{\small NMS}     &\\
    \arrayrulecolor{gray}\midrule
    Deng et al. \cite{89}       &{\small RGB-D}      &\xmark                &{\small R}  &$\mathbf{x}, \mathbf{d}, \theta_y$          &$L_{1_s}$         &{\small CNN}         &\xmark                     &{\small NMS}           &category\\
    Lahoud et al. \cite{87}     &{\small RGB-D}      &\xmark                &{\small R}  &$\mathbf{d}$                                &$L_{1_s}$         &{\small MLP}         &{\small context-info}      &\xmark           &category\\
    PointFusion \cite{e157}     &{\small RGB-LIDAR}  &\xmark                &{\small R}  &$\{\mathbf{p}_{3D_i}\}_{i=1}^{8}$           &$L_{1_s}$         &{\small CNN}         &\xmark                     &{\small scoring} &category\\
    \arrayrulecolor{black}\bottomrule
    & & & & {\small \textbf{$3$D} }\\
    \arrayrulecolor{black}\toprule
    {\small Deep SS} \cite{82}  &{\small Depth}       &{\small $3$D grid}       &{\small R}         &$\mathbf{x}, \mathbf{d}$                    &$L_{1_s}$            &{\small RPN ($3$D)}      &{\small ORN}           &{\small NMS}          &category\\
    \textbf{refinement step}    &{\small RGB-D}       &{\small $3$D grid}       &{\small R}         &$\mathbf{x}, \mathbf{d}$                    &$L_{1_s}$            &{\small ORN}  &\xmark                 &{\small size pruning} & \\
    \arrayrulecolor{gray}\midrule
    Li et al. \cite{e161}       &{\small LIDAR}       &{\small cylindrical}     &{\small R}         &$\{\mathbf{p}_{3D_i}\}_{i=1}^{8}$           &$L_2$                &{\small CNN}  &\xmark                 &{\small NMS}          &category\\
    Bo Li \cite{e150}           &{\small LIDAR}       &{\small $3$D grid}       &{\small R}         &$\{\mathbf{p}_{3D_i}\}_{i=1}^{8}$           &$L_2$                &{\small CNN}  &\xmark                 &{\small NMS}          &category\\
    {\small PIXOR} \cite{e156}  &{\small LIDAR}       &{\small BEV}             &{\small R}         &$x, y, d_w, d_l, \theta_y$                  &$L_{1_s}$            &{\small CNN}  &\xmark                 &{\small NMS}          &category\\
    VoxelNet \cite{83}          &{\small LIDAR}       &{\small $3$D grid, BEV}    &{\small R}         &$\mathbf{x}, \mathbf{d}, \theta_y$          &$L_{1_s}$            &{\small RPN}  &\xmark                 &\xmark                &category\\
    \arrayrulecolor{gray}\midrule
    {\small MV$3$D} \cite{86}     &{\small LIDAR}       &{\small BEV}             &{\small R}         &$\mathbf{x}, \mathbf{d}$          &$L_{1_s}$            &{\small CNN}  &{\small FusionNet}     &{\small NMS}          &category\\
    \textbf{refinement step}    &{\small RGB-LIDAR}   &{\small BEV, FV}         &{\small R}         &$\{\mathbf{p}_{3D_i}\}_{i=1}^{8}$           &$L_{1_s}$            &{\small CNN}  &\xmark                 &{\small NMS}          &\\
    \arrayrulecolor{gray}\midrule
    Liang et al. \cite{e151}    &{\small RGB-LIDAR}   &{\small BEV}             &{\small R}         &$\mathbf{x}, \mathbf{d}, \theta_y$          &$L_{1_s}$            &{\small CNN}  &\xmark                 &{\small NMS}          &category\\
    \arrayrulecolor{gray}\midrule
    {\small AVOD} \cite{124}    &{\small RGB-LIDAR}   &{\small BEV}             &{\small R}         &$\mathbf{x}, \mathbf{d}$                    &$L_{1_s}$            &{\small RPN}  &{\small CNN scoring}   &{\small NMS}          &category\\
    \textbf{refinement step}    &{\small RGB-LIDAR}   &{\small BEV}             &{\small R}         &$\{\mathbf{p}_{2D_i}\}_{i=1}^{4}, h_1, h_2, \theta_y$ &$L_{1_s}$            &{\small CNN}  &\xmark                 &{\small NMS}          &        \\
    \arrayrulecolor{black}\bottomrule
    & & & & {\small \textbf{$6$D} }\\
    \arrayrulecolor{black}\toprule
    BB8 \cite{40}                 &{\small RGB}        &\xmark    &{\small R}            &$\{\mathbf{p}_{3D_i}^{proj}\}_{i=1}^{8}$         &$L_2$                        &{\small CNN}        &{\small PnP}             &\xmark   &instance\\
    Tekin et al. \cite{38}        &{\small RGB}        &\xmark    &{\small R}            &$\{\mathbf{p}_{3D_i}^{proj}\}_{i=1}^{8}, \mathbf{p}_{2D_c}$          &$L_2$                        &{\small CNN}        &{\small PnP}             &\xmark   &instance\\ 
    Oberweger et al. \cite{b173}  &{\small RGB}        &\xmark    &{\small R \& S}       &$\{\mathbf{p}_{3D_i}^{proj}\}_{i=1}^{8}$         &$L_2$                        &{\small CNN}        &{\small PnP \& RANSAC}   &\xmark   &instance\\
    CDPN \cite{b177}              &{\small RGB}        &\xmark    &{\small R \& S}       &$\mathbf{x}$                              &$L_1$                        &{\small CNN}        &{\small PnP \& RANSAC}   &\xmark   &instance\\
    Pix2Pose \cite{b178}          &{\small RGB}        &\xmark    &{\small R \& S}       &$\mathbf{x}$                              &$L_1$                        &{\small CNN}        &{\small PnP \& RANSAC}   &\xmark   &instance\\
    Hu et al. \cite{b169}         &{\small RGB}        &\xmark    &{\small R \& S}       &$\{\mathbf{p}_{3D_i}^{proj}\}_{i=1}^{8}$         &$L_1$                        &{\small CNN}        &{\small EPnP}            &\xmark   &instance\\
    PVNet \cite{b170}             &{\small RGB}        &\xmark    &{\small R \& S}       &$\{\mathbf{p}_{3D_i}^{proj}\}_{i=1}^{8}$         &$L_{1_s}$                    &{\small CNN}        &{\small EPnP \& LMA}     &\xmark   &instance\\
    {\small IHF} \cite{31}        &{\small Depth}      &\xmark    &{\small S}            &$\mathbf{x}, \mathbf{\Theta}$             &{\small offset \& pose entr} &{\small HF}         &{\small co-tra}          &\xmark   &instance\\
    Sahin et al. \cite{32}        &{\small Depth}      &\xmark    &{\small S}            &$\mathbf{x}, \mathbf{\Theta}$             &{\small offset \& pose entr} &{\small HF}         &{\small co-tra}          &\xmark   &instance\\
    {\small LCHF} \cite{4}        &{\small RGB-D}      &\xmark    &{\small S}            &$\mathbf{x}, \mathbf{\Theta}$             &{\small offset \& pose entr} &{\small HF}         &{\small co-tra}          &\xmark   &instance\\
    Doumanoglou et al. \cite{35}  &{\small RGB-D}      &\xmark    &{\small S}            &$\mathbf{x}, \mathbf{\Theta}$             &{\small offset \& pose entr} &{\small HF}         &{\small joint reg}       &\xmark   &instance\\
    DenseFusion \cite{b171}       &{\small RGB-D}      &\xmark    &{\small R}            &$\mathbf{x}, \mathbf{\Theta}$             &$L_1$                        &{\small CNN}        &{\small it ref}          &\xmark   &instance\\
    \arrayrulecolor{black}\bottomrule
  \end{tabular}
}
\end{minipage}%
\label{table_reg}
\end{table*}
\subsection{Regression}
Figure \ref{fig_reg} demonstrates the overall schematic representation of the regression-based methods. The training and the testing phases of the regression-based methods are similar to that of the classification-based methods. During an off-line stage, regressors are trained based on real or synthetic annotated data using the related loss functions. In an on-line stage, a real image is taken as input by any of the regressors. $2$D-driven $3$D methods \cite{117,89,87,e157} first extract a $2$D BB around the object of interest ($2$D BB generation block), which is then lifted to $3$D. Input pre-processing (input pre-processing block) is conducted by several methods \cite{117,82,e161,e150,e156,83,86,e151,124} to make the input data suitable for the trained regressor. $6$D object pose estimators \cite{40,38,b173,b177,b178,b169,b170,31,32,4,35,b171} extract features from the input images, and using the trained regressor, estimate objects' $6$D pose. Several methods further refine the output of the trained regressors \cite{117,87,82,86,124,40,38,b173,b177,b178,b169,b170,31,32,4,35,b171} (refinement block), and finally hypothesize the object pose after filtering.\\
\indent Table \ref{table_reg} elaborates the regression-based methods. Wang et al. \cite{117} argue that the performance gap in between the LIDAR-based and the RGB-based methods is not the quality of the data, but the representation. They convert image-based depth maps acquired from RGB (Mo/St) to pseudo-LiDAR (p-LIDAR) representation, which is then fed as input to the regression-based detection algorithm AVOD \cite{124} (This representation is also given as input to FrustumPNet \cite{116} and will be discussed in the \textit{classification \& regression} subsection). The RGB-D-based method of Deng et al. \cite{89} first detects $2$D BBs using Multiscale Combinatorial Grouping (MCG) \cite{79, 146, 147}. Engineering a Fast R-CNN-based architecture, it regresses $\mathbf{x}, \mathbf{d}$, and $\theta_y$. Another RGB-D-based $2$D-driven $3$D method \cite{87} generates $2$D BB hypotheses using Faster R-CNN, and then regresses $\mathbf{d}$ exploiting a Multi-Layer Perceptron (MLP), whose training is based on $L_{1_s}$. Unlike \cite{89}, the hypotheses produced by MLP are further refined utilizing the context information (context-info) available in the $3$D space. PointFusion \cite{e157} takes RGB and LIDAR images of a $2$D detected object as input. The features extracted from the RGB channel using a CNN and from raw depth data using a PointNet are combined in a fusion net to regress $3$D BBs.\\
\indent Regression-based $3$D object detectors directly reason about $3$D BBs of the objects of interest. Deep SS \cite{82} introduces the first fully conv $3$D region proposal network (RPN), which converts the input depth volume into the $3$D voxel grid and regresses $\mathbf{x}$ and $\mathbf{d}$. A joint object recognition network (ORN) scores the proposals taking their $3$D volumes and corresponding RGB images. Both nets are trained based on the $L_{1_s}$ loss. The proposal-free, single stage $3$D detection method of PIXOR \cite{e156} generates pixel-wise predictions, each of which corresponds to a $3$D object estimate. As it operates on the Bird's Eye View (BEV) representation of LIDAR point cloud, it is amenable to real-time inference. VoxelNet \cite{83}, as an end-to-end trainable net, interfaces sparse LIDAR point clouds with RPN representing voxelized raw point cloud as sparse 4D tensors via a feature learning net. Using this representation, a Faster R-CNN-based RPN generates $3$D detection, regressing the $\mathbf{x}$, $\mathbf{d}$, and $\theta_y$ parameters whose training is employed with $L_{1_s}$. MV$3$D \cite{86} processes the point cloud of LIDAR and acquires the BEV image. It firstly generates proposals regressing $\mathbf{x}$ and $\mathbf{d}$. Then, FusionNet \cite{b208} fuses the front view (FV) of LIDAR and RGB images of proposals to produce final detections regressing the $3$D box corners $\{\mathbf{p}_{3D_i}\}_{i=1}^{8}$. Liang et al. \cite{e151} is another end-to-end trainable method taking the advantages of both LIDAR and RGB information. It fuses RGB features onto the BEV feature map to regress the parameters $\mathbf{x}$, $\mathbf{d}$, and $\theta_y$. Aggregate View Object Detection (AVOD) net \cite{124} uses LIDAR point clouds and RGB images to generate features that are shared by two sub-nets. The first sub-net, an RPN, regresses the parameters $\mathbf{x}$ and $\mathbf{d}$ to produce $3$D BB proposals, which are then fed to the second sub-net that refines the proposals further regressing $\{\mathbf{p}_{2D_i}\}_{i=1}^{4}$, $h_1$, $h_2$, and $\theta_y$.\\
\indent The regression-based $6$D object pose estimators are designed to estimate full $6$D pose parameters. BB8 \cite{40} firstly segments the object of interest in RGB image, and then feed the region into a CNN which is trained to predict the projections of $8$ corners of the $3$D BB. Full $6$D pose is resolved applying a PnP algorithm the $2$D-$3$D correspondences. The method additionally addresses the symmetry issue, narrowing down the range of rotation around the axis of symmetry of the training samples from $0$ to the angle of symmetry $\gamma$. Tekin et al. \cite{38} follow a similar approach to estimate objects' $6$D pose, however, unlike BB8, they do not employ any segmentation and directly regress $\{\mathbf{p}_{3D_i}^{proj}\}_{i=1}^{8}$ along with $\mathbf{p}_{2D_c}$. Oberweger et al. \cite{b173} present a method which is engineered to handle occlusion. The method extracts multiple patches from an image region centered on the object and predicts heatmaps over the $2$D projections of $3$D points. The heatmaps are aggregated and the $6$D pose is estimated using a PnP algorithm. CDPN \cite{b177} disentangles the estimation of translation and rotation, separately predicting $3$D coordinates for all image pixels and indirectly computing the rotation parameters from the predicted coordinates via PnP. Pix2Pose \cite{b178} addresses the challenges of the problem, i) occlusion: estimating the $3$D coordinates per-pixel and generative adversarial training, ii) symmetry: introducing a novel $L_1$-based transformer loss, and iii) lack of precise $3$D object models: using RGB images without textured models during training phase. Full $6$D pose is obtained employing PnP \& RANSAC over the predicted $3$D coordinates. PVNet \cite{b170} estimates full $6$D poses of the objects of interest under severe occlusion or truncation. To this end, it firstly regresses the vectors from pixels to the projections of the corners of the BBs, and then votes for the corner locations. Lastly, the $6$D pose is computed via EPnP \& Levenberg-Marquardt Algorithm (LMA). IHF \cite{31}, Sahin et al. \cite{32}, LCHF \cite{4}, and Doumanoglou et al. \cite{35} directly regress the $6$D pose parameters training Hough Forests (HF) with offset \& pose entropy functions. IHF \cite{31} and Sahin et al. \cite{32} are engineered to work in depth images, while LCHF \cite{4} and Doumanoglou et al. \cite{35} take RGB-D as input. The $6$D pose parameters are further refined conducting a co-training scheme \cite{31,32,4} and employing joint registration (joint reg) \cite{35}.\\
\begin{figure}[!t]
\centering
\includegraphics[width=3.4in]{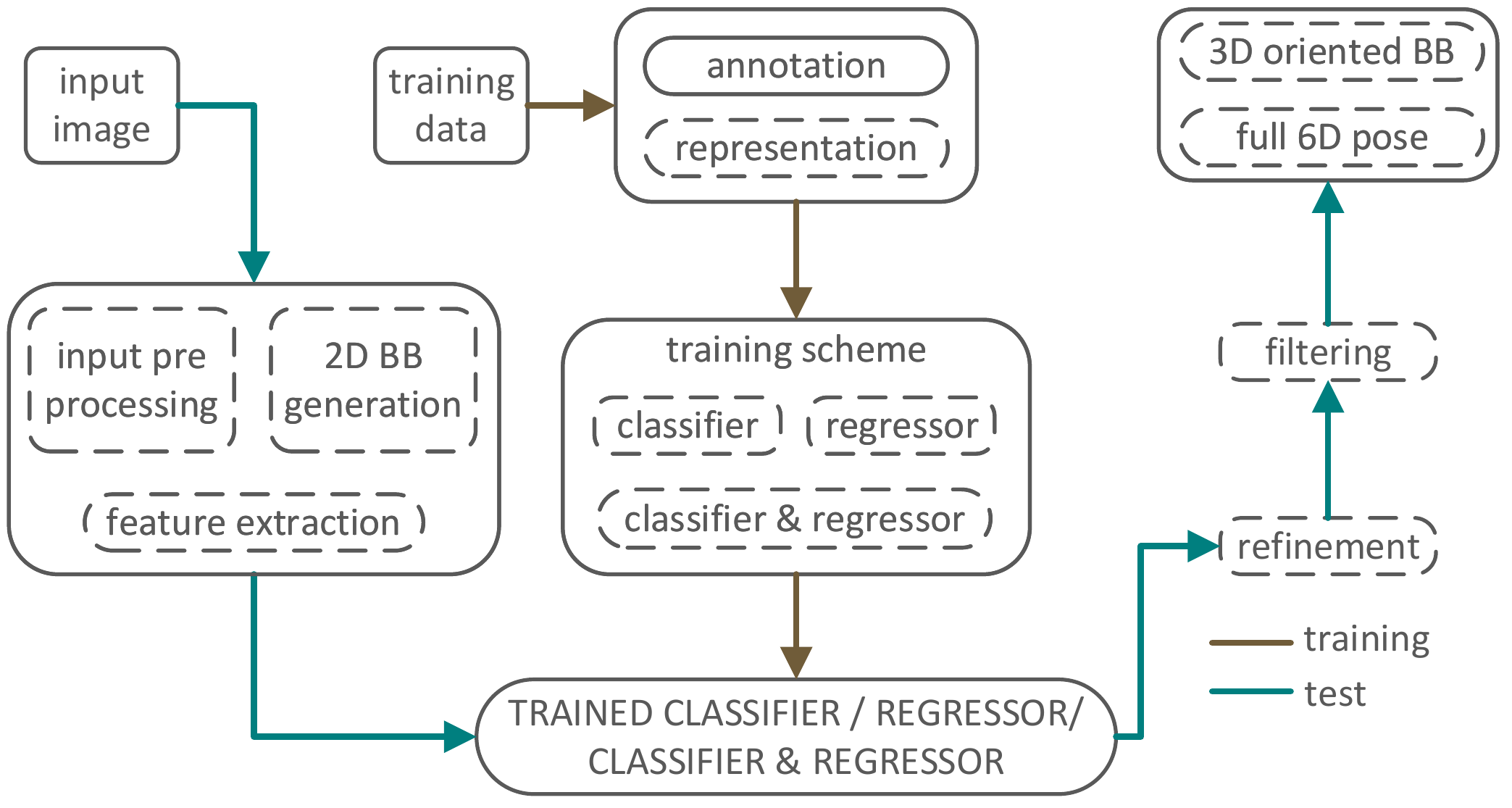}
\caption{Overall schematic representation of classification \& regression-based methods: blocks drawn with continuous lines are employed by all methods while dashed-line blocks are operated by clusters of specific methods (see the text and Table \ref{table_cla_reg} for details).}
\label{fig_cla_reg}
\vspace{-1.5em}
\end{figure}
\indent The $2$D-driven $3$D methods, $3$D methods, and the ones in \cite{40,38,b171} are trained using real data, while the training of the methods in \cite{31,32,4,35} are conducted with synthetic renderings. \cite{b173,b177,b178,b169,b170} use both real and synthetic images to train. The $2$D-driven $3$D methods and the $3$D BB detectors work at the level of categories, and the $6$D methods \cite{40,38,b173,b177,b178,b169,b170,31,32,4,35,b171} work at instance-level.
\begin{table*}[t]
\tiny
\fontsize{10}{12.2}\selectfont
\caption{Classification \& Regression-based methods ({\small BEV: \textbf{B}ird's-\textbf{E}ye \textbf{V}iew}, {\small CNN: \textbf{C}onvolutional \textbf{N}eural \textbf{N}etwork}, {\small CRF: \textbf{C}onditional \textbf{R}andom \textbf{F}ield}, {\small ICP: \textbf{I}terative \textbf{C}losest \textbf{P}oint}, {\small Mo/St: \textbf{Mo}no/\textbf{St}ereo}, {\small NMS: \textbf{N}on-\textbf{M}aximum \textbf{S}uppression}, {\small p-LIDAR: \textbf{p}seudo-\textbf{LIDAR}}, {\small PnP: \textbf{P}erspective-\textbf{n}-\textbf{P}oint}, {\small R: \textbf{R}eal data}, {\small RF: \textbf{R}andom \textbf{F}orest}, {\small S: \textbf{S}ynthetic data}, {\small SSD: \textbf{S}ingle-\textbf{S}hot \textbf{D}etector}, {\small s-SVM: \textbf{s}tructured-\textbf{S}upport \textbf{V}ector \textbf{M}achine}, {\small SVR: \textbf{S}upport \textbf{V}ector \textbf{R}egressor}, for symbols see Table \ref{table_denot})}
\centering
\setlength\tabcolsep{3pt}
\renewcommand{\arraystretch}{1.2}
\begin{minipage}{0.999\textwidth}
\centering
\resizebox{0.999\columnwidth}{!}{
  \begin{tabular}{l c c c c c c c c c c c c}
    \arrayrulecolor{black}\toprule
    {\small\textbf{method}}                   &{\small\textbf{input}}   &{\small\textbf{input}}  &{\small\textbf{training}}    &{\small\textbf{classification}}  &{\small\textbf{classifier}} &{\small\textbf{trained}}     &{\small\textbf{regression}}    &{\small\textbf{regressor}}   &{\small\textbf{trained}}                                               &{\small\textbf{refinement}}   &{\small\textbf{filtering}}   &{\small\textbf{level}}\\
                                      &               &{\small\textbf{pre-processing}}  &{\small\textbf{data}}  &{\small\textbf{parameters}}      &{\small\textbf{training}}   &{\small\textbf{classifier}}  &{\small\textbf{parameters}}   &{\small\textbf{training}}    &{\small\textbf{regressor}} &{\small\textbf{step}}         &                     &\\
    \arrayrulecolor{black}\bottomrule
    & & & & & {\small \textbf{$2$D-DRIVEN $3$D} }\\
    \arrayrulecolor{black}\toprule 
    Mousavian et al. \cite{91}  &{\small RGB}        &\xmark                &{\small R}  &$\theta_y$  &$L_{ce}$  &{\small CNN}  &$\mathbf{d}, \theta_y$                      &$L_2$  &{\small CNN}         &{\small $2$D constraints}  &\xmark           &category\\
    Xu et al. \cite{e154}       &{\small RGB}        &\xmark                &{\small R}  &$\theta_y$  &$L_{ce}$  &{\small CNN}  &$\mathbf{x}, \mathbf{d}, \theta_y$          &$L_{1_s}$         &{\small CNN}         &\xmark                     &\xmark           &category\\
    Wang et al. \cite{117}      &{\small RGB(Mo/St)} &{\small p-LIDAR}      &{\small R}  &$\mathbf{d}, \theta_y$  &$L_{ce}$  &{\small CNN}  &$\mathbf{x}, \mathbf{d}, \theta_y$          &$L_{1_s}$         &{\small FrustumPNet} &\xmark                     &\xmark           &category\\
    \arrayrulecolor{gray}\midrule
    {\small MonoPSR} \cite{122} &{\small RGB}        &\xmark                &{\small R}  &$\theta_y$ &$L_{ce}$ &{\small CNN} &$\mathbf{d}, \theta_y$                      &$L_{1_s}$         &{\small CNN}         &{\small CNN scoring}       &\xmark           &category\\
    \textbf{refinement step}    &{\small RGB}        &\xmark                &{\small R}  &\xmark &\xmark &\xmark &$y,z$                                       &$L_{1_s}$         &{\small CNN}         &\xmark                     &\xmark           &        \\
    \arrayrulecolor{gray}\midrule                                
    FrustumPNet \cite{116}      &{\small RGB-D}      &\xmark  &{\small R}  &$\mathbf{d}, \theta_y$ &$L_{ce}$ &{\small CNN} &$\mathbf{x}, \mathbf{d}, \theta_y$          &$L_{1_s}$         &{\small PointNet}    &\xmark                     &\xmark           &category\\
    \arrayrulecolor{black}\bottomrule
    & & & & & {\small \textbf{$3$D} }\\
    \arrayrulecolor{black}\toprule 
    Chen et al. \cite{121}          &{\small RGB(St)}   &{\small $3$D grid}       &{\small R}              &$\mathbf{x}, \mathbf{d}, \theta_y$    &$IoU_{3D}$       &{\small s-SVM}            &\xmark      &\xmark        &\xmark      &CNN  &{\small NMS}   &category\\
    \textbf{refinement step}        &{\small RGB(Mo)}   &\xmark       &{\small R}              &\xmark                  &\xmark            &\xmark                    &$x,y,d_w,d_h, \theta_y$   &$L_{1_s}$         &{\small CNN}    &\xmark            &{\small NMS}   &\\
    \arrayrulecolor{gray}\midrule
    Chen et al. \cite{88}           &{\small RGB}       &\xmark             &{\small R}              &$\mathbf{x}, \mathbf{d}, \theta_y$    &$IoU_{3D}$       &{\small s-SVM}                  &\xmark                 &\xmark                 &\xmark      &CNN  &{\small NMS} &category\\
    \textbf{refinement step}        &{\small RGB}       &\xmark             &{\small R}              &\xmark                  &\xmark            &\xmark                    &$x,y,d_w,d_h,\theta_y$           &$L_{1_s}$         &{\small CNN}    &\xmark            &{\small NMS}   &        \\  
    \arrayrulecolor{gray}\midrule
    DeepContext\cite{81}            &{\small Depth}     &{\small $3$D grid}       &{\small R \& S} &$\mathbf{x}, \theta_y$          &$L_{ce}$      &{\small CNN}                  &\xmark                 &\xmark                 &\xmark      &{\small $3$DContextNet} &\xmark   &category\\
    \textbf{refinement step}        &{\small Depth}     &{\small $3$D grid}       &{\small R \& S} &\xmark                  &\xmark            &\xmark                    &$\mathbf{x}, \mathbf{d}$       &$L_{1_s}$         &{\small CNN}    &\xmark            &\xmark   &        \\
    \arrayrulecolor{gray}\midrule
    Lin et al. \cite{78}             &{\small RGB-D}    &\xmark &{\small R}     &\xmark       &\xmark   &\xmark         &$\mathbf{x}$   &$L_{il}$   &{\small SVR}  &{\small CRF}  &{\small NMS} &category\\
    \textbf{refinement step}         &{\small RGB-D}    &\xmark &{\small R}     &$\mathbf{x}$ &$L_{0/1}$ &{\small CRF}   &\xmark         &\xmark             &\xmark        &\xmark        &\xmark       &\\
    \arrayrulecolor{gray}\midrule
    {\small SECOND} \cite{e158} &{\small LIDAR}       &{\small $3$D grid, BEV}                   &{\small R}   &$\theta_y$ &$L_{ce}$ &{\small CNN}      &$\mathbf{x}, \mathbf{d}, \theta_y$          &$L_{1_s}$            &{\small CNN}  &\xmark                 &\xmark                &category\\
    PointPillars \cite{119}         &{\small LIDAR}     &{\small p-pillars} &{\small R}              &$\theta_y$                &$L_{ce}$      &{\small SSD}                  &$\mathbf{x}, \mathbf{d}, \theta_y$       &$L_{1_s}$         &{\small SSD}    &\xmark            &{\small NMS}   &category\\
    \arrayrulecolor{gray}\midrule
    PointRCNN \cite{118}        &{\small LIDAR}       &\xmark     &{\small R}   &$x, z, \theta_y$ &$L_{ce}$ &{\small CNN}      &$\mathbf{x}, \mathbf{d}, \theta_y$          &$L_{1_s}$ &{\small CNN}  &{\small CNN, canonical ref} &{\small NMS}          &category \\
    \textbf{refinement step}    &{\small LIDAR}       &\xmark     &{\small R}   &$x, z, \theta_y$ &$L_{ce}$ &{\small CNN}      &$\mathbf{x}, \mathbf{d}, \theta_y$          &$L_{1_s}$ &{\small CNN}  &\xmark                 &{\small NMS}          &         \\
    \arrayrulecolor{black}\bottomrule
    & & & & & {\small \textbf{$6$D} }\\
    \arrayrulecolor{black}\toprule
    Brachmann et al. \cite{30}    &{\small RGB}        &\xmark    &{\small R}  &$\mathbf{x}$ &$IG$ &{\small RF}     &$\mathbf{x}$                              &$L_1$          &{\small RF}         &{\small PnP \& RANSAC \& log-like}          &\xmark   &instance\\
    SSD-$6$D \cite{37}              &{\small RGB}        &\xmark                 &{\small R \& S}         &$\theta_{cvp}, \theta_{ip}$   &$L_{ce}$      &{\small CNN}                  &$\{\mathbf{p}_{2D_i}\}_{i=1}^{4}$     &$L_{1_s}$         &{\small CNN}    &ICP          &{\small NMS}   &instance\\
    \arrayrulecolor{gray}\midrule
    DPOD \cite{b176}                &{\small RGB}        &\xmark                 &{\small R \& S}         &$\mathbf{x}$                &$L_{ce}$            &{\small CNN}                  &\xmark              &\xmark           &\xmark    &{\small CNN} &\xmark   &instance\\
    \textbf{refinement step}        &{\small RGB}        &\xmark                 &{\small R \& S}         &\xmark              &\xmark             &\xmark                             &$\mathbf{x}, \mathbf{\Theta}$                           &$L_1$           &{\small CNN}    &\xmark   &\xmark   &\\
    \arrayrulecolor{gray}\midrule
    Li et al. \cite{b172}           &{\small RGB-D}      &\xmark                 &{\small R \& S}         &$\mathbf{x}, \mathbf{\Theta}$        &$L_{ce}$    &{\small CNN}                  &$\mathbf{x}, \mathbf{\Theta}$                           &$L_2$           &{\small CNN}    &{\small multi view} &\xmark   &instance\\
    \textbf{refinement step}        &{\small RGB-D}      &\xmark                 &{\small R \& S}         &$\mathbf{x}, \mathbf{\Theta}$        &$L_{ce}$    &{\small CNN}                  &$\mathbf{x}, \mathbf{\Theta}$                           &$L_2$           &{\small CNN}    &{\small ICP}   &\xmark   &\\
    \arrayrulecolor{black}\bottomrule
  \end{tabular}
}
\end{minipage}%
\label{table_cla_reg}
\end{table*}
\subsection{Classification \& Regression}
Figure \ref{fig_cla_reg} depicts the overall schematic representation of the classification \& regression-based methods. Unlike the previous categories of methods, \textit{i.e.}, classification-based and regression-based, this category performs the classification and regression tasks within a single architecture. The methods can firstly do the classification, the outcomes of which are cured in a regression-based refinement step \cite{121,88,81, b176} or vice versa \cite{78}, or can do the classification and regression in a single-shot process \cite{91,e154,117,122,116,e158,119,118,30,37,b172}.\\
\indent Table \ref{table_cla_reg} expands on the classification \& regression-based methods. Mousavian et al. \cite{91} modify MS-CNN \cite{139} so that the parameters $\mathbf{d}$ and $\theta_y$ are regressed in addition to the $2$D BB of the object of interest. The regression of $\mathbf{d}$ is conducted by the $L_2$ loss, while the bin-based discrete-continuous loss is applied to firstly discretize $\theta_y$ into $n$ overlapping bins, and then to regress the angle within each bin. The input of MonoPSR \cite{122} is an RGB image, which is not subjected to any pre-processing. Once the $2$D BB proposals for the the object of interest are generated using MS-CNN \cite{139}, MonoPSR hypothesizes $3$D proposals, which are then fed into a \textit{CNN scoring} refinement step. The scoring net, in order to eliminate low confidence proposals and to produce final detection, regresses $y$ and $z$. RGB-D-based method, FrustumPNet \cite{116}, detects $2$D BBs in RGB images using FPN \cite{135}. It does not pre-process the depth channel of the input RGB-D image, since it is capable of processing raw point clouds, as in PointNets \cite{148, 149}. Once the $2$D BB proposals are gathered, they are lifted to the $3$D space classifying and regressing $\mathbf{d}$ and $\theta_y$, and regressing $\mathbf{x}$. The regression of $\mathbf{x}$ is provided training the net with $L_{1_s}$, while $\mathbf{d}$ and $\theta_y$ are regressed using the bin-based version of $L_{1_s}$, as employed in Faster R-CNN.\\
\indent The method of Chen et al. \cite{121} pre-processes the input RGB (St) images to obtain voxelized point clouds of the scene, and using this representation, detects the $3$D BBs of the objects of interest in two stages: in the first stage, an SVM is trained to generate $3$D BB proposals classifying $\mathbf{x}$, $\mathbf{d}$, and $\theta_y$. In the second stage, the BB proposals are scored to produce final detections. Another work of Chen et al. \cite{88} employs the same strategy as in \cite{121}, taking a monocular RGB image as input. The depth-based method, DeepContext \cite{81}, is formed of two CNNs. The first CNN is trained using $L_{ce}$ to generate $3$D BB proposals classifying $\mathbf{x}$ and $\theta_y$. The proposals are then sent to the second CNN, which further regresses $\mathbf{x}$ and $\mathbf{d}$ to produce the final detections. RGB-D based method \cite{78} first regresses $\mathbf{x}$ using Support Vector Regressor (SVR), which is further refined using a CRF whose training is based on the loss $L_{0/1}$. PointPillars \cite{119} converts the input depth image of LIDAR into the pseudo image of point-pillars (p-pillars), and using this representation, outputs $\mathbf{x}$, $\mathbf{d}$, and $\theta_y$ to generate $3$D BBs of the objects of interest. PointRCNN \cite{118} detects the $3$D BBs in two stages: it first directly processes raw depth images acquired from LIDAR sensors to produce box proposals. In the second stage, generated proposals are refined using another net which transforms the points of each proposal to canonical coordinates. The parameters and the loss functions of the first stage are used to produce final detection results.\\
\indent Brachmann et al. \cite{30} present a random forest-based (RF) architecture which lifts the $3$D detection to $6$D space employing RANSAC \& log-likelihood algorithms. SSD-$6$D \cite{37} simultaneously classifies $\theta_{cvp}$ and $\theta_{ip}$ and regresses the corners $\{\mathbf{p}_{2D_i}\}_{i=1}^{4}$ of $2$D BBs. The training of the net is based on $L_{ce}$ and $L_{1_s}$. The output parameters of the net is lifted to the $6$D space, employing an ICP-based refinement along with the utilization of the camera intrinsics.\\
\indent The $2$D-driven $3$D methods and $3$D methods are trained using real data. DeepContext \cite{81} is trained on the partially synthetic training depth images which exhibit a variety of different local object appearances, and real data are used to fine tune the method. The $2$D-driven $3$D methods and the $3$D BB detectors work at the level of categories, and the $6$D methods \cite{30,37,b176,b172} work at instance-level.
\subsection{Template Matching}
\begin{figure}[!t]
\centering
\includegraphics[width=3.0in]{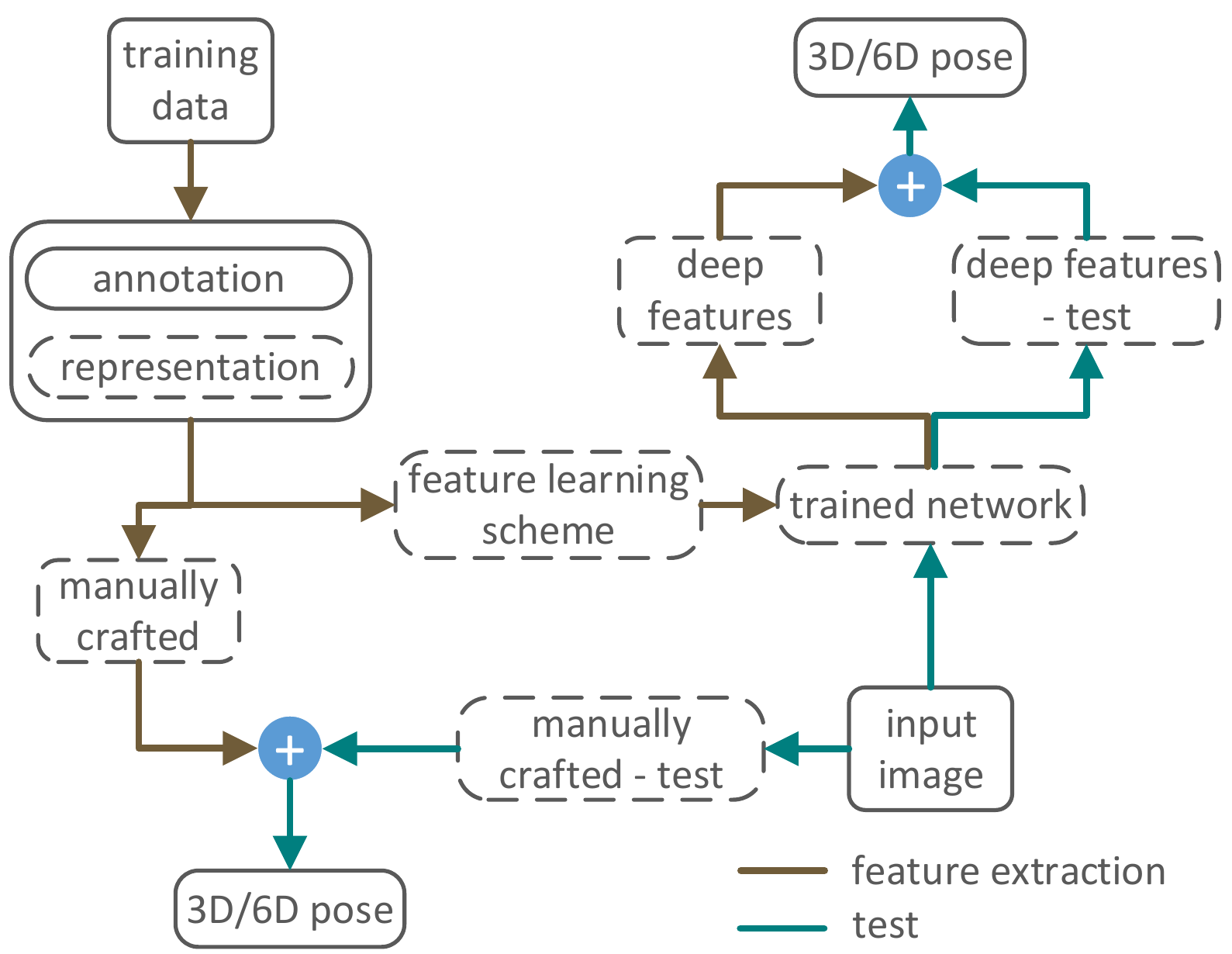}
\caption{Overall schematic representation of template matching methods: blocks drawn with continuous lines are employed by all methods while dashed-line blocks are operated by clusters of specific methods (see the text and Table \ref{table_template} for details).}
\label{fig_template}
\vspace{-1.4em}
\end{figure}

The overall schematic representation of the template matching methods is given in Fig. \ref{fig_template}.\\
\noindent \textbf{Feature Extraction Phase.} During an off-line feature extraction phase, $3$D pose \cite{b190, 34, 41, b191} or $6$D pose \cite{b186, b189, b188, b192, b187, 23, 2, 24, 25} annotated templates involved in the training data are represented with robust feature descriptors. Features are manually-crafted utilizing the available shape, geometry, and appearance information \cite{b186, b189, b188, b192, b187, 23, 2, 25}, and the recent paradigm in the field is to deep learn those using neural net architectures \cite{b190, 34, 41, b191}.\\
\noindent \textbf{Testing Phase.} A template-based method takes an image $\mathbf{I}$ as input on which it runs a sliding window during an on-line test phase. Each of the windows is represented with feature descriptors (either manually-crafted or deep features) and is compared with the templates stored in a memory. The matching is employed in the feature space, the distances between each window and the template set are computed. The pose parameters of a template are assigned to the window that has the closest distance with that template.\\
\indent Table \ref{table_template} details the template matching methods. Payet et al. \cite{b186} introduce Bag of Boundaries (BOB) representation, the histograms of the right contours belonging to the foreground object, and formulates the problem as matching the BOBs in the test image to the set of shape templates. The representation improves the robustness across clutter. Ulrich et al. \cite{b189} match test images to the set of templates using edge features. The method first estimates the discrete pose, which is further refined using the $2$D match and the corresponding $3$D camera pose in an LMA algorithm. Since the increase in the number of the templates in a database slows down the performance of template matching-based methods, Konishi et al. \cite{b188} present a method with a representation of robust to change in appearance Perspectively Cumulated Orientation Feature (PCOF) and an efficient search algorithm of Hierarchical Pose Trees (HPT). $6$D pose parameters initially estimated by edge matching-based Real-time Attitude and PosItion Determination-Linear Regressor (RAPID-LR) \cite{b192} are further refined by extracting HOG features on those edges and matching to the template database by RAPID-HOG distance. The method originally proposed for object tracking \cite{b187} can handle temporary tracking loss occurring when the object is massively occluded or is going out of the camera's viewpoint. It detects the object of interest matching the test image to the template database based on Temporally Consistent Local Color (TCLC) histograms. Liu et al. \cite{23}, estimating the $6$D poses of objects merely taking depth image as input, extract depth edges. MTTM \cite{b190} firstly computes the segmentation mask in the test image and then employs template matching over RoIs within a CNN framework, where the pose parameters are encoded with quaternions. Linemod \cite{2} and Hodan et al. \cite{25} represent each pair of templates by surface normals and color gradients, while \cite{24} trains SVMs to learn weights, which are then embedded into AdaBoost. Wohlhart et al. \cite{34} define the triplet loss $L_{tri}$ and the pair-wise loss $L_{pair}$ for training CNNs. $L_{tri}$ enlarges the Euclidean distance between descriptors from two different objects and makes the Euclidean distance between descriptors from the same object representative of the similarity between their poses. $L_{pair}$ makes the descriptors robust to noise and other distracting artifacts such as changing illumination. Balntas et al. \cite{41} further improve $L_{tri}$ and $L_{pair}$ guiding the templates with pose during CNN training ($L_{pose}$ loss). Zakharov et al. \cite{b191} introduce dynamic margin for the loss functions $L_{tri}$ and $L_{tri}$, resulting faster training time and enabling the use of smaller-size descriptors with no loss of accuracy. The methods \cite{34,41}, and \cite{b191} are based on RGB-D, and the deep features are learnt based on synthetic and real data.\\
\begin{table*}[t]
\tiny
\fontsize{10}{12.2}\selectfont
\caption{Template matching-based methods ({\small BOB: \textbf{B}ag \textbf{O}f \textbf{B}oundaries}, {\small CNN: \textbf{C}onvolutional \textbf{N}eural \textbf{N}etwork}, {\small FDCM: \textbf{F}ast \textbf{D}irectional \textbf{C}hamfer \textbf{M}atching}, {\small HOG: \textbf{H}istogram of \textbf{O}riented \textbf{G}radients}, {\small HPT: \textbf{H}ierarchical \textbf{P}ose \textbf{T}ree}, {\small ICP: \textbf{I}terative \textbf{C}losest \textbf{P}oint}, {\small k-NN: \textbf{k}-\textbf{N}earest \textbf{N}eighbor}, {\small LMA: \textbf{L}evenberg-\textbf{M}arquardt \textbf{A}lgorithm}, {\small PCOF: \textbf{P}erspectively \textbf{C}umulated \textbf{O}rientation \textbf{F}eature}, {\small PnP: \textbf{P}erspective-\textbf{n}-\textbf{P}oint}, {\small PSO: \textbf{P}article \textbf{S}warm \textbf{O}ptimization}, {\small R: \textbf{R}eal data}, {\small RAPID-LR: \textbf{R}eal-time \textbf{A}ttitude and \textbf{P}os\textbf{I}tion \textbf{D}etermination-\textbf{L}inear \textbf{R}egressor}, {\small S: \textbf{S}ynthetic data}, {\small SVM: \textbf{S}upport \textbf{V}ector \textbf{M}achine}, {\small TCLC: \textbf{T}emporally \textbf{C}onsistent \textbf{L}ocal \textbf{C}olor histograms}, for symbols see Table \ref{table_denot})}
\centering
\setlength\tabcolsep{14pt}
\renewcommand{\arraystretch}{1.2}
\begin{minipage}{0.999\textwidth}
\centering
\resizebox{0.999\columnwidth}{!}{
  \begin{tabular}{l c c c c c c c c c}
    \arrayrulecolor{black}\toprule
    {\small\textbf{method}} &{\small\textbf{input}} &{\small\textbf{training}} &{\small\textbf{annotation}} &{\small\textbf{manually crafted}} &{\small\textbf{feature learning}} &{\small\textbf{learnt}} &{\small\textbf{matching}} &{\small\textbf{refinement}}   &{\small\textbf{level}}\\
                    &               &{\small\textbf{data}}     &                    &{\small\textbf{feature}}                    &{\small\textbf{scheme}}    &{\small\textbf{features}}     &                  &{\small\textbf{step}}  & \\
    \arrayrulecolor{black}\bottomrule
    & & & & {\small \textbf{$6$D}} \\
    \arrayrulecolor{black}\toprule 
    Payet et al. \cite{b186}      &{\small RGB}    &{\small R}      &$\mathbf{x}, \mathbf{\Theta}$  &{\small BOB}                &\xmark                          &\xmark              &{\small convex opt}  &\xmark              &category\\
    Ulrich et al. \cite{b189}     &{\small RGB}    &{\small S}      &$\mathbf{x}, \mathbf{\Theta}$  &{\small edge}               &\xmark                          &\xmark              &{\small dot pro}     &{\small LMA}        &instance\\
    Konishi et al. \cite{b188}    &{\small RGB}    &{\small S}      &$z, \mathbf{\Theta}$           &{\small PCOF}               &\xmark                          &\xmark              &{\small HPT}         &{\small PnP}        &instance\\
    \arrayrulecolor{gray}\midrule
    {\small RAPID-LR} \cite{b192} &{\small RGB}    &{\small S}      &$\mathbf{x}, \mathbf{\Theta}$  &{\small edge}               &\xmark                          &\xmark              &{\small RAPID-LR}    &{\small RAPID-HOG}  &instance\\
    \textbf{refinement step}      &{\small RGB}    &{\small S}      &$\mathbf{x}, \mathbf{\Theta}$  &{\small HOG}                &\xmark                          &\xmark              &{\small RAPID-HOG}   &\xmark              &\\
    \arrayrulecolor{gray}\midrule
    Tjaden et al. \cite{b187}     &{\small RGB}    &{\small R \& S} &$\mathbf{x}, \mathbf{\Theta}$  &{\small TCLC}               &\xmark                          &\xmark              &{\small $2$D Euc}    &\xmark              &instance\\
    Liu et al. \cite{23}          &{\small Depth}  &{\small S}      &$\mathbf{x}, \mathbf{\Theta}$  &{\small edge}               &\xmark                          &\xmark              &{\small FDCM}        &{\small multi-view} &instance\\
    {\small MTTM} \cite{b190}     &{\small Depth}  &{\small S}      &$\mathbf{\Theta}$              &\xmark                      &$L_{tri}$, $L_{pose}$           &{\small CNN}        &{\small NN}          &{\small ICP}        &instance\\
    Linemod \cite{2}              &{\small RGB-D}  &{\small S}      &$\mathbf{x}, \mathbf{\Theta}$  &{\small normal, color grad} &\xmark                          &\xmark              &{\small NN}          &{\small ICP}        &instance\\
    Rios-Cabrera \cite{24}        &{\small RGB-D}  &{\small S}      &$\mathbf{x}, \mathbf{\Theta}$  &\xmark                      &{\small SVM}                    &{\small Ada-Boost}  &{\small NN}          &\xmark              &instance\\
    Hodan et al. \cite{25}        &{\small RGB-D}  &{\small S}      &$\mathbf{x}, \mathbf{\Theta}$  &{\small normal, color grad} &\xmark                          &\xmark              &{\small NN}          &{\small PSO}        &instance\\
    Wohlhart et al. \cite{34}     &{\small RGB-D}  &{\small R \& S} &$\mathbf{\Theta}$              &\xmark                      &$L_{tri}$, $L_{pair}$           &{\small CNN}        &{\small kNN}         &\xmark              &instance\\
    Balntas et al. \cite{41}      &{\small RGB-D}  &{\small R \& S} &$\mathbf{\Theta}$              &\xmark                      &$L_{tri}$, $L_{pose}$           &{\small CNN}        &{\small kNN}         &\xmark              &instance\\
    Zakharov et al. \cite{b191}   &{\small RGB-D}  &{\small R \& S} &$\mathbf{\Theta}$              &\xmark                      &$L_{tri_d}$, $L_{pair_d}$       &{\small CNN}        &{\small NN}          &\xmark              &instance\\
    \arrayrulecolor{black}\bottomrule
  \end{tabular}
}
\end{minipage}%
\label{table_template}
\end{table*}
\indent This family of the methods formulate the problem globally and represent the templates in the set and windows extracted from input images by feature descriptors holistically. Distortions along object borders arising from occlusion and clutter in the test processes mainly degrade the performance of these methods. Several imperfections of depth sensors, such as missing depth values, noisy measurements, \textit{etc.} impair the surface representations at depth discontinuities, causing extra degradation in the methods' performance. Other drawback is matching features extracted during test to a set of templates, and hence, it cannot easily be generalized well to unseen ground truth annotations.
\subsection{Point-pair Feature Matching}
\begin{figure}[!t]
\centering
\includegraphics[width=3.0in]{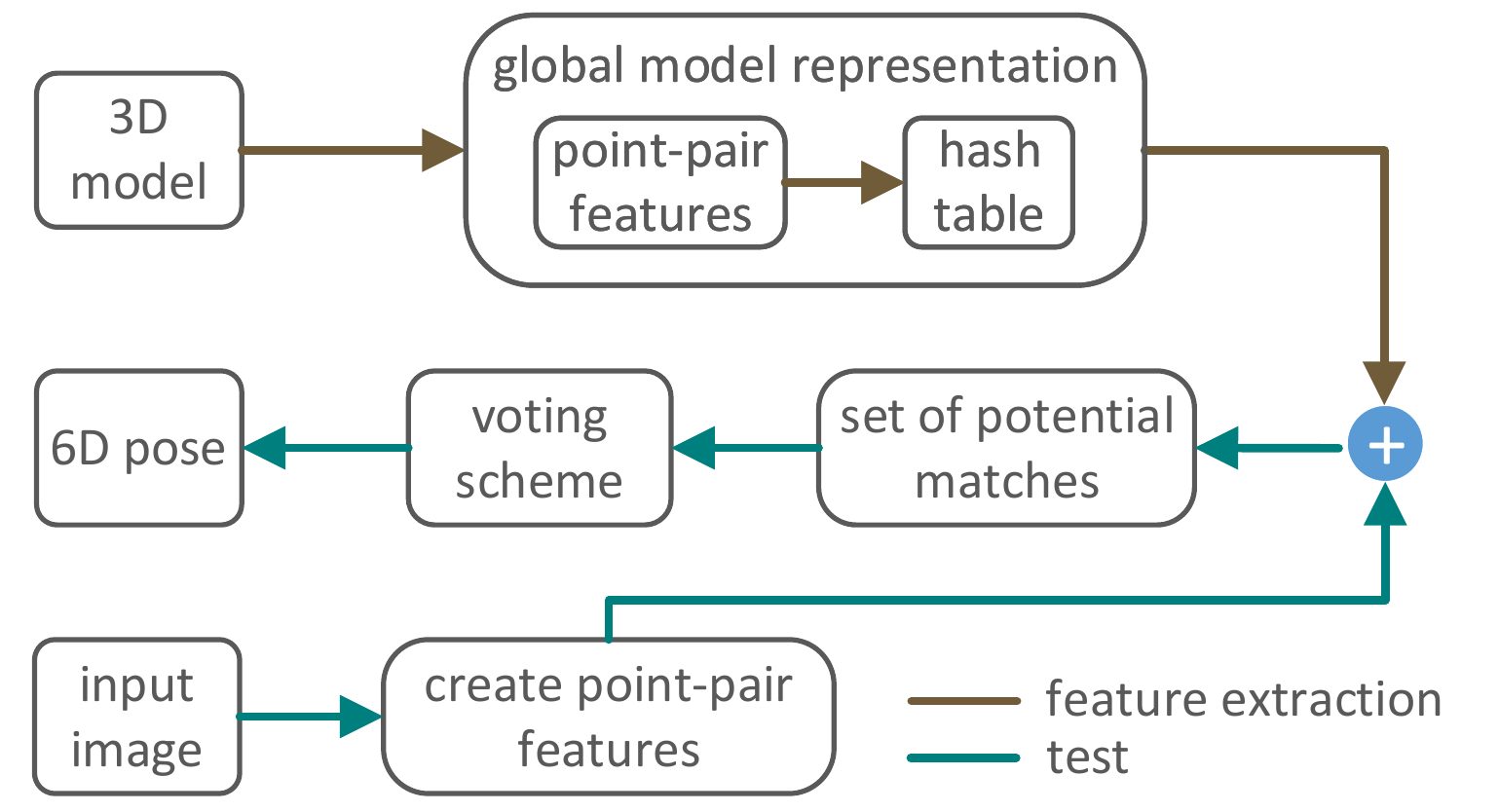}
\caption{Overall schematic representation of point-pair feature matching methods: (off-line phase) $3$D model of an object of interest is globally represented with point-pair features (PPFs). (on-line phase) Point-pair features created from the input depth image are compared with the global model representation in order to estimate the $6$D pose.}
\label{fig_point_pair}
\vspace{-1.5em}
\end{figure}

Figure \ref{fig_point_pair} represents the overall schematic layout of the point-pair feature matching methods. During an off-line phase, the global representation of the $3$D model of an object of interest is formed by point-pair features (PPF) stored in a hash table. In an online phase, the point-pair features extracted in the test image are compared with the global model representation, from which the generated set of potential matches vote for the pose parameters.\\
\indent We detail this category of methods initially focusing on the forefront publication from Drost et al. \cite{17}, and then presenting the approaches developed to improve PPF matching performance. Given two points $\mathbf{m}_1$ and $\mathbf{m}_2$ with normals $\mathbf{n}_1$ and $\mathbf{n}_2$, PPF is defined as follows \cite{17}:
\begin{equation}
\small
\mathbf{PPF}(\mathbf{m}_1, \mathbf{m}_2) = ( \vert \vert \mathbf{dist} \vert \vert_2, \angle (\mathbf{n}_1, \mathbf{dist}), \angle (\mathbf{n}_2, \mathbf{dist}), \angle (\mathbf{n}_1, \mathbf{n}_2))
\end{equation}
where $\mathbf{dist} = \mathbf{m}_2 - \mathbf{m}_1$, and $\angle (\mathbf{a}, \mathbf{b}) \in [0;\pi]$ denotes the angle between two vectors $\mathbf{a}$ and $\mathbf{b}$. Hence, PPF describes the relative translation and rotation of two oriented points. The global description of the model $M$ is created computing the feature vector PPF for all point pairs $ \{(\mathbf{m}_r, \mathbf{m}_i)\} \in M$ and is represented as a hash table indexed by the feature vector $\mathbf{PPF}$ \cite{16, 17}. Similar feature vectors are grouped together so that they are located in the same bin of a hash table. Note that, global model description represents a mapping from the feature space to the model.\\
\indent A set of point pair features are computed from the input test image $\mathbf{I}$: A reference point $\mathbf{s}_{ref}$ is firstly selected, and then all other points $\{ \mathbf{s}_i \}$ available in $\mathbf{I}$ are paired with the reference point. Created point pairs $\{(\mathbf{s}_{ref}, \mathbf{s}_i)\}$ are compared with the ones stored in the global model representation. This comparison is employed in feature space, and a set of potential matches are resulted \cite{14, 16, 17}. The rotation angle $\alpha$ between matched point pairs $(\mathbf{m}_r, \mathbf{m}_i)$ and $(\mathbf{s}_{ref}, \mathbf{s}_i)$ is calculated as follows:
\begin{equation}
\small
\mathbf{s}_i = \mathbf{T_s}^{-1} \mathbf{R_x}(\alpha) \mathbf{T_m} \mathbf{m}_i
\end{equation}
where $\mathbf{T_s}$ and $\mathbf{T_m}$ are the transformation matrices that translate $\mathbf{s}_{ref}$ and $\mathbf{m}_r$ into the origin and rotates their normals $\mathbf{n_r^s}$ and $\mathbf{n_r^m}$ onto the \textbf{x}-axis \cite{17}. Once the rotation angle $\alpha$ is calculated, the local coordinate $(\mathbf{m}_r, \alpha)$ is voted \cite{14, 16, 17, 21, 22}.\\
\indent Initial attempts on improving the performance of point-pair matching methods have been on deriving novel features \cite{15,18,19,20,21,144}. \cite{15} augments the descriptors with the visibility context such as dimension, convexity, concativity to discard false matches in range data. \cite{18}, only relying on several scanned views of a target object, exploits color information for PPFs. Boundary points with directions and boundary line segments are also utilized \cite{19}. Point-pair descriptors from both intensity and range are used along with geometric edge extractor to optimize surface and silhouette overlap of the scene and the model \cite{20}.\\
\indent PPF matching methods show underperformance due to similar-looking distractors, occlusion, clutter, large planar surfaces, and sensor noise \cite{b211}. Their performance is further elevated by novel sampling and voting strategies \cite{22,143,43}. The weighted voting function learnt in the max-margin learning framework \cite{22} selects and ranks discriminative features by optimizing a discriminative cost function. \cite{43} exploits object size for sampling to reduce the run-time and handles the sensor noise when voting using the neighboring bins of the lookup table. The run-time performance is also addressed in \cite{145} paralleling the voting process. Recently presented deep architectures learn local descriptors from raw point clouds, which are encoded with PPFs \cite{b213} or additionally exploiting normal representations \cite{b214}.
\begin{table*}[t]
\fontsize{8}{8.2}\selectfont
\caption{Datasets utilized to test the reviewed $3$D BB detectors ({\small I/O: \textbf{I}ndoor/\textbf{O}utdoor}, {\small Mo/St: \textbf{Mo}no/\textbf{St}ereo}).}
\vspace{0.8em}
\centering
\setlength\tabcolsep{4pt}
\renewcommand{\arraystretch}{1.5}
\begin{minipage}{\textwidth}
\centering
\resizebox{0.79\columnwidth}{!}{
\begin{tabular}{ l c c c c c c c c}
    \arrayrulecolor{black}\toprule
    {\small\textbf{Dataset}} &{\small\textbf{Modality}} &{\small\textbf{\# Images}} &{\small\textbf{\# Training}} &{\small\textbf{\# Test}} &{\small\textbf{\# Obj. Classes}} &{\small\textbf{\# annotated obj.}} &{\small\textbf{I/O}} & {\small\textbf{Level}}\\ 
    \arrayrulecolor{black}\midrule
    
    {\small KITTI \cite{126}}       &{\small RGB(Mo/St), LIDAR}  &{\small$14999$}     &{\small$7481$}   &{\small$7518$}   &{\small $3$}      &{\small$80256$}     &{\small O}   &Category\\
    
    {\small SUN RGB-D v1 \cite{128}}   &{\small RGB-D}               &{\small$10335$}     &{\small $5285$}  &{\small $5050$}  &{\small $800$}    &{\small$64595$}     &{\small I} &Category\\
    
    {\small SUN RGB-D v2 \cite{128}}   &{\small RGB-D}               &{\small$13195$}     &{\small$10335$}  &{\small$2860$}   &{\small $19$}     &{\small$67667$}     &{\small I} &Category\\

    {\small NYU-Depth v2 \cite{127}}      &{\small RGB-D}               &{\small$1449$}      &{\small$795$}    &{\small$654$}    &{\small $894$}    &{\small $35064$}    &{\small I} &Category\\

    {\small PASCAL$3$D+ \cite{b204}}  &{\small RGB}                &{\small$30899$}     &{\small \xmark}      &{\small \xmark}      &{\small $12$}     &{\small~$30899$}    &{\small I \& O} &Category\\
    \arrayrulecolor{black}\bottomrule
  \end{tabular}
}
\label{table_$3$D_datasets} 
\end{minipage}%
\end{table*}
\begin{table*}[t]
\fontsize{8}{8.2}\selectfont
\caption{Datasets used to test the reviewed $6$D pose estimators ({\small C: \textbf{C}lutter}, {\small MI: \textbf{M}ultiple \textbf{I}nstance}, {\small O: \textbf{O}cclusion}, {\small SC: \textbf{S}evere \textbf{C}lutter}, {\small SO: \textbf{S}evere \textbf{O}cclusion}, {\small VP: \textbf{V}iew\textbf{P}oint}, {\small SLD: \textbf{S}imilar-\textbf{L}ooking \textbf{D}istractors}, {\small BP: \textbf{B}in \textbf{P}icking}, {\small LI: \textbf{LI}ght}).}
\vspace{0.8em}
\centering
\setlength\tabcolsep{7pt}
\renewcommand{\arraystretch}{1.3}
\begin{minipage}{\textwidth}
\centering
\resizebox{0.79\columnwidth}{!}{
\begin{tabular}{ l c c c c c}
    \arrayrulecolor{black}\toprule
{\small\textbf{Dataset}}    & {\small\textbf{Challenge}}                & {\small\textbf{\# Objects}} & {\small\textbf{Modality}} & {\small\textbf{\# Total Frame}}  & {\small\textbf{Level}}\\ 
    \arrayrulecolor{black}\midrule
{\small RU-APC \cite{b194}} & {\small VP \& C }                         &{\small$14$}                 &{\small RGB-D}             &{\small5964}                      &Instance\\
    
{\small LINEMOD \cite{2}}   & {\small VP \& C \& TL }                   &{\small$15$}                 &{\small RGB-D}             &{\small18273}                     &Instance\\
    
{\small MULT-I \cite{4}}    & {\small VP \& C \& TL \& O \& MI }        &{\small$6$}                  &{\small RGB-D}             &{\small2067}                      &Instance\\
    
{\small OCC \cite{28}}      & {\small VP \& C \& TL \& SO }             &{\small$8$}                  &{\small RGB-D}             &{\small8916}                      &Instance\\

{\small BIN-P \cite{35}}    & {\small VP \& SC \& SO \& MI \& BP }      &{\small$2$}                  &{\small RGB-D}             &{\small177}                       &Instance\\
    
{\small T-LESS \cite{42}}   & {\small VP \& C \& TL \& O \& MI \& SLD } &{\small$30$}                 &{\small RGB-D}             &{\small10080}                     &Instance\\
    
{\small TUD-L}              & {\small VP \& LI }                        &{\small$3$}                  &{\small RGB-D}             &{\small23914}                     &Instance\\
    
{\small TYO-L}              & {\small VP \& C \& LI }                   &{\small$21$}                 &{\small RGB-D}             &{\small1680}                      &Instance\\
    
    \arrayrulecolor{black}\bottomrule
  \end{tabular}
}
\label{table_1} 
\end{minipage}%
\end{table*}
\section{Datasets and Metrics}
\label{ch4_datasets}
In this section, we present the datasets most widely used to test the $3$D BB detectors and full $6$D pose estimators reviewed in this paper and detail the metrics utilized to measure the methods' performance. 
\subsection{Datasets}
\begin{figure}[!t]
\centering
\includegraphics[height=3.4in]{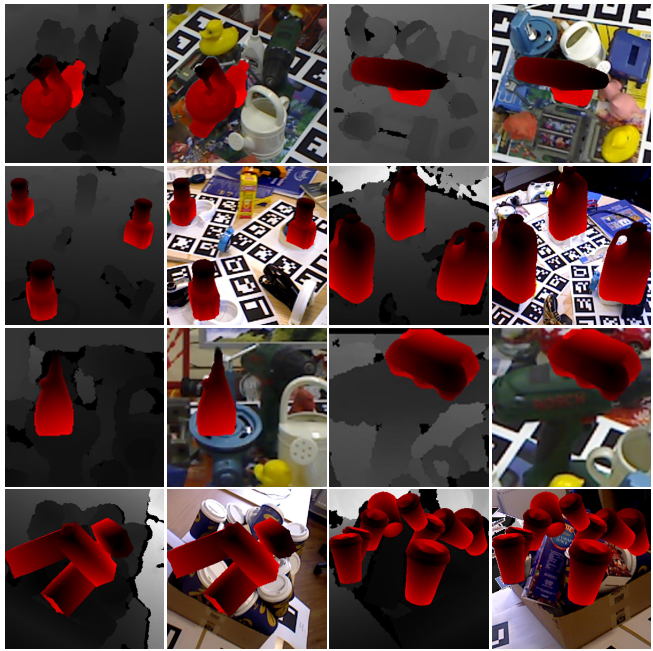}
\caption{Sample datasets for $6$D object pose recovery. Row-wise, the $1^{st}$ benchmark concerns texture-less objects at varying viewpoint with cluttered background, the $2^{nd}$ is interested in multi-instance, the $3^{rd}$ has scenes with severely occluded objects, the $4^{th}$ reflects the challenges found in bin-picking scenarios.}
\label{fig5}
\end{figure}

KITTI \cite{126}, SUN RGB-D \cite{128}, NYU-Depth v2 \cite{127}, and PASCAL$3$D+ \cite{b204} are the datasets on which the $3$D BB detectors are commonly tested (see Table \ref{table_$3$D_datasets}). KITTI has $14999$ images, $7481$ of which are used to train the detectors, and the remaining is utilized to test. It has $3$ discriminative categories, \textit{car, pedestrian}, and \textit{cyclist}, with a total number of $80256$ labeled objects. SUN RGB-D \cite{128} has $10335$ RGB-D images in total, $3784$ and $1159$ of which are captured by Kinect v2 and IntelRealSense, respectively, while the remaining are collected from the datasets of NYU-Depth v2 \cite{127}, B$3$DO \cite{b195}, and SUN$3$D \cite{b196}. For the $10,335$ RGB-D images, there are $64,595$ $3$D BB annotations with accurate orientations for about $800$ object categories ($2^{nd}$ row of Table \ref{table_$3$D_datasets}). There are around $~5K$ training and $~5K$ test images in the dataset for $37$ object categories\footnote{This statistics is obtained from \cite{b197}.}. In the $2017$ SUN RGB-D $3$D object detection challenge\footnote{http://rgbd.cs.princeton.edu/challenge.html}, existing $10335$ images along with their $64595$ $3$D BB annotations are used as training data, and newly acquired $2860$ images, which have $3072$ annotated object BBs, are utilized to test the methods' performance on $19$ object classes ($3^{rd}$ row of Table \ref{table_$3$D_datasets}). The NYU-Depth v2 dataset \cite{127} involves $1449$ images, $795$ of which are of training, and the rest is for the test. Despite the fact that it has $894$ object categories, there are mainly $19$ object categories on which the methods are evaluated \textit{(bathtub, bed, bookshelf, box, chair, counter, desk, door, dresser, garbage bin, lamp, monitor, nightstand, pillow, sink, sofa, table, tv, toilet)}. $12$ object categories of PASCAL VOC $2012$ \cite{7} \textit{(aeroplane, bicycle, boat, bottle, bus, car, chair, dining table, motorbike, sofa, train, and tvmonitor)} are augmented with $3$D annotations in PASCAL$3$D+ \cite{b204}. For each category, more images are added from ImageNet \cite{6}, resulting in a total number of $30899$ images with annotated objects. Apart from the datasets detailed in Table \ref{table_$3$D_datasets}, we briefly mention other related $3$D object detection datasets \cite{b198, b199, b200, b201, b202, b203}. The RGB-D object dataset \cite{b198} involves $51$ object classes, including $300$ geometrically different object instances with clean background. The dataset in \cite{b199} provides indoor images with $3$D annotations for several IKEA objects. NYC$3$DCars \cite{b200} and EPFL Cars \cite{b201} are presented for the \textit{car} category. The images of the former are captured in the street scenes of New York City, and the latter one includes $2299$ images of $20$ car instances, which are taken at different azimuth angles but similar elevation and close distances. Table-Top Pose dataset \cite{b202} is formed of $480$ images of $3$ categories, \textit{mouse, mug}, and \textit{stapler}, and the one in \cite{b203} annotates the subsets of $4$ categories of ImageNet, \textit{bed, chair, sofa}, and \textit{table}, with $3$D BBs.\\
\indent LINEMOD \cite{2}, MULT-I \cite{4}, OCC \cite{28}, BIN-P \cite{35}, and T-LESS \cite{42} are the datasets most frequently used to test the performances of full $6$D pose estimators. In a recently proposed benchmark for $6$D object pose estimation \cite{5}, these datasets are refined and are presented in a unified format along with three new datasets (Rutgers Amazon Picking Challenge (RU-APC) \cite{b194}, TU Dresden Light (TUD-L), and Toyota Light (TYO-L)). The table which details the parameters of these datasets (Table $1$, pp. $25$, in \cite{5}) is reformatted in our paper in Table \ref{table_1} to present the datasets' challenges:\\
\noindent \textbf{Viewpoint (VP) \& Clutter (C).} RU-APC \cite{b194} dataset involves the test scenes in which objects of interest are located at \textit{varying viewpoints} and \textit{cluttered backgrounds}.\\
\noindent \textbf{VP \& C \& Texture-less (TL).} Test scenes in the LINEMOD \cite{2} dataset involve \textit{texture-less} objects at varying viewpoints with cluttered backgrounds. There are $15$ objects, for each of which more than $1100$ real images are recorded. The sequences provide views from $0-360$ degree around the object, $0-90$ degree tilt rotation, $\mp 45$ degree in-plane rotation, and $650-1150$ mm object distance.\\
\noindent \textbf{VP \& C \& TL \& Occlusion (O) \& Multiple Instance (MI).} Occlusion is one of the main challenges that makes the datasets more difficult for the task of object detection and $6$D pose estimation. In addition to close and far range $2$D and $3$D clutter, testing sequences of the Multiple-Instance (MULT-I) dataset \cite{4} contain \textit{foreground occlusions} and \textit{multiple object instances}. In total, there are approximately $2000$ real images of $6$ different objects, which are located at the range of $600-1200$ mm. The testing images are sampled to produce sequences that are uniformly distributed in the pose space by $[0^\circ, 360^\circ ]$, $[-80^\circ, 80^\circ ]$, and $[-70^\circ, 70^\circ ]$ in the yaw, roll, and pitch angles, respectively.\\
\noindent \textbf{VP \& C \& TL \& Severe Occlusion (SO).} Occlusion, clutter, texture-less objects, and change in viewpoint are the most well-known challenges that could successfully be dealt with the state-of-the-art $6$D object detectors. However, \textit{heavy existence} of these challenges severely degrades the performance of $6$D object detectors. Occlusion (OCC) dataset \cite{28} is one of the most difficult datasets in which one can observe up to  $70-80 \%$ occluded objects. OCC includes the extended ground truth annotations of LINEMOD: in each test scene of the LINEMOD \cite{2} dataset, various objects are present, but only ground truth poses for one object are given. Brachmann et al. \cite{28} form OCC considering the images of one scene (benchvise) and annotating the poses of $8$ additional objects.\\
\noindent \textbf{VP \& SC \& SO \& MI \& Bin Picking (BP).} In \textit{bin-picking} scenarios, multiple instances of the objects of interest are arbitrarily stocked in a bin, and hence, the objects are inherently subjected to severe occlusion and severe clutter. Bin-Picking (BIN-P) dataset \cite{35} is created to reflect such challenges found in industrial settings. It includes $177$ test images of $2$ textured objects under varying viewpoints.\\
\noindent \textbf{VP \& C \& TL \& O \& MI \& Similar-Looking Distractors (SLD).} \textit{Similar-looking distractor(s)} along with similar-looking object classes involved in the datasets strongly confuse recognition systems causing a lack of discriminative selection of shape features. Unlike the above-mentioned datasets and their corresponding challenges, the T-LESS \cite{42} dataset particularly focuses on this problem. The RGB-D images of the objects located on a table are captured at different viewpoints covering $360$ degrees rotation, and various object arrangements generate occlusion. Out-of-training objects, similar-looking distractors (planar surfaces), and similar-looking objects cause $6$ DoF methods to produce many false positives, particularly affecting the depth modality features. T-LESS has $30$ texture-less industry-relevant objects, and $20$ different test scenes, each of which consists of $504$ test images.\\
\noindent \textbf{VP \& Light (LI).} TUD-L is a dedicated dataset for measuring the performances of the $6$D pose estimators across different ambient and \textit{lighting} conditions. The objects of interest are subjected to $8$ different lighting conditions in the test images, in the first frame of which object pose is manually aligned to the scene using $3$D object model. The initial pose is then propagated through the sequence using ICP.\\
\noindent \textbf{VP \& C \& LI.} As in the TUD-L dataset, different lighting condition is also featured in TYO-L dataset, where the target objects are located in a cluttered background. Its $21$ objects are captured on a table-top setup under $5$ different illumination.\\
\indent We also briefly mention about new datasets which are utilized on measuring the performance of full $6$D pose estimators. YCB-Video dataset \cite{39} is formed of $92$ video sequences, $80$ of which are used for training, while the remaining sequences are of the testing phase. The total number of real images are about $133$K, and it additionally contains around $80$K synthetically rendered images to be utilized for training. The test images, where $21$ objects taken from the YCB dataset \cite{b205, b206} exist, are subjected to significant image noise, varying lighting conditions, and occlusion in cluttered background. JHUScene-50 \cite{b207} dataset contains $5$ different indoor environments including office workspaces, robot manipulation platforms, and large containers. Each environment contains $10$ scenes, each of which is composed of $100$ test frames with densely cluttered multiple object instances. There are $10$ hand tool objects, $5000$ test images, and $22520$ labeled poses in total Sample datasets for $6$D object pose recovery are shown in Fig. \ref{fig5}.
\subsection{Metrics: Correctness of an Estimation}
Several evaluation metrics have been proposed to determine the correctness of $3$D BB estimations and full $6$D pose hypotheses.
\subsubsection{Metrics for 3D Bounding Boxes} We firstly present the metrics for $3$D BB detectors.\\
\noindent \textbf{Intersection over Union (IoU) \cite{7}.} This metric is originally presented to evaluate the performance of the methods working in $2$D space \cite{7}. Given the estimated and ground truth BBs $B$ and $\bar{B}$ and assuming that they are aligned with image axes, it determines the area of intersection $B \cap \bar{B}$, and the area of union $B \cup \bar{B}$, and then comparing these two, outputs the overlapping ratio $\omega_\mathrm{IoU_{2D}}$:
\begin{equation}
\small
\omega_\mathrm{IoU_{2D}} = \frac{B \cap \bar{B}}{B \cup \bar{B}}
\label{eq9}
\end{equation}
According to Eq. \ref{eq9}, a predicted box is considered to be correct (true positive) if the overlapping ratio $\omega_\mathrm{IoU_{2D}}$ is more than the threshold $\tau_\mathrm{IoU_{2D}} = 0.5$ \cite{7}. This metric is further extended to work with $3$D volumes calculating overlapping ratio $\omega_\mathrm{IoU_{3D}}$ over $3$D BBs \cite{93}. The extended version assumes that $3$D BBs are aligned with gravity direction, but makes no assumption on the other two axes.\\
\noindent \textbf{Average Precision (AP) \cite{7}.} It depicts the shape of the Precision/Recall (PR) curve. Dividing the recall range  $ r = [0 \quad 1]$ into a set of $11$ equal levels, it finds the mean precision $p$ at this set \cite{7}:
\begin{equation}
\small
\mathrm{AP} = \frac{1}{11} \sum_{r \in \{ 0,0.1,..,1 \}} p_{\mathrm{interp}}(r)
\label{eq10}
\end{equation}
The precision $p$ at each recall level $r$ is interpolated by taking the maximum precision measured for a method for which the corresponding recall exceeds $r$:
\begin{equation}
\small
p_{\mathrm{interp}}(r) = \underset{\tilde{r}:\tilde{r} \geq r}{\mathrm{max}} p(\tilde{r})
\label{eq10a}
\end{equation}
where $p(\tilde{r})$ is measured precision at recall $\tilde{r}$.\\
\noindent \textbf{Average Orientation Similarity (AOS) \cite{126}.} This metric firstly measures the object detection performance of any detector in $2$D image plane using the AP metric in \cite{7}, and then, evaluates the joint detection and $3$D orientation estimation performance as follows \cite{126}:
\begin{equation}
\small
\mathrm{AOS} = \frac{1}{11} \sum_{r \in \{ 0,0.1,..,1 \}} \underset{\tilde{r}:\tilde{r} \geq r}{\mathrm{max}} sim(\tilde{r}),
\label{eqnewww}
\end{equation}
where, the orientation similarity at recall $r$, $sim(r)$, is a normalized $([0...1])$ version of the cosine similarity given below:
\begin{equation}
\small
sim(r) = \frac{1}{\vert \mathcal{D}(r) \vert} \sum_{i \in \mathcal{D}(r)} \frac{1+\cos \Delta_{\theta}^{(i)}}{2} \delta_{i}.
\label{eqcosine}
\end{equation}
In Eq. \ref{eqcosine}, $\mathcal{D}(r)$ depicts the set of all object detections at recall level $r$, $\Delta_{\theta}^{(i)}$ is the difference in angle between estimated
and ground truth orientation of detection $i$. $\delta_{i}$ is set to $1$ if detection $i$ is assigned to a ground truth BB (overlaps by at least $50 \%$), and $\delta_{i} = 0$ if it has not been assigned. This penalizes multiple detections which explain a single object.\\
\noindent \textbf{Average Viewpoint Precision (AVP) \cite{b204}.} As in \cite{126}, this metric measures the performance of any detector jointly considering $2$D object detection and $3$D orientation. An estimation is accepted as correct if and only if the BB overlap is larger than $50 \%$ and the viewpoint is correct (\textit{i.e.}, the two viewpoint labels are the same in discrete viewpoint space or the distance between the two viewpoints is smaller than some threshold in continuous viewpoint space). Then, a Viewpoint Precision-Recall (VPR) curve, the area under of which is equal to average viewpoint precision, is drawn.
\subsubsection{Metrics for Full 6D Pose} Once we detail the metrics for $3$D BB detectors, we next present the metrics for the methods of full $6$D pose estimators.\\
\noindent \textbf{Translational and Angular Error.} As being independent from the models of objects, this metric measures the correctness of a hypothesis according to the followings \cite{17}: i) $\mathcal{L}_2$ norm between the ground truth and estimated translations $\bar{\mathbf{x}}$ and $\mathbf{x}$, ii) the angle computed from the axis-angle representation of ground truth and estimated rotation matrices $\bar{\mathbf{R}}$ and $\mathbf{R}$:
\begin{equation}
\small
\omega_\mathrm{TE} = \vert \vert (\bar{\mathbf{x}} - \mathbf{x}) \vert \vert_2,
\label{eq7}
\end{equation}
\vspace{-1em}
\begin{equation}
\small
\omega_\mathrm{RE} = \arccos(\Tr(\mathbf{R} \bar{\mathbf{R}}^{-1} - 1)/2).
\label{eq8}
\end{equation}
According to Eqs. \ref{eq7} and \ref{eq8}, a hypothesis is accepted as correct if the scores $\omega_\mathrm{TE}$ and $\omega_\mathrm{RE}$ are below predefined thresholds $\tau_\mathrm{TE}$ and $\tau_\mathrm{RE}$.\\
\noindent \textbf{$2$D Projection.} This metric takes the $3$D model of an object of interest as input and projects the model's vertices $\mathbf{v}$ onto the image plane at both ground truth and estimated poses. The distance of the projections of corresponding vertices is determined as follows:
\begin{equation}
\small
\omega_\mathrm{2Dproj} = \frac{1}{\vert \mathcal{V} \vert} \sum_{\mathbf{v} \in \mathcal{V}} \vert \vert \mathbf{K}  \bar{\mathbf{R}} \mathbf{v} - \mathbf{K} \mathbf{R} \mathbf{v} \vert \vert_{2}
\label{eq10b}
\end{equation}
where $\mathcal{V}$ is the set of all object model vertices, $\mathbf{K}$ is the camera matrix. Homogenous coordinates are normalized before calculating the $\mathcal{L}_2$ norm. An estimated pose is accepted as correct if the average re-projection error $\omega_\mathrm{2Dproj}$ is below $5$px. As this error is basically computed using the $3$D model of an object of interest, it can also be calculated using its point cloud.\\
\noindent \textbf{Average Distance (AD).} This is one of the most widely used metrics in the literature \cite{2}. Given the ground truth $(\bar{\mathbf{x}}, \bar{\mathbf{\Theta}})$ and estimated $(\mathbf{x}, \mathbf{\Theta})$ poses of an object of interest $O$, this metric outputs $\omega_\mathrm{AD}$, the score of the AD between $(\bar{\mathbf{x}}, \bar{\mathbf{\Theta}})$ and $(\mathbf{x}, \mathbf{\Theta})$. It is calculated over all points $\mathbf{s}$ of the $3$D model $M$ of the object of interest:
\begin{equation}
\small
\omega_\mathrm{AD} = \underset{\mathbf{s} \in M}{\mathrm{avg}} \vert \vert (\bar{\mathbf{R}} \mathbf{s} + \bar{\mathbf{T}}) - (\mathbf{R} \mathbf{s} + \mathbf{T}) \vert \vert
\label{eq3}
\end{equation}
where $\bar{\mathbf{R}}$ and $\bar{\mathbf{T}}$ depict rotation and translation matrices of the ground truth pose $(\bar{\mathbf{x}}, \bar{\mathbf{\Theta}})$, while $\mathbf{R}$ and $\mathbf{T}$ represent rotation and translation matrices of the estimated pose $(\mathbf{x}, \mathbf{\Theta})$. Hypotheses ensuring the following inequality are considered as correct:
\begin{equation}
\small
\omega_\mathrm{AD} \leq z_{\omega} \Phi
\label{eq4}
\end{equation}
where $\Phi$ is the diameter of the $3$D model $M$, and $z_{\omega}$ is a constant that determines the coarseness of a hypothesis which is assigned as correct. Note that, Eq. \ref{eq3} is valid for objects whose models are not ambiguous or do not have any subset of views under which they appear to be ambiguous. In case the model $M$ of an object of interest has indistinguishable views, Eq. \ref{eq3} transforms into the following form:
\begin{equation}
\small
\omega_\mathrm{AD} = \underset{\mathbf{s}_1 \in M}{\mathrm{avg}}\underset{\mathbf{s}_2 \in M}{\mathrm{min}} \vert \vert ( \bar{\mathbf{R}} \mathbf{s}_1 + \bar{\mathbf{T}}) - ( \mathbf{R} \mathbf{s}_2 + \mathbf{T}) \vert \vert
\label{eq5}
\end{equation}
where $\omega_\mathrm{AD}$ is calculated as the AD to the closest model point. This function employs many-to-one point matching and significantly promotes symmetric and occluded objects, generating lower $\omega_\mathrm{AD}$ scores.\\
\noindent \textbf{Visible Surface Discrepancy (VSD).} This metric has recently been proposed to eliminate ambiguities arising from object symmetries and occlusions \cite{3}. The model $M$ of an object of interest $O$ is rendered at both ground truth $(\bar{\mathbf{x}}, \bar{\mathbf{\Theta}})$ and estimated $(\mathbf{x}, \mathbf{\Theta})$ poses, and the depth maps $\bar{\mathbf{D}}$ and $\mathbf{D}$ of the renderings are intersected with the test depth image itself $\mathbf{I_D}$ in order to generate the visibility masks $\mathbf{\bar{V}}$ and $\mathbf{V}$. By comparing the generated masks, $\omega_\mathrm{VSD}$, the score that determines whether an estimation is correct, according to a pre-defined threshold $\tau$, is computed as follows:
\begin{equation}
\small
\omega_\mathrm{VSD} =
\underset{q \in \bar{\mathbf{V}} \cup \mathbf{V} }{\mathrm{avg}}
\begin{cases} 
0 & \text{if $q \in \bar{\mathbf{V}} \cap \mathbf{V} \, \wedge \, |\bar{\mathbf{D}}(q) - \mathbf{D} (q)| < \tau$} \\
1 & \text{otherwise}
\end{cases}
\end{equation}
where $q$ depicts pixel. As formulized, the score $\omega_\mathrm{VSD}$ is calculated only over the visible part of the model surface, and thus the indistinguishable poses are treated as equivalent.\\
\noindent \textbf{\textit{Sym} Pose Distance.} VSD \cite{3} is presented as an ambiguity-invariant pose error function. The acceptance of a pose hypothesis as correct depends on the plausibility of the estimated pose given the available data. However, numerous applications, particularly in the context of active vision, rely on a precise pose estimation and cannot be satisfied by plausible hypotheses, and hence VSD is considered problematic for evaluation. When objects of interest are highly occluded, the number of plausible hypotheses can be infinitely large. \textit{Sym} pose distance deals with the ambiguities arising from the symmetry issue under a theoretically backed framework. Within this framework, this metric outputs the score $\omega_{\textit{Sym}}$ between poses $\mathcal{P}_1$ and $\mathcal{P}_2$:
\begin{equation}
\small
\omega_{\textit{Sym}} = \underset{\mathbf{G_1, G_2} \in G}{\mathrm{min}} \sqrt{ \frac{1}{A} \int_{A} \vert \vert \mathbf{W_2} \circ \mathbf{G_2(x)} - \mathbf{W_1} \circ \mathbf{G_1(x)} \vert \vert^{2} da},
\label{eq11}
\end{equation}
where the pose $\mathcal{P}_1$ is identified to the equivalency class $\{ \mathbf{W_1} \circ \mathbf{G_1 (x)} | \mathbf{G_1} \in G \}$, and the pose $\mathcal{P}_2$ is identified to the equivalency class $\{ \mathbf{W_2} \circ \mathbf{G_2 (x)} | \mathbf{G_2} \in G \}$. $A$ is the surface area of the object of interest, $\mathbf{W_1}$ and $\mathbf{W_2}$ are rigid transformations, and $\mathbf{G_1}$ and $\mathbf{G_2}$ are the symmetry groups in $G \in SE(3)$.\\

\noindent \textbf{Implementation Details.} In this study, we employ a twofold evaluation strategy for the $6$D detectors using both AD and VSD metrics: i) Recall. The hypotheses on the test images of every object are ranked, and the hypothesis with the highest weight is selected as the estimated $6$D pose. Recall value is calculated comparing the number of correctly estimated poses and the number of the test images of the interested object. ii) F1 scores. Unlike recall, all hypotheses are taken into account, and F1 score, the harmonic mean of precision and recall values, is presented.
\section{Multi-modal Analyses}
\label{ch4_Exp_Res}
We analyze $17$ baselines on the datasets with respect to both challenges and the architectures. Two of the baselines \cite{2, 4} are our own implementations. The color gradients and surface normal features, presented in \cite{2}, are computed using the built-in functions and classes provided by OpenCV. The features in LCHF \cite{4} are the part-based version of the features introduced in \cite{2}. Hence, we inherit the classes given by OpenCV in order to generate part-based features used in LCHF. We train each method for the objects of interest by ourselves, and using the learnt detectors, we test those on all datasets. Note that, the methods use only foreground samples during training/template generation. In this section, \enquote{LINEMOD} refers to the dataset, whilst \enquote{Linemod} is used to indicate the baseline itself.
\subsection{Analyses Based on Average Distance}
Utilizing the AD metric, we compare the chosen baselines along with the challenges, i) regarding the recall values that each baseline generates on every dataset, ii) regarding the F1 scores. The coefficient $z_{\omega}$ is $0.10$, and in case we use different thresholds, we will specifically indicate in the related parts.
\begin{table*}[!t]
\small
\fontsize{8}{8.2}\selectfont
\caption{Methods' performance are depicted object-wise based on recall values computed using the Average Distance (AD) evaluation protocol.}
\centering

\setlength\tabcolsep{5pt}
{\renewcommand{\arraystretch}{1.3}
\begin{subtable}{\linewidth}\centering
{\begin{tabular}[t]{l c c c c c c c c c c c c c c c}
\hline
\textbf{Method} &\textbf{input} &\textbf{ape} &\textbf{bvise} &\textbf{cam} &\textbf{can} &\textbf{cat} &\textbf{driller} &\textbf{duck} &\textbf{box} &\textbf{glue} &\textbf{hpuncher} &\textbf{iron} &\textbf{lamp} &\textbf{phone} &\textbf{AVER}\\  
\hline
   Brach et al. \cite{30}    &RGB-D  &98.1   &99.0   &99.7  &99.7   &99.1   &100   &96.2   &99.7   &99.0    &98.0   &99.9   &99.5   &99.6  &99.0\\
    Kehl et al. \cite{36}    &RGB-D  &96.9   &94.1   &97.7  &95.2   &97.4   &96.2  &97.3   &99.9   &78.6    &96.8   &98.7   &96.2   &92.8  &95.2\\
   Brach et al. \cite{28}    &RGB-D  &85.4   &98.9   &92.1  &84.4   &90.6   &99.7  &92.7   &91.1   &87.9    &97.9   &98.8   &97.6   &86.1  &92.6\\
           LCHF \cite{4}     &RGB-D  &84.0	 &95.0   &72.0  &74.0   &91.0   &92.0  &91.0   &48.0   &55.0    &89.0   &72.0   &90.0   &69.0  &78.6\\
  Cabrera et al. \cite{24}   &RGB-D  &95.0	 &98.9   &98.2  &96.3   &99.1   &94.3  &94.2   &99.8   &96.3    &97.5   &98.4   &97.9   &88.3  &96.5\\
         Linemod \cite{2}    &RGB-D  &95.8   &98.7   &97.5  &95.4   &99.3   &93.6  &95.9   &99.8   &91.8    &95.9   &97.5   &97.7   &93.3  &96.3\\
    Hodan et al. \cite{25}   &RGB-D  &93.9   &99.8   &95.5  &95.9   &98.2   &94.1  &94.3   &100    &98.0    &88.0   &97     &88.8   &89.4  &94.9\\
         Hashmod \cite{44}   &RGB-D  &95.6   &91.2   &95.2  &91.8   &96.1   &95.1  &92.9   &99.9   &95.4    &95.9   &94.3   &94.9   &91.3  &94.6\\
  Hinters et al. \cite{43}   &RGB-D  &98.5   &99.8   &99.3  &98.7   &99.9   &93.4  &98.2   &98.8   &75.4    &98.1   &98.3   &96.0   &98.6  &96.4\\
    Drost et al. \cite{17}   &Depth  &86.5   &70.7   &78.6  &80.2   &85.4   &87.3  &46.0   &97.0   &57.2    &77.4   &84.9   &93.3   &80.7  &78.9\\
           SSD-6D \cite{37}  &RGB    &65.0   &80.0   &78.0  &86.0   &70.0   &73.0  &66.0   &100    &100     &49.0   &78.0   &73.0   &79.0  &76.7\\
              BB8 \cite{40}  &RGB    &40.4   &91.8   &55.7  &64.1   &62.6   &74.4  &44.3   &57.8   &41.2    &67.2   &84.7   &76.5   &54.0  &62.7\\
     Tekin et al. \cite{38}  &RGB    &21.6   &81.8   &36.6  &68.8   &41.8   &63.5  &27.2   &69.6   &80.0    &42.6   &75.0   &71.1   &47.7  &56.0\\
     Brach et al. \cite{30}  &RGB    &33.2   &64.8   &38.4  &62.9   &42.7   &61.9  &30.2   &49.9   &31.2    &52.8   &80.0   &67.0   &38.1  &50.3\\ 
\hline
\end{tabular}}
\vspace{0.5em}
\caption{LINEMOD}\label{table_2a}
\end{subtable}%
}

\setlength\tabcolsep{5pt}
{\renewcommand{\arraystretch}{1.3}
\begin{subtable}{\linewidth}\centering
{\begin{tabular}[t]{l c c c c c c c c}
\hline
\textbf{Method}           &\textbf{input}   &\textbf{camera}  &\textbf{cup}  &\textbf{joystick}   &\textbf{juice}   &\textbf{milk}   &\textbf{shampoo} & \textbf{AVER}\\
\hline
   LCHF \cite{4}  &RGB-D &52.5      &99.8   &98.3         &99.3    &92.7   &97.2   &90.0\\
 Linemod \cite{2} &RGB-D &18.3      &99.2   &85.0         &51.6    &72.2   &53.1   &63.2\\
\hline
\end{tabular}}
\vspace{0.5em}
\caption{MULT-I}\label{table_2b}
\end{subtable}%
}

\setlength\tabcolsep{5pt}
{\renewcommand{\arraystretch}{1.3}
\begin{subtable}{\linewidth}\centering
{\begin{tabular}[t]{l c c c c c c c c c c}
\hline
\textbf{Method} &\textbf{input} &\textbf{ape} &\textbf{can} &\textbf{cat} &\textbf{driller} &\textbf{duck} &\textbf{box} &\textbf{glue} &\textbf{hpuncher} &\textbf{AVER}\\ 
\hline
         PoseCNN \cite{39}   &RGB-D    &76.2	 &87.4  &52.2   &90.3   &77.7  &72.2   &76.7   &91.4   &78.0\\
    Michel et al. \cite{33}  &RGB-D    &80.7	 &88.5  &57.8   &94.7   &74.4  &47.6   &73.8   &96.3   &76.8\\
    Krull et al. \cite{29}   &RGB-D    &68.0	 &87.9  &50.6   &91.2   &64.7  &41.5   &65.3   &92.9   &70.3\\
    Brach et al. \cite{28}   &RGB-D    &53.1	 &79.9  &28.2   &82.0   &64.3  &9.0    &44.5   &91.6   &56.6\\
             LCHF \cite{4}   &RGB-D    &48.0	 &79.0  &38.0   &83.0   &64.0  &11.0   &32.0   &69.0   &53.0\\
    Hinters et al. \cite{43} &RGB-D    &81.4	 &94.7  &55.2   &86.0   &79.7  &65.5   &52.1   &95.5   &76.3\\
           Linemod \cite{2}  &RGB-D    &21.0	 &31.0  &14.0   &37.0   &42.0  &21.0   &5.0    &35.0   &25.8\\
         PoseCNN \cite{39}   &RGB      &9.6	     &45.2  &0.93   &41.4   &19.6  &22.0   &38.5   &22.1   &25.0\\
    \hline
\end{tabular}}
\vspace{0.5em}
\caption{OCC}\label{table_2c}
\end{subtable}%
}

\setlength\tabcolsep{5pt}
{\renewcommand{\arraystretch}{1.3}
\begin{subtable}{\linewidth}\centering
{\begin{tabular}[t]{l c c c c}
\hline
\textbf{Method}              &\textbf{input}    &\textbf{cup} &\textbf{juice}  &\textbf{AVER}\\
\hline
	          LCHF \cite{4}  &RGB-D             &90.0	      &89.0            &90.0\\
	  Brach et al. \cite{28} &RGB-D             &89.4         &87.6            &89.0\\
	       Linemod \cite{2}  &RGB-D             &88.0	      &40.0            &64.0\\
\hline
\end{tabular}}
\vspace{0.5em}
\caption{BIN-P}\label{table_$2$D}
\end{subtable}%
}
\label{table_2}
\vspace{-1.5em}
\end{table*}
\begin{figure*}[!t]
\centering
\includegraphics[width=\linewidth]{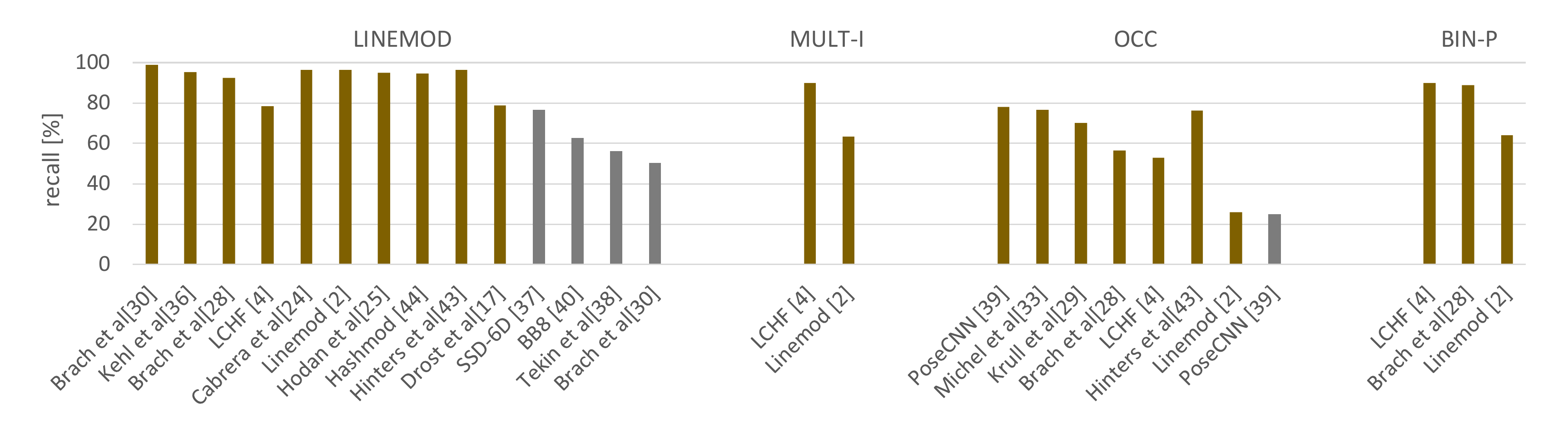}
\caption{Success of each baseline on every dataset is shown, recall values are computed using the Average Distance (AD) metric.}
\label{fig6}
\vspace{-1em}
\end{figure*}
\begin{figure*}[!t]
\centering
\includegraphics[width=\linewidth]{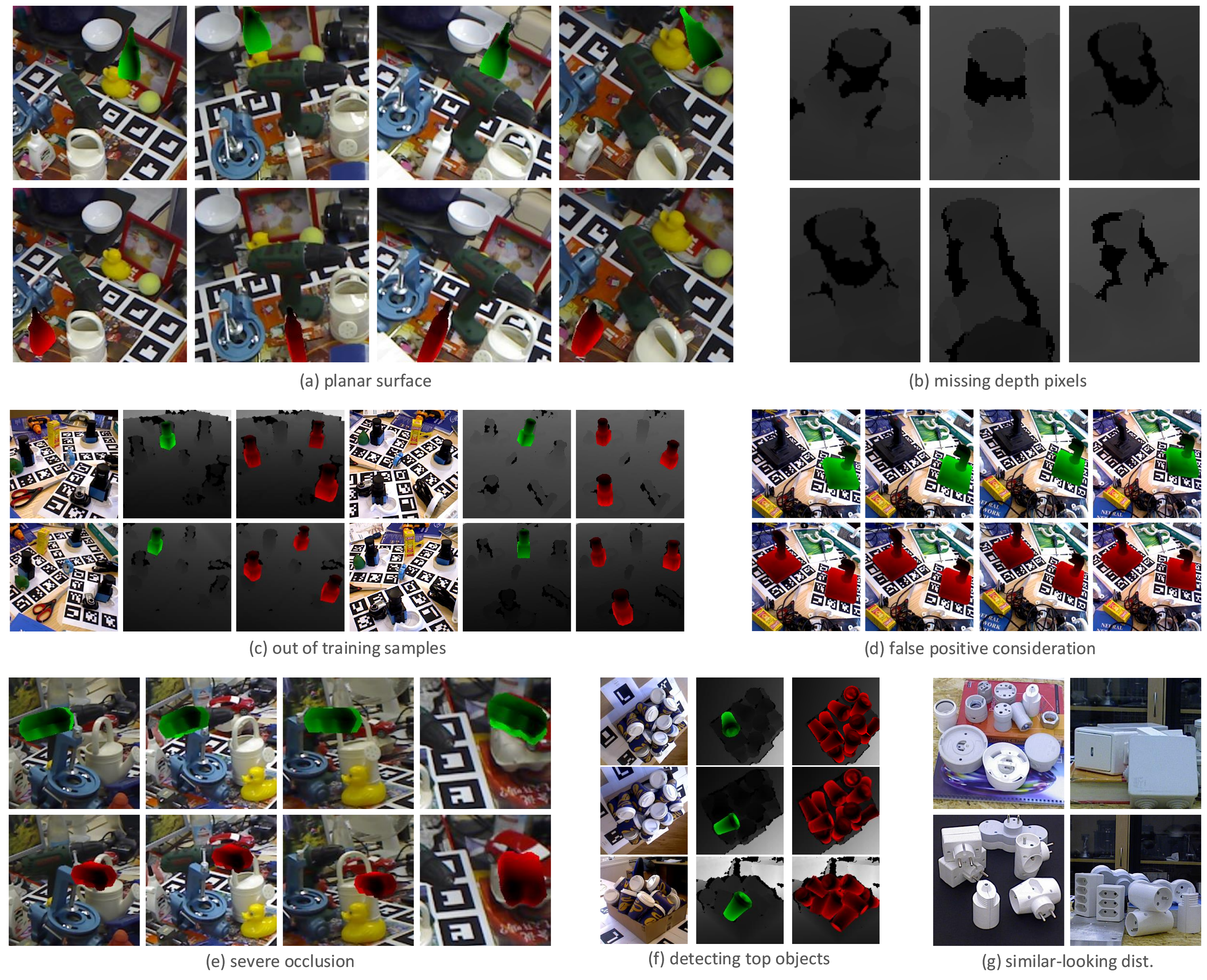}
\caption{Challenges encountered during test are exemplified (green renderings are hypotheses, and the red ones are ground truths). (For interpretation of the references to color in this figure legend, the
reader is referred to the web version of this article.)}
\label{fig7}
\vspace{-1em}
\end{figure*}
\begin{table*}[!t]
\small
\fontsize{8}{8.2}\selectfont
\caption{Methods' performance are depicted object-wise based on F1 scores computed using the Average Distance (AD) evaluation protocol.}
\centering

\setlength\tabcolsep{5pt}
{\renewcommand{\arraystretch}{1.3}
\begin{subtable}{\linewidth}\centering
{\begin{tabular}[t]{l c c c c c c c c c c c c c c c}
\hline
\textbf{Method} &\textbf{input} &\textbf{ape} &\textbf{bvise} &\textbf{cam} &\textbf{can} &\textbf{cat} &\textbf{dril} &\textbf{duck} &\textbf{box} &\textbf{glue} &\textbf{hpuncher} &\textbf{iron} &\textbf{lamp} &\textbf{phone} & \textbf{AVER}\\ 
\hline
     Kehl et al. \cite{36}  &RGB-D &0.98 &0.95 &0.93 &0.83 &0.98 &0.97 &0.98 &1 &0.74 &0.98 &0.91 &0.98 &0.85 &0.93\\
             LCHF \cite{4}  &RGB-D &0.86 &0.96 &0.72 &0.71 &0.89 &0.91 &0.91 &0.74 &0.68 &0.88 &0.74 &0.92 &0.73 &0.82\\
           Linemod \cite{2} &RGB-D  &0.53 &0.85 &0.64 &0.51 &0.66 &0.69 &0.58 &0.86 &0.44 &0.52 &0.68 &0.68 &0.56 &0.63\\
          SSD-6D \cite{37}  &RGB &0.76 &0.97 &0.92 &0.93 &0.89 &0.97 &0.80 &0.94 &0.76 &0.72 &0.98 &0.93 &0.92 &0.88\\
\hline
\end{tabular}}
\vspace{0.5em}
\caption{LINEMOD}\label{table_3a}
\end{subtable}%
}

\setlength\tabcolsep{5pt}
{\renewcommand{\arraystretch}{1.3}
\begin{subtable}{\linewidth}\centering
{\begin{tabular}[t]{l c c c c c c c c}
\hline
\textbf{Method}      &\textbf{input}         &\textbf{camera}  &\textbf{cup}  &\textbf{joystick}  &\textbf{juice}  &\textbf{milk}  &\textbf{shampoo} & \textbf{AVER}\\
\hline
Doum et al. \cite{35}     &RGB-D     &0.93  &0.74  &0.92	 &0.90  &0.82  &0.51 &0.8\\
Kehl et al. \cite{36}     &RGB-D     &0.38  &0.97  &0.89	 &0.87  &0.46  &0.91 &0.75\\
          LCHF \cite{4}   &RGB-D     &0.39  &0.89  &0.55	 &0.88  &0.40  &0.79 &0.65\\
Drost et al. \cite{17}    &RGB-D     &0.41  &0.87  &0.28	 &0.60  &0.26  &0.65 &0.51\\
        Linemod \cite{2}  &RGB-D     &0.37  &0.58  &0.15     &0.44  &0.49  &0.55 &0.43\\	
     SSD-6D \cite{37}     &RGB     &0.74  &0.98  &0.99  &0.92  &0.78  &0.89 &0.88\\
\hline
\end{tabular}}
\vspace{0.5em}
\caption{MULT-I}\label{table_3b}
\end{subtable}%
}

\setlength\tabcolsep{5pt}
{\renewcommand{\arraystretch}{1.3}
\begin{subtable}{\linewidth}\centering
{\begin{tabular}[t]{l c c c c c c c c c c}
\hline
\textbf{Method} &\textbf{input} &\textbf{ape} &\textbf{can} &\textbf{cat} &\textbf{driller} &\textbf{duck} &\textbf{box} &\textbf{glue} &\textbf{hpuncher} & \textbf{AVER}\\ 
\hline
            LCHF \cite{4}   &RGB-D   &0.51	 &0.77     &0.44   &0.82   &0.66     &0.13     &0.25    &0.64  &0.53\\
          Linemod \cite{2}  &RGB-D   &0.23	 &0.31     &0.17   &0.37   &0.43     &0.19     &0.05    &0.30  &0.26\\
  Brach et al. \cite{30}    &RGB     &\xmark &\xmark        &\xmark      &\xmark      &\xmark        &\xmark        &\xmark       &\xmark     &0.51\\
  Tekin et al. \cite{38}    &RGB     &\xmark &\xmark        &\xmark      &\xmark      &\xmark        &\xmark        &\xmark       &\xmark     &0.48\\
   SSD-6D \cite{37}    &RGB     &\xmark &\xmark        &\xmark      &\xmark      &\xmark        &\xmark        &\xmark       &\xmark     &0.38\\
\hline
\end{tabular}}
\vspace{0.5em}
\caption{OCC}\label{table_3c}
\end{subtable}%
}

\setlength\tabcolsep{5pt}
{\renewcommand{\arraystretch}{1.3}
\begin{subtable}{\linewidth}\centering
{\begin{tabular}[t]{l c c c c}
\hline
\textbf{Method}              &\textbf{input}   &\textbf{cup}   &\textbf{juice} &\textbf{AVER}\\
\hline
             LCHF \cite{4}   &RGB-D   &0.48  &0.29  &0.39\\
    Doum et al. \cite{35}    &RGB-D   &0.36  &0.29  &0.33\\
           Linemod \cite{2}  &RGB-D   &0.48  &0.20  &0.34\\
\hline
\end{tabular}}
\vspace{0.5em}
\caption{BIN-P}\label{table_$3$D}
\end{subtable}%
}
\label{table_3}
\vspace{-1.5em}
\end{table*}
\subsubsection{Recall-only Discussions}
\label{ch4_sub_recal}
Recall-only discussions are based on the numbers provided in Table \ref{table_2} and Fig. \ref{fig6}.\\
\noindent \textbf{Clutter, Viewpoint, Texture-less objects.} Highest recall values are obtained on the LINEMOD dataset (see Fig. \ref{fig6}), meaning that the state-of-the-art methods for $6$D object pose estimation can successfully handle the challenges, clutter, varying viewpoint, and texture-less objects. LCHF, detecting more than half of the objects with over $80 \%$ accuracy, worst performs on \enquote{box} and \enquote{glue} (see Table \ref{table_2} (a)), since these objects have planar surfaces, which confuses the features extracted in depth channel (example images are given in Fig. \ref{fig7} (a)).\\
\noindent \textbf{Occlusion.} In addition to the challenges involved in LINEMOD, occlusion is introduced in MULT-I. Linemod's performance decreases, since occlusion affects holistic feature representations in color and depth channels. LCHF performs better on this dataset than Linemod. Since LCHF is trained using the parts coming from positive training images, it can easily handle occlusion, using the information acquired from occlusion-free parts of the target objects. However, LCHF degrades on \enquote{camera}. In comparison with the other objects in the dataset, \enquote{camera} has relatively smaller dimensions. In most of the test images, there are non-negligible amount of missing depth pixels (Fig. \ref{fig7} (b)) along the borders of this object, and thus confusing the features extracted in depth channel. In such cases, LCHF is liable to detect similar-looking out of training objects and generate many false positives (see Fig. \ref{fig7} (c)). The hypotheses produced by LCHF for \enquote{joystick} are all considered as false positive (Fig. \ref{fig7} (d)). When we re-evaluate the recall that LCHF produces on the \enquote{joystick} object setting $z_{\omega}$ to the value of $0.15$, we observe $89 \%$ accuracy.\\
\noindent \textbf{Severe Occlusion.} OCC involves challenging test images where the objects of interest are cluttered and severely occluded. The best performance on this dataset is caught by PoseCNN \cite{39}, and there is still room for improvement in order to fully handle this challenge. Despite the fact that the distinctive feature of this benchmark is the existence of \enquote{severe occlusion}, there are occlusion-free target objects in several test images. In case the test images of a target object include unoccluded and/or naively occluded samples (with the occlusion ratio up to $40 \% -50 \%$ of the object dimensions) in addition to severely occluded samples, methods produce relatively higher recall values (\textit{e.g.}, \enquote{can, driller, duck, holepuncher}, Table \ref{table_2} (c)). On the other hand, when the target object has additionally other challenges such as planar surfaces, methods' performance (LCHF and Linemod) decreases (\textit{e.g.}, \enquote{box}, Fig. \ref{fig7} (e)).\\
\noindent \textbf{Severe Clutter.} In addition to the challenges discussed above, BIN-P inherently involves severe clutter, since it is designed for bin-picking scenarios, where objects are arbitrarily stacked in a pile. According to the recall values presented in Table \ref{table_2} (d), LCHF and Brachmann et al. \cite{28} perform $ 25 \%$ better than Linemod. Despite having severely occluded target objects in this dataset, there are unoccluded/relatively less occluded objects at the top of the bin. Since our current analyses are based on the top hypothesis of each method, the produced success rates show that the methods can recognize the objects located on top of the bin with reasonable accuracy (Fig. \ref{fig7} (f)).\\
\noindent \textbf{Similar-Looking Distractors.} We test both Linemod and LCHF on the T-LESS dataset. Since most of the time the algorithms fail, we do not report quantitative analyses, instead we discuss our observations from the experiments. The dataset involves various object classes with strong shape and color similarities. When the background color is different than that of the objects of interest, color gradient features are successfully extracted. However, the scenes involve multiple instances, multiple objects similar in shape and color, and hence, the features queried exist in the scene at multiple locations. The features extracted in depth channel are also severely affected from the lack of discriminative selection of shape information. When the objects of interest have planar surfaces, the detectors cannot easily discriminate foreground and background in depth channel, since these objects in the dataset are relatively smaller in dimension (see Fig. \ref{fig7} (g)).\\
\noindent \textbf{Part-based vs. Holistic approaches.} Holistic methods \cite{2, 17, 43, 39, 37} formulate the detection problem globally. Linemod \cite{2} represents the windows extracted from RGB and depth images by the surface normals and color gradients features. Distortions along the object borders arising from occlusion and clutter, that is, the distortions of the color gradient and surface normal information in the test processes, mainly degrade the performance of this detector. Part-based methods \cite{4, 28, 35, 30} extract parts in the given image. Despite the fact that LCHF uses the same kinds of features as in Linemod, LCHF detects objects extracting parts, thus making the method more robust to occlusion and clutter.\\
\noindent \textbf{Template-based vs. Random forest-based.} Template-based methods, \textit{i.e.}, Linemod, match the features extracted during test to a set of templates, and hence, they cannot easily be generalized well to unseen ground truth annotations, that is, the translation and rotation parameters in object pose estimation. Methods based on random forests \cite{4, 28, 30} efficiently benefit the randomization embedded in this learning tool, consequently providing good generalization performance on new unseen samples.\\
\noindent \textbf{RGB-D vs. Depth.} Methods utilizing both RGB and depth channels demonstrate higher recall values than methods that are of using only depth, since RGB provides extra clues to ease the detection. This is depicted in Table \ref{table_2} (a) where learning- and template-based methods of RGB-D perform much better than point-to-point technique \cite{17} of depth channel.\\
\noindent \textbf{RGB-D vs. RGB (CNN structures).} More recent paradigm is to adopt CNNs to solve $6$D object pose estimation problem taking RGB images as inputs \cite{37, 39}. The methods \cite{37,38,40} working in the RGB channel in Table \ref{table_2} are based on CNN structure. According to the numbers presented in Table \ref{table_2}, RGB-based SSD-$6$D \cite{37} and RGB-D-based LCHF achieve similar performance. These recall values show the promising performance of CNN architectures across random forest-based learning methods.\\
\indent Robotic manipulators that pick and place the items from conveyors, shelves, pallets, \textit{etc.}, need to know the pose of one item per RGB-D image, even though there might be multiple items in its workspace. Hence our recall-only analyses mainly target to solve the problems that could be encountered in such cases. Based upon the analyses currently made, one can make important implications, particularly from the point of the performances of the detectors. On the other hand, recall-based analyses are not enough to illustrate which dataset is more challenging than the others. This is especially true in crowded scenarios where multiple instances of target objects are severely occluded and cluttered. Therefore, in the next part, we discuss the performances of the baselines from another aspect, regarding precision-recall curves and F1 scores, where the $6$D detectors are investigated sorting all detection scores across all images.
\begin{figure*}[!t]
\centering
\includegraphics[width=\linewidth]{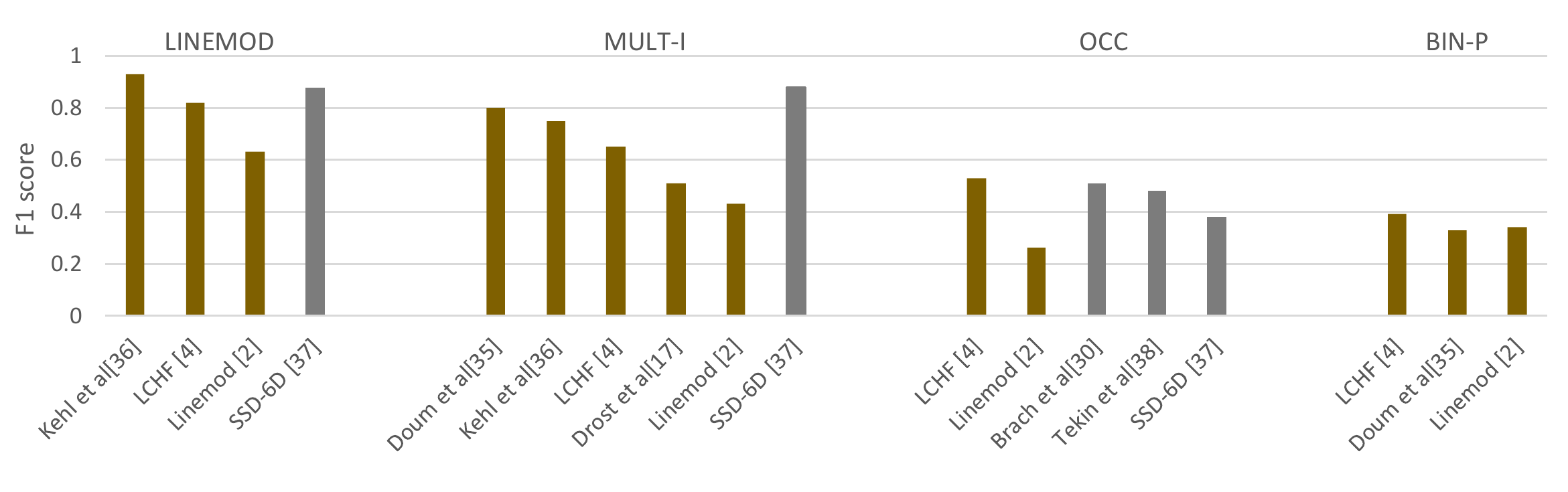}
\caption{Success of each baseline on every dataset is shown, F1 scores are computed using the Average Distance (AD) metric.}
\label{fig8}
\end{figure*}
\begin{figure*}[!t]
\centering
\includegraphics[width=14cm]{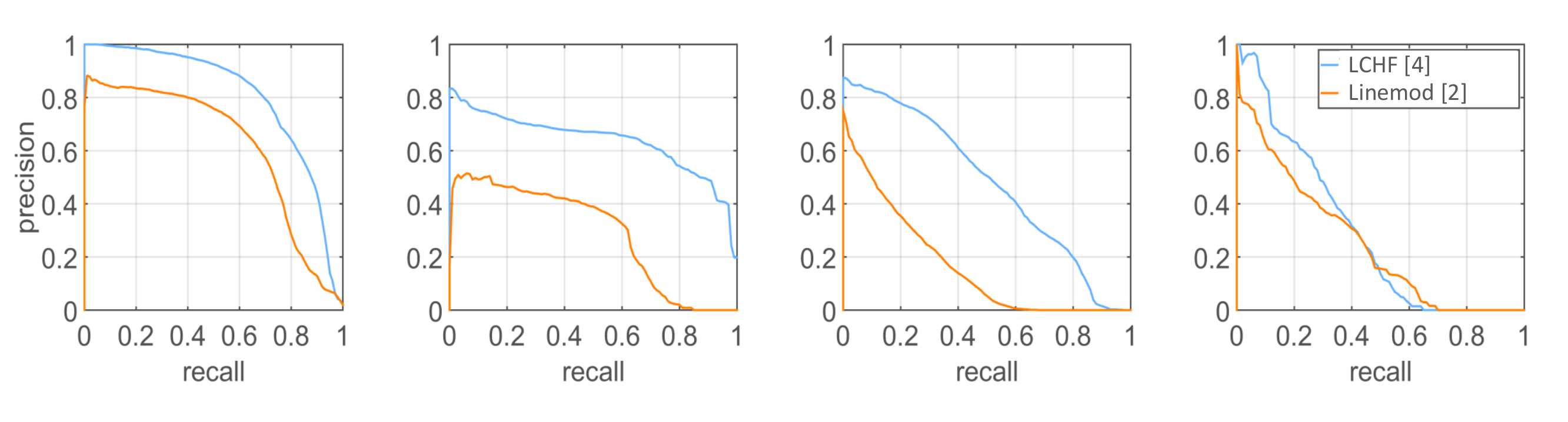}
\caption{Precision-recall curves of averaged F1 scores for LCHF \cite{4} and Linemod \cite{2} are shown: from left to right, LINEMOD, MULT-I, OCC, BIN-P.}
\label{fig9}
\end{figure*}
\begin{figure*}[!t]
\centering
\includegraphics[width=\linewidth]{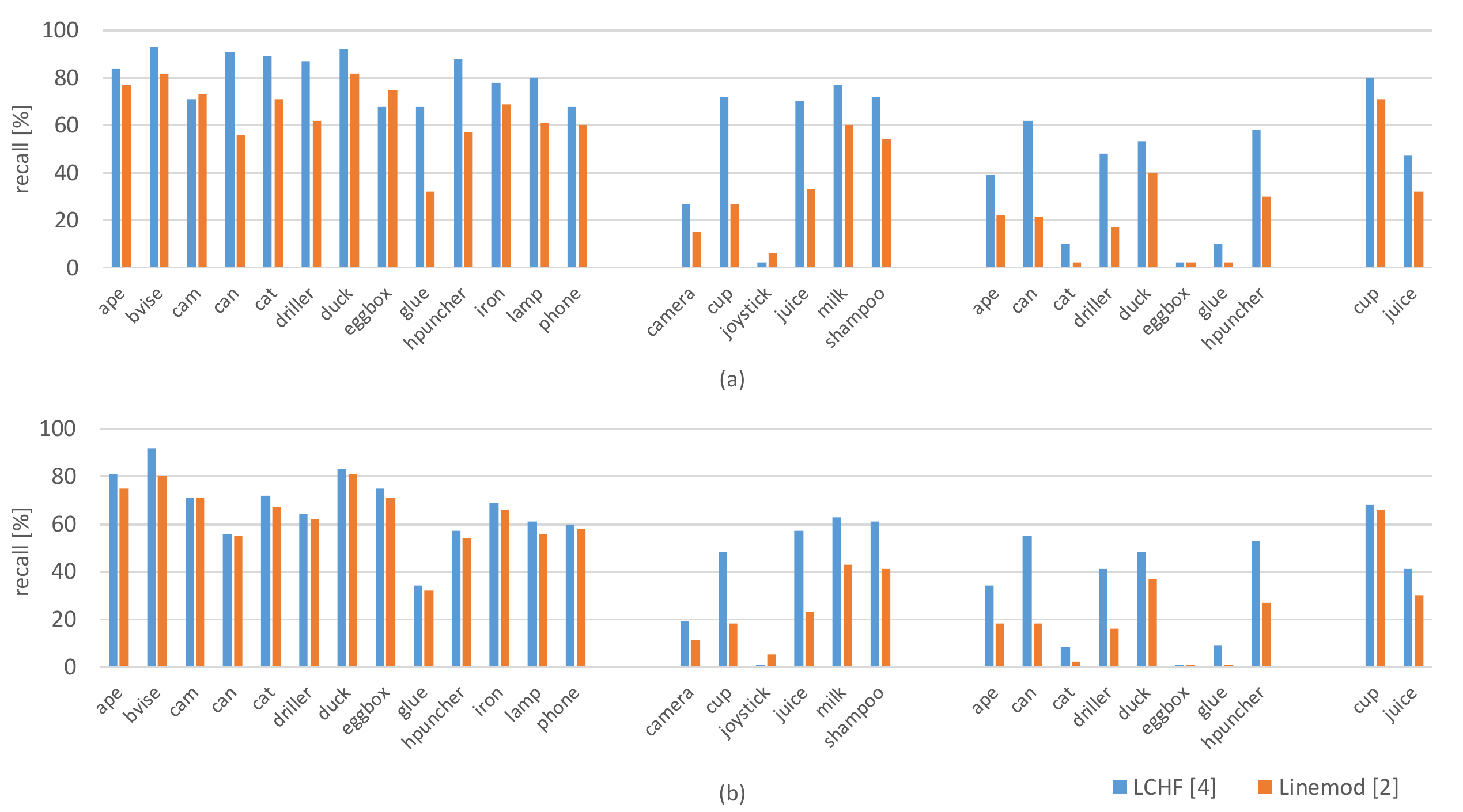}
\caption{Baselines are compared using Visible Surface Discrepancy (VSD) with respect to recall values. The performances of the baselines are depicted object-wise: from left to right, LINEMOD, MULT-I, OCC, and BIN-P. Parameters: (a) $\delta = 20$, $\tau = 100$, and $t = 0.5$, (b) $\delta = 15$, $\tau = 100$, and $t = 0.5$.}
\label{fig11}
\end{figure*}
\begin{figure*}[!t]
\centering
\includegraphics[width=\linewidth]{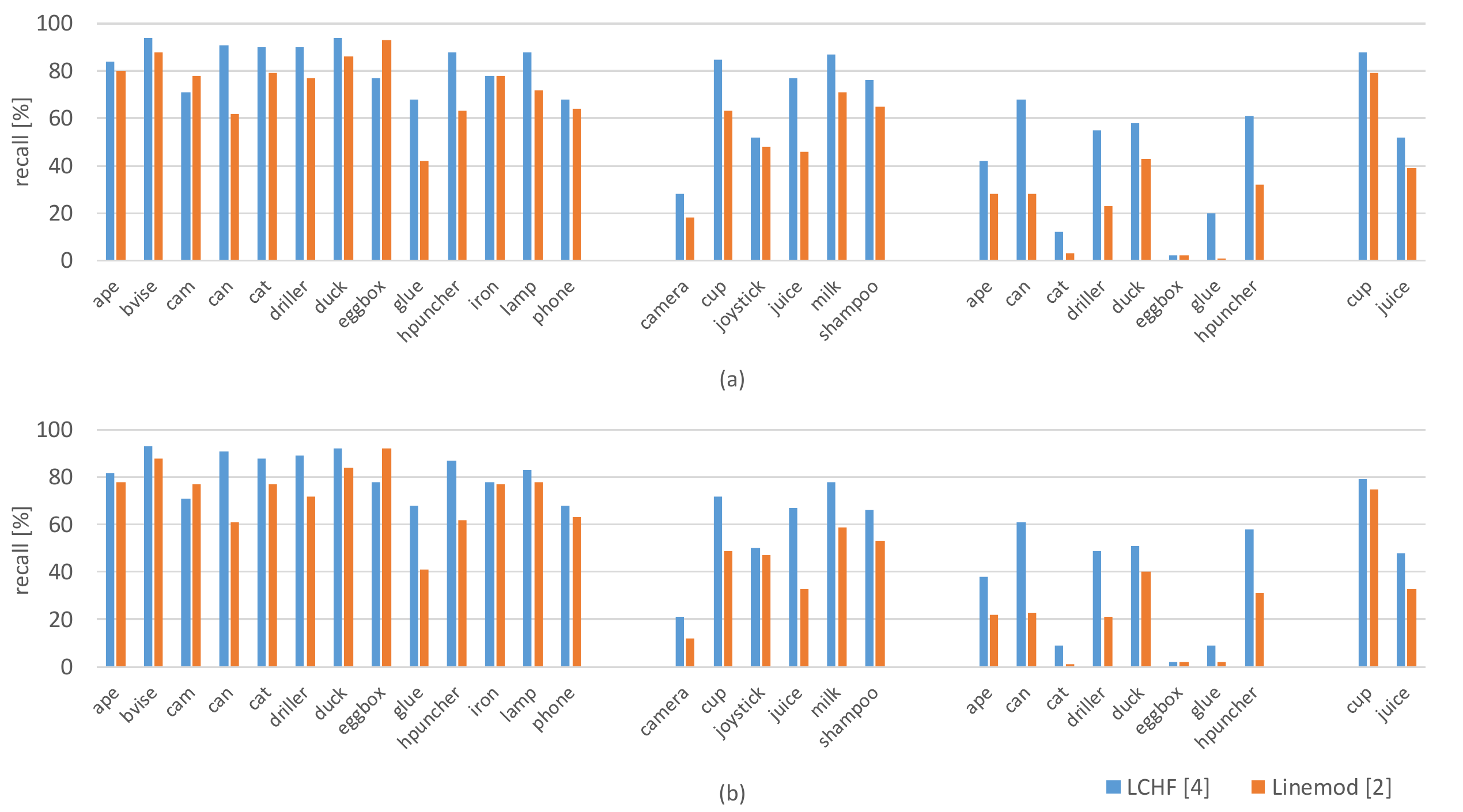}
\caption{Baselines are compared using Visible Surface Discrepancy (VSD) with respect to recall values. The performances of the baselines are depicted object-wise: from left to right, LINEMOD, MULT-I, OCC, and BIN-P. Parameters: (a) $\delta = 20$, $\tau = 100$, and $t = 0.6$, (b) $\delta = 15$, $\tau = 100$, and $t = 0.6$.}
\label{fig12}
\end{figure*}
\begin{figure*}[!t]
\centering
\includegraphics[width=\linewidth]{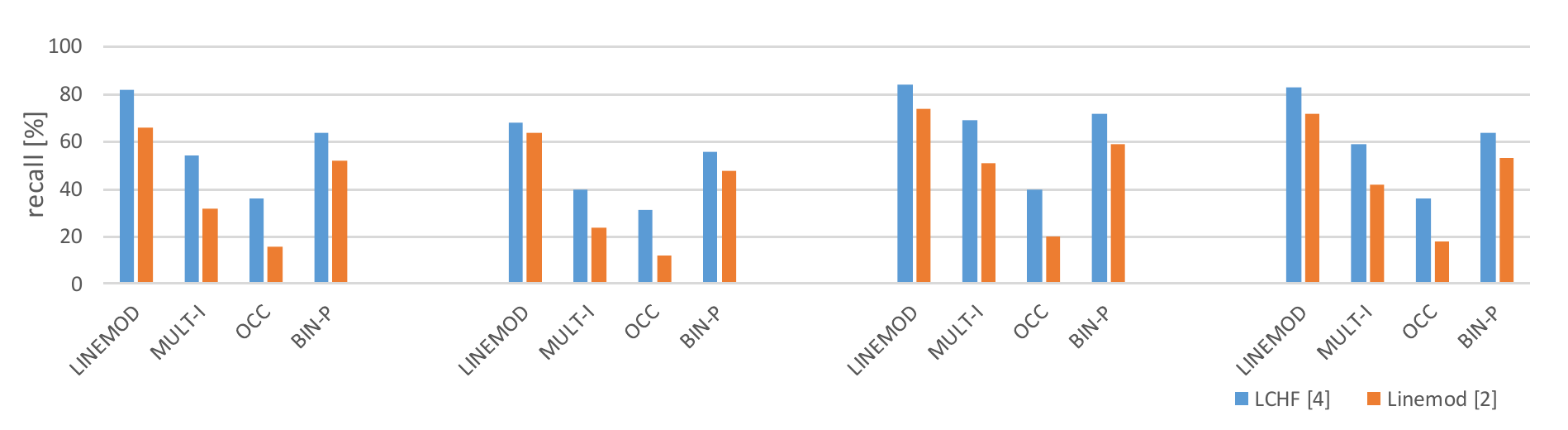}
\caption{Success of each baseline on every dataset is demonstrated averaging the recall values of individual objects. (a) $\delta = 20$, $\tau = 100$, $t = 0.5$, (b) $\delta = 15$, $\tau = 100$, $t = 0.5$, (c) $\delta = 20$, $\tau = 100$, $t = 0.6$, (d) $\delta = 15$, $\tau = 100$, $t = 0.6$.}
\label{fig13}
\vspace{-1em}
\end{figure*}
\begin{figure}[!t]
\centering
\includegraphics[width=\linewidth]{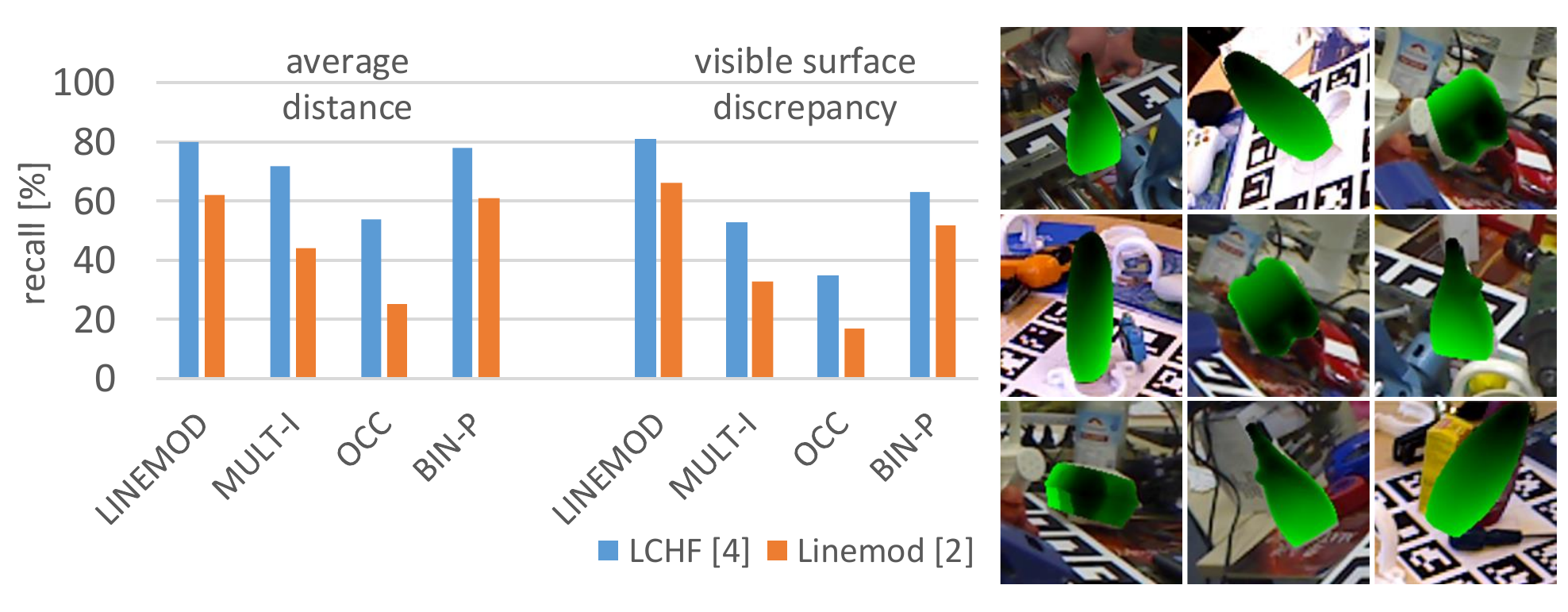}
\caption{Evaluations of LCHF and Linemod are compared with respect to both AD ($z_\omega = 0.1$) and VSD ($\delta = 20$, $\tau = 100$, and $t = 0.5$). Samples on the right are considered as false positive with respect to AD, whilst VSD deems correct.}
\label{fig14}
\vspace{-1em}
\end{figure}
\subsubsection{Precision-Recall Discussions}
\label{ch4_sub_pr}
Our precision-recall discussions are based on the F1 scores provided in Table \ref{table_3}, and Fig. \ref{fig8}.\\
\indent We first analyze the performance of the methods \cite{36, 4, 2, 37} on the LINEMOD dataset. On the average, Kehl et al. \cite{36} outperforms other methods proving the superiority of learning deep features. Despite estimating $6$D in RGB images, SSD-$6$D \cite{37} exhibits the advantages of using CNN structures for $6$D object pose estimation. LCHF and Linemod demonstrate lower performance, since the features used by these methods are manually-crafted. The comparison between Fig. \ref{fig6} and Fig. \ref{fig8} reveals that the results produced by the methods have approximately the same characteristics on the LINEMOD dataset, with respect to recall and F1 scores.\\
\indent The methods tested on the MULT-I dataset \cite{35,36,4,17,2} utilize the geometry information inherently provided by depth images. Despite this fact, SSD-$6$D \cite{37}, estimating $6$D pose only from RGB images, outperforms other methods clearly proving the superiority of using CNNs for the $6$D problem over other structures.\\
\indent LCHF \cite{4} and Brachmann et al. \cite{30} best perform on OCC with respect to F1 scores. As this dataset involves test images where highly occluded objects are located, the reported results depict the importance of designing part-based solutions.\\
\indent The most important difference is observed on the BIN-P dataset. While the success rates of the detectors on this dataset are higher than $60 \%$ with respect to the recall values (see Fig. \ref{fig6}), according to the presented F1 scores, their performance are less than $40 \%$. When we take into account all hypotheses and the challenges particular to this dataset, which are severe occlusion and severe clutter, we observe strong degradation in the accuracy of the detectors.\\
\indent In Fig. \ref{fig9}, we lastly report precision-recall curves of LCHF and Linemod. Regarding these curves, one can observe that as the datasets are getting more difficult, from the point of challenges involved, the methods produce less accurate results.\\
\vspace{-1em}
\subsection{Analyses Based on Visible Surface Discrepancy}
The analyses presented so far have been employed using the AD metric. We continue our discussions computing recall values for the two methods LCHF \cite{4} and Linemod \cite{2}, using the VSD metric. VSD renders the model of an object of interest at both ground truth and estimated poses, and intersects the depth maps of renderings with the test image in order to compute visibility masks. In the calculation of the visibility masks, there are several important parameters: i) the tolerance $\delta$, ii) the misalignment tolerance $\tau$, and iii) the threshold $t$. We firstly set $\delta = 20$ mm, $\tau = 100$ mm, and $t = 0.5$, and the bar chart in Fig. \ref{fig11} (a) demonstrates the  object-wise accuracy of each baseline on the LINEMOD, MULT-I, OCC, BIN-P datasets, respectively. Both baselines demonstrate over $60 \%$ recall on more than half of the objects involved in the LINEMOD dataset. The object \enquote{glue} is the one on which the methods degrade. In the MULT-I dataset, part-based LCHF performs better than Linemod. However, LCHF fails on detecting \enquote{joystick}, whilst the Linemod detector shows approximately $7 \%$ recall on this object. The baselines underperform on OCC, since this dataset has the test images where the objects are subjected to severe occlusion. The objects \enquote{cat}, \enquote{eggbox}, and \enquote{glue} are the most difficult ones, for which the performances of the methods are limited to at most $10 \%$ recall. Apart from severe occlusion, \enquote{eggbox} and \enquote{glue} have planar surfaces, an extra factor that impairs the features extracted in depth channel. Both baselines produce higher recall values on the \enquote{cup} object involved in the BIN-P dataset, whilst the methods' performance decreases when the concerned object is \enquote{juice}, since multiple planar surfaces in the scene cause severe clutter for the detectors.\\
\indent Next, we present the accuracy of each method altering the parameters of VSD. Figure \ref{fig11} (b) shows the recall values when $\delta = 15$, $\tau = 100$, and $t = 0.5$. Note that these are also the parameters used in the experiments of the original publication of VSD \cite{3}. Slightly decreasing the parameter $\delta$ gives rise lower recall values, however, we cannot see any significant difference from the point of characteristics of the bar charts. Unlike the parameter $\delta$, when we change the threshold $t$, one can observe remarkable discrepancies for several objects. Figure \ref{fig12} demonstrates the accuracy of each method when the parameters $\delta$, $\tau$, $t$ are set to $20$ mm, $100$ mm, $0.6$, and $15$ mm, $100$ mm, $0.6$, respectively. The comparison between Fig. \ref{fig11} and Fig. \ref{fig12} depicts that the moderation in the parameter $t$ gives rise approximately $50 \%$ accuracy on the \enquote{joystick} object for both LCHF and Linemod.\\
\indent We lastly average the object-wise results shown in Figs. \ref{fig11} and \ref{fig12}, and report the success of each baseline for each dataset in Fig. \ref{fig13}. According to this figure, highest recall values are reached when $\delta = 20$, $\tau = 100$, and $t = 0.6$. According to the visual results, we found that the set of parameters, $\delta = 20$, $\tau = 100$, and $t = 0.5$, is the one most appropriate.\\
\indent In Fig. \ref{fig14} the evaluations of LCHF and Linemod are compared with respect to both AD ($z_\omega = 0.1$) and VSD ($\delta = 20$, $\tau = 100$, and $t = 0.5$). The characteristic of both chart is the same, that is, both methods, according to AD and VSD, perform best on the LINEMOD dataset, whilst worst on OCC. On the other hand, the main advantage of VSD is that it features ambiguity-invariance: since it is designed to evaluate the baselines over the visible parts of the objects, it gives more robust measurements across symmetric objects. Sample images in Fig. \ref{fig14} show the hypotheses of symmetric objects which are considered as false positive according to the AD metric, whilst VSD accepts those as correct. 
\vspace{-0.5em}
\section{Summary and Observations}
\label{Summary} 
Review studies available in the literature introduced important implications for the problem of object pose recovery, however, they mainly addressed generalized object detection in which the reviewed methods and discussions were restricted to visual level in $2$D. In this paper, we have taken the problem a step further and have presented a comprehensive review of the methods ranging from $3$D BB detectors to full $6$D pose estimators. We have formed a discriminative categorization of the methods based on their problem modeling approaches. Along with the examination on the datasets and evaluations metrics, we have conducted multi-modal analyses investigating methods' performance in RGB and depth channels.\\
\indent 2D-driven 3D BB detection methods enlarge the 2D search space using the available appearance and geometry information in the 3D space along with RGB images. The methods directly detecting 3D BBs either use monocular RGB images together with contextual models as well as semantics or utilize stereo images, RGB-D cameras, and LIDAR sensors. On one hand, RGB provides dense measurements, however it is not efficient for handling objects located far away. On the other hand, LIDAR sensors precisely calculate depth information while its downside is the sparsity. Capitalizing their advantages, recent methods fuse two channels.\\
\indent Template matching, point-pair feature matching, and random forest-based learning methods for 6D object pose estimation have dominated the area within the first half of the past decade. Typically working in RGB-D or in depth channel, accurate results have been obtained on textured-objects at varying viewpoints with cluttered backgrounds. Part-based solutions can handle the occlusion problem better than the ones global, using the information acquired from occlusion-free parts of the target objects. In the BOP challenge, point-pair feature matching methods have shown overperformance, however, heavy existence of occlusion and clutter severely affects the detectors. Similar-looking distractors along with similar-looking object classes seem the biggest challenge for such methods, since the lack of discriminative selection of shape features strongly confuse recognition systems. Hence, the community has initiated learning deep discriminative features. The most recent paradigm is to adopt end-to-end trainable CNN architectures, and solve the problem in RGB-only.\\
\indent As $3$D amodal BB detection methods are engineered to work at the level of categories, they are mainly evaluated on the KITTI \cite{126}, NYU-Depth v2 \cite{127}, and SUN RGB-D \cite{128} datasets, where the objects of interest are aligned with the gravity direction. Unlike the $3$D methods, full $6$D object pose estimators are typically designed to work at the level of instances where the models of the objects of interest are utilized. $6$D methods are tested on the datasets, such as LINEMOD \cite{2}, MULT-I \cite{4}, T-LESS \cite{42}, which have $6$D object pose annotations. Exploiting the advantages of both category- and instance-level methods, $6$D object pose estimation could be addressed at the level of categories. To this end, the random forest-based ISA \cite{129} utilizes skeleton structures of models along with Semantically Selected Centers. Recently presented deep net in \cite{130} estimates Normalized Object Coordinate Space, and has no dependence on the objects' $3$D models. These methods could further be improved explicitly handling the challenges of the categories, \textit{e.g.}, intra-class variations, distribution shifts. The creation of novel 6D annotated datasets for categories can allow the community to measure the robustness of the methods.\\

\ifCLASSOPTIONcaptionsoff
  \newpage
\fi


\begin{thebibliography}{1}

\bibitem{1}
C. Eppner, S. Hofer, R. Jonschkowski, R. Martin-Martin, A. Sieverling, V. Wall, and O. Brock, 
\emph{Lessons from the Amazon Picking Challenge: Four aspects of building robotic systems}, 
Proceedings of Robotics: Science and Systems, 2016.

\bibitem{2}
S. Hinterstoisser, V. Lepetit, S. Ilic, S. Holzer, G. Bradski, K. Konolige, and N. Navab,
\emph{Model based training, detection, and pose estimation of texture-less $3$D objects in heavily cluttered scenes},
ACCV, 2012.

\bibitem{3}
T. Hodan, J. Matas, and S. Obdrzalek, 
\emph{On evaluation of $6$D object pose estimation},
ECCVW, 2016.

\bibitem{4}
A. Tejani, D. Tang, R. Kouskouridas, and T-K. Kim,
\emph{Latent-class Hough forests for $3$D object detection and pose estimation}, 
ECCV, 2014.

\bibitem{5}
T. Hodan, F. Michel, E. Brachmann, W. Kehl, A. Buch, D. Kraft, B. Drost, J. Vidal, S. Ihrke, X. Zabulis, C. Sahin, F. Manhardt, F. Tombari, T-K. Kim, J. Matas, and C. Rother,
\emph{BOP: Benchmark for $6$D object pose estimation},
ECCV, 2018.

\bibitem{6}
J. Deng, W. Dong, R. Socher, L.J. Li, K. Li, and L. Fei-Fei,
\emph{ImageNet: A large-scale hierarchical image database}, 
CVPR, 2009.

\bibitem{7}
M. Everingham, L.V. Gool, C.K. Williams, J. Winn, and A. Zisserman,
\emph{The PASCAL Visual Object Classes (VOC) challenge},
IJCV, 2010.

\bibitem{8}
S.K. Divvala, D. Hoiem, J.H. Hays, A.A. Efros, and M. Hebert, 
\emph{An empirical study of context in object detection},
CVPR, 2009.

\bibitem{9}
D. Hoiem, Y. Chodpathumwan, and Q. Dai,
\emph{Diagnosing error in object detectors},
ECCV, 2012.

\bibitem{10}
O. Russakovsky, J. Deng, Z. Huang, A.C. Berg, and L. Fei-Fei, 
\emph{Detecting avocados to zucchinis: What have we done, and where are we going?} 
ICCV, 2013.

\bibitem{11}
A. Torralba and A.A. Efros,
\emph{Unbiased look at dataset bias},
CVPR, 2011.

\bibitem{12}
M. Everingham, S.A. Eslami, L.V. Gool, C.K. Williams, J. Winn, and A. Zisserman,
\emph{The PASCAL Visual Object Classes challenge: A retrospective},
IJCV, 2015.

\bibitem{13}
O. Russakovsky, J. Deng, H. Su, J. Krause, S. Satheesh, S. Ma, Z. Huang, A. Karpathy, A. Khosla, M. Bernstein, and A.C. Berg, 
\emph{ImageNet large scale visual recognition challenge},
IJCV, 2015.

\bibitem{14}
R. B. Rusu, N. Blodow, and M. Beetz, 
\emph{Fast point feature histograms (FPFH) for $3$D registration},
ICRA, 2009.

\bibitem{15}
E. Kim and G. Medioni, 
\emph{$3$D object recognition in range images using visibility context},
IROS, 2011.

\bibitem{16}
C. Papazov and D. Burschka, 
\emph{An efficient RANSAC for $3$D object recognition in noisy and occluded scenes},
Asian Conference on Computer Vision, 2010.

\bibitem{17}
B. Drost, M. Ulrich, N. Navab, and S. Ilic, 
\emph{Model globally, match locally: Efficient and robust $3$D object recognition},
CVPR, 2010.

\bibitem{18}
C. Choi and H. I. Christensen, 
\emph{$3$D pose estimation of daily objects using an RGB-D camera},
IROS, 2012.

\bibitem{19}
C. Choi, Y. Taguchi, O. Tuzel, M.L. Liu, and S. Ramalingam, 
\emph{Voting-based pose estimation for robotic assembly using a $3$D sensor},
International Conference on Robotics and Automation, 2012.

\bibitem{20}
B. Drost and S. Ilic, 
\emph{$3$D Object Detection and Localization using Multimodal Point Pair Features},
$3$D Imaging, Modeling, Processing, Visualization and Transmission, 2012.

\bibitem{21}
R.P. de Figueiredo, P. Moreno, and A. Bernardino, 
\emph{Fast $3$D Object Recognition of Rotationally Symmetric Objects},
Pattern Recognition and Image Analysis, 2013.

\bibitem{22}
O. Tuzel, M.Y. Liu, Y. Taguchi, and A. Raghunathan, 
\emph{Learning to Rank $3$D Features},
ECCV, 2014.

\bibitem{23}
M. Y. Liu, O. Tuzel, A. Veeraraghavan, Y. Taguchi, T. K. Marks, and R. Chellappa, 
\emph{Fast object localization and pose estimation in heavy clutter for robotic bin picking},
IJRR, 2012.

\bibitem{24}
R. Rios-Cabrera and T. Tuytelaars, 
\emph{Discriminatively trained templates for $3$D object detection: A real time scalable approach},
ICCV, 2013.

\bibitem{25}
T. Hodan, X. Zabulis, M. Lourakis, S. Obdrzalek, and J. Matas, 
\emph{Detection and fine $3$D pose estimation of texture-less objects in RGB-D images},
IROS, 2015.

\bibitem{26}
J. Sock, S. H. Kasaei, L. S. Lopes, and T-K. Kim, 
\emph{Multi-view $6$D Object Pose Estimation and Camera Motion Planning using RGBD Images},
3rd International Workshop on Recovering $6$D Object Pose, 2017.

\bibitem{27}
U. Bonde, V. Badrinarayanan, and R. Cipolla, 
\emph{Robust instance recognition in presence of occlusion and clutter},
ECCV, 2014.

\bibitem{28}
E. Brachmann, A. Krull, F. Michel, S. Gumhold, J. Shotton, and C. Rother, 
\emph{Learning $6$D object pose estimation using $3$D object coordinates},
ECCV, 2014.

\bibitem{29}
A. Krull, E. Brachmann, F. Michel, M. Y. Yang, S. Gumhold, and C. Rother, 
\emph{Learning analysis-by-synthesis for $6$D pose estimation in RGB-D images},
ICCV, 2015.

\bibitem{30}
E. Brachmann, F. Michel, A. Krull, M.Y. Yang, S. Gumhold, and C. Rother, 
\emph{Uncertainty-Driven $6$D Pose Estimation of Objects and Scenes from a Single RGB Image},
CVPR, 2016.

\bibitem{31}
C. Sahin, R. Kouskouridas, and T-K. Kim, 
\emph{Iterative Hough Forest with Histogram of Control Points for 6 DoF Object Registration from Depth Images},
IROS, 2016.

\bibitem{32}
C. Sahin, R. Kouskouridas, and T-K. Kim, 
\emph{A Learning-based Variable Size Part Extraction Architecture for $6$D object pose recovery in depth images},
Journal of Image and Vision Computing 63, 2017, 38-50.

\bibitem{33}
F. Michel, A. Kirillov, E. Brachmann, A. Krull, S. Gumhold, B. Savchynskyy, and C. Rother, 
\emph{Global hypothesis generation for $6$D object pose estimation},
CVPR, 2017.

\bibitem{34}
P. Wohlhart and V. Lepetit, 
\emph{Learning descriptors for object recognition and $3$D pose estimation},
CVPR, 2015.

\bibitem{35}
A. Doumanoglou, R. Kouskouridas, S. Malassiotis, and T-K. Kim, 
\emph{Recovering $6$D Object Pose and Predicting Next-Best-View in the Crowd},
CVPR, 2016.

\bibitem{36}
W. Kehl, F. Milletari, F. Tombari, S. Ilic, and N. Navab, 
\emph{Deep learning of local RGB-D patches for $3$D object detection and $6$D pose estimation},
ECCV, 2016.

\bibitem{37}
W. Kehl, F. Manhardt, F. Tombari, S. Ilic, and N. Navab, 
\emph{{SSD-$6$D}: Making RGB-based $3$D detection and $6$D pose estimation great again},
CVPR, 2017.

\bibitem{38}
B. Tekin, S. N. Sinha, and P. Fua, 
\emph{Real-Time Seamless Single Shot $6$D Object Pose Prediction},
CVPR, 2018.

\bibitem{39}
Y. Xiang, T. Schmidt, V. Narayanan, and D. Fox, 
\emph{PoseCNN: A Convolutional Neural Network for $6$D Object Pose Estimation in Cluttered Scenes},
Robotics: Science and Systems (RSS), 2018.

\bibitem{40}
M. Rad and V. Lepetit, 
\emph{{BB8}: A Scalable, Accurate, Robust to Partial Occlusion Method for Predicting the $3$D Poses of Challenging Objects without Using Depth},
ICCV, 2017.

\bibitem{41}
V. Balntas, A. Doumanoglou, C. Sahin, J. Sock, R. Kouskouridas, and T-K. Kim, 
\emph{Pose Guided RGBD Feature Learning for $3$D Object Pose Estimation},
ICCV, 2017.

\bibitem{42}
T. Hodan, P. Haluza, S. Obdrzalek, J. Matas, M. Lourakis, and X. Zabulis, 
\emph{{T-LESS}: An RGB-D dataset for $6$D pose estimation of texture-less objects},
WACV, 2017.

\bibitem{43}
S. Hinterstoisser, V. Lepetit, N. Rajkumar, and K. Konolige, 
\emph{Going further with point-pair features},
ECCV, 2016.

\bibitem{44}
W. Kehl, F. Tombari, N. Navab, S. Ilic, and V. Lepetit, 
\emph{Hashmod: A Hashing Method for Scalable $3$D Object Detection},
BMVC, 2015.

\bibitem{45}
A. Hettiarachchi and D. Wigdor,
\emph{Annexing reality: Enabling opportunistic use of everyday objects as tangible proxies in augmented reality},
Proceedings of the 2016 CHI Conference on Human Factors in Computing Systems, pp. 1957-1967. ACM, 2016.

\bibitem{46}
D. Lindlbauer, J. Mueller, and M. Alexa,
\emph{Changing the Appearance of Real-World Objects By Modifying Their Surroundings},
Proceedings of the 2017 CHI Conference on Human Factors in Computing Systems, pp. 3954-3965. ACM, 2017.

\bibitem{47}
F. Born and M. Masuch,
\emph{Increasing Presence in a Mixed Reality Application by Integrating a Real Time Tracked Full Body Representation},
International Conference on Advances in Computer Entertainment, pp. 46-60. Springer, Cham, 2017.

\bibitem{48}
R. Xiao, J. Schwarz, N. Throm, A. D. Wilson, and H. Benko,
\emph{MRTouch: Adding Touch Input to Head-Mounted Mixed Reality},
IEEE transactions on visualization and computer graphics 24, no. 4 (2018): 1653-1660.

\bibitem{49}
M. Sra, S. Garrido-Jurado, C. Schmandt, and P. Maes,
\emph{Procedurally generated virtual reality from $3$D reconstructed physical space},
Proceedings of the 22nd ACM Conference on Virtual Reality Software and Technology, pp. 191-200. ACM, 2016.

\bibitem{50}
S-Y. Teng, T-S. Kuo, C. Wang, C-H. Chiang, D-Y. Huang, L. Chan, and B-Y. Chen,
\emph{PuPoP: Pop-up Prop on Palm for Virtual Reality},
The 31st Annual ACM Symposium on User Interface Software and Technology, pp. 5-17. ACM, 2018.

\bibitem{51}
D. Lindlbauer and A. D. Wilson, 
\emph{Remixed Reality: Manipulating Space and Time in Augmented Reality},
Proceedings of the 2018 CHI Conference on Human Factors in Computing Systems, p. 129. ACM, 2018.

\bibitem{52}
D. Harley, A. P. Tarun, D. Germinario, and A. Mazalek, 
\emph{Tangible VR: Diegetic tangible objects for virtual reality narratives},
Proceedings of the 2017 Conference on Designing Interactive Systems, pp. 1253-1263. ACM, 2017.

\bibitem{53}
I. F. Mondragon, P. Campoy, C. Martinez, and M. A. Olivares-Mendez,
\emph{$3$D pose estimation based on planar object tracking for UAVs control},
ICRA, 2010. 

\bibitem{54}
I. Dryanovski, W. Morris, and J. Xiao,
\emph{An open-source pose estimation system for micro-air vehicles},
ICRA, 2011.

\bibitem{55}
A. Causo, Z.H. Chong, R. Luxman, Y.Y. Kok, Z. Yi, W.C. Pang, R. Meixuan, Y.S. Teoh, W. Jing, H.S. Tju, and I.M. Chen,
\emph{A robust robot design for item picking},
ICRA, 2018.

\bibitem{56}
J. Huang and M. Cakmak, 
\emph{Code3: A system for end-to-end programming of mobile manipulator robots for novices and experts},
Proceedings of the 2017 ACM/IEEE International Conference on Human-Robot Interaction ACM, 2017.

\bibitem{57}
R. Patel, R. Curtis, B. Romero, and N. Correll,
\emph{Improving grasp performance using in-hand proximity and contact sensing},
Robotic Grasping and Manipulation Challenge (pp. 146-160), 2016.

\bibitem{58}
M. Schwarz, A. Milan, A. S. Periyasamy, and S. Behnke,
\emph{RGB-D object detection and semantic segmentation for autonomous manipulation in clutter},
The International Journal of Robotics Research, 37(4-5), 437-451, 2018.

\bibitem{60}
A. Zeng, S. Song, K.T. Yu, E. Donlon, F.R. Hogan, M. Bauza, D. Ma, O. Taylor, M. Liu, E. Romo, and N. Fazeli,
\emph{Robotic pick-and-place of novel objects in clutter with multi-affordance grasping and cross-domain image matching},
ICRA, 2018.

\bibitem{62}
M. Zhu, K.G. Derpanis, Y. Yang, S. Brahmbhatt, M. Zhang, C. Phillips, M. Lecce, and K. Daniilidis,
\emph{Single image $3$D object detection and pose estimation for grasping},
ICRA, 2014.

\bibitem{64}
M. Blösch, S. Weiss, D. Scaramuzza, and R. Siegwart, 
\emph{Vision based MAV navigation in unknown and unstructured environments},
ICRA, 2010.

\bibitem{65}
C. Sahin, T.K Kim,
\emph{Recovering 6D object pose: A review and multi-modal analysis},
ECCV Workshops, 2018.

\bibitem{66}
E. Sucar and J. B. Hayet, 
\emph{Bayesian scale estimation for monocular SLAM based on generic object detection for correcting scale drift},
ICRA, 2018.

\bibitem{67}
K. Tateno, F. Tombari, I. Laina, and N. Navab,
\emph{CNN-SLAM: Real-time dense monocular SLAM with learned depth prediction},
CVPR, 2017.

\bibitem{68}
P. Parkhiya, R. Khawad, J.K. Murthy, B. Bhowmick, and K. M. Krishna,
\emph{Constructing Category-Specific Models for Monocular Object-SLAM},
ICRA, 2018.

\bibitem{69}
J. McCormac, R. Clark, M. Bloesch, A. Davison, and S. Leutenegger,
\emph{Fusion++: Volumetric Object-Level SLAM},
International Conference on $3$D Vision ($3$DV) (pp. 32-41), 2018.

\bibitem{70}
S. Choudhary, L. Carlone, C. Nieto, J. Rogers, Z. Liu, H.I. Christensen, and F. Dellaert,
\emph{Multi robot object-based SLAM},
International Symposium on Experimental Robotics (pp. 729-741). Springer, 2016.

\bibitem{71}
M. Hosseinzadeh, K. Li, Y. Latif, and I. Reid,
\emph{Real-Time Monocular Object-Model Aware Sparse SLAM},
ICRA, 2019.

\bibitem{72}
D. Gálvez-López, M. Salas, J. D. Tardós, and J. M. M. Montiel, 
\emph{Real-time monocular object SLAM},
Robotics and Autonomous Systems, 75, 435-449, 2016.

\bibitem{73}
B. Mu, S. Y. Liu, L. Paull, J. Leonard, and J. P. How,
\emph{SLAM with objects using a nonparametric pose graph},
IROS, 2016.

\bibitem{74}
R. F. Salas-Moreno, R. A. Newcombe, H. Strasdat, P. H. Kelly, and A. J. Davison,
\emph{SLAM++: Simultaneous localisation and mapping at the level of objects},
CVPR, 2013.

\bibitem{75}
C. Sahin, M. Unel,
\emph{Under vehicle perception for high level safety measures using a catadioptric camera system},
IECON 39th Annual Conference of the IEEE Industrial Electronics Society, IEEE, 2013, pp. 4306–4311.

\bibitem{76}
C. Zach, A. Penate-Sanchez, and M.T. Pham,
\emph{A dynamic programming approach for fast and robust object pose recognition from range images},
CVPR, 2015.

\bibitem{77}
J. Sock, K.I. Kim, C. Sahin, and T-K. Kim,
\emph{Multi-Task Deep Networks for Depth-Based $6$D Object Pose and Joint Registration in Crowd Scenarios},
BMVC, 2018.

\bibitem{78}
D. Lin, S. Fidler, and R. Urtasun,
\emph{Holistic scene understanding for $3$D object detection with RGBD cameras},
ICCV, 2013.
 
\bibitem{79}
S. Gupta, R. Girshick, P. Arbelaez, and J. Malik,
\emph{Learning rich features from RGB-D images for object detection and segmentation},
ECCV, 2014.

\bibitem{80}
M. Engelcke, D. Rao, D.Z. Wang, C.H. Tong, and I. Posner,
\emph{Vote$3$Deep: Fast object detection in $3$D point clouds using efficient convolutional neural networks},
ICRA, 2017.

\bibitem{81}
Y. Zhang, M. Bai, P. Kohli, S. Izadi, and J. Xiao
\emph{DeepContext: Context-encoding neural pathways for $3$D holistic scene understanding},
ICCV, 2017.

\bibitem{82}
S. Song and J. Xiao,
\emph{Deep sliding shapes for amodal $3$D object detection in RGB-D images},
CVPR, 2016.

\bibitem{83}
Y. Zhou and O. Tuzel,
\emph{VoxelNet: End-to-end learning for point cloud based $3$D object detection},
CVPR, 2018.

\bibitem{85}
S. Gupta, P. Arbelaez, R. Girshick, and J. Malik,
\emph{Aligning $3$D models to RGB-D images of cluttered scenes},
CVPR, 2015.

\bibitem{86}
X. Chen, H. Ma, J. Wan, B. Li, and T. Xia,
\emph{Multi-view $3$D object detection network for autonomous driving},
CVPR, 2017.

\bibitem{87}
J. Lahoud and B. Ghanem,
\emph{$2$D-driven $3$D object detection in RGB-D images},
ICCV, 2017.

\bibitem{88}
X. Chen, K. Kundu, Z. Zhang, H. Ma, S. Fidler, and R. Urtasun,
\emph{Monocular $3$D object detection for autonomous driving},
In Proceedings of the IEEE Conference on Computer Vision and Pattern Recognition, 2016.

\bibitem{89}
Z. Deng and L. Jan Latecki,
\emph{Amodal detection of $3$D objects: Inferring $3$D bounding boxes from $2$D ones in RGB-depth images},
In Proceedings of the IEEE Conference on Computer Vision and Pattern Recognition (pp. 5762-5770), 2017.

\bibitem{90}
S. Gupta, P. Arbelaez, and J. Malik,
\emph{Perceptual organization and recognition of indoor scenes from RGB-D images},
In Proceedings of the IEEE Conference on Computer Vision and Pattern Recognition (pp. 564-571), 2013.

\bibitem{91}
A. Mousavian, D. Anguelov, J. Flynn, and J. Kosecka,
\emph{$3$D bounding box estimation using deep learning and geometry},
In Proceedings of the IEEE Conference on Computer Vision and Pattern Recognition (pp. 7074-7082), 2017.

\bibitem{92}
J. Papon and M. Schoeler,
\emph{Semantic pose using deep networks trained on synthetic RGB-D},
In Proceedings of the IEEE International Conference on Computer Vision (pp. 774-782), 2015.

\bibitem{93}
S. Song and J. Xiao,
\emph{Sliding shapes for $3$D object detection in depth images},
In European conference on computer vision (pp. 634-651). Springer, Cham., 2014.

\bibitem{94}
Y. Xiang, W. Choi, Y. Lin, and S. Savarese,
\emph{Data-driven $3$D voxel patterns for object category recognition},
In Proceedings of the IEEE Conference on Computer Vision and Pattern Recognition (pp. 1903-1911), 2015.

\bibitem{101}
U. Asif, M. Bennamoun, and F.A. Sohel,
\emph{RGB-D object recognition and grasp detection using hierarchical cascaded forests},
IEEE Transactions on Robotics, 33(3), pp.547-564, 2017.

\bibitem{102}
M. Bleyer, C. Rhemann, and C. Rother,
\emph{Extracting $3$D scene-consistent object proposals and depth from stereo images},
In European Conference on Computer Vision (pp. 467-481). Springer, Berlin, Heidelberg, 2012.

\bibitem{104}
H. Jiang,
\emph{Finding approximate convex shapes in RGBD images},
In European Conference on Computer Vision (pp. 582-596). Springer, Cham, 2014.

\bibitem{105}
H. Jiang and J. Xiao,
\emph{A linear approach to matching cuboids in RGBD images},
In Proceedings of the IEEE Conference on Computer Vision and Pattern Recognition (pp. 2171-2178), 2013.

\bibitem{106}
S.H. Khan, X. He, M. Bennamoun, F. Sohel, and R. Togneri,
\emph{Separating objects and clutter in indoor scenes},
In Proceedings of the IEEE Conference on Computer Vision and Pattern Recognition (pp. 4603-4611), 2015.

\bibitem{111}
D.Z. Wang and I. Posner,
\emph{Voting for Voting in Online Point Cloud Object Detection},
In Robotics: Science and Systems (Vol. 1, No. 3, pp. 10-15607), 2015.

\bibitem{113}
R.B. Rusu, N. Blodow, Z. Marton, A. Soos, and M. Beetz,
\emph{Towards $3$D object maps for autonomous household robots},
In 2007 IEEE/RSJ International Conference on Intelligent Robots and Systems (pp. 3191-3198). IEEE.

\bibitem{114}
R.B. Rusu, Z.C. Marton, N. Blodow, M. Dolha, and M. Beetz,
\emph{Towards $3$D point cloud based object maps for household environments},
Robotics and Autonomous Systems, 56(11), pp.927-941.

\bibitem{115}
D. Gehrig, P. Krauthausen, L. Rybok, H. Kuehne, U.D. Hanebeck, T. Schultz, and R. Stiefelhagen,
\emph{Combined intention, activity, and motion recognition for a humanoid household robot},
In 2011 IEEE/RSJ International Conference on Intelligent Robots and Systems (pp. 4819-4825). IEEE.

\bibitem{116}
C.R. Qi, W. Liu, C. Wu, H. Su, and L.J. Guibas,
\emph{Frustum pointnets for $3$D object detection from RGB-D data},
In Proceedings of the IEEE Conference on Computer Vision and Pattern Recognition (pp. 918-927), 2018.

\bibitem{117}
Y. Wang, W.L. Chao, D. Garg, B. Hariharan, M. Campbell, and K.Q. Weinberger,
\emph{Pseudo-lidar from visual depth estimation: Bridging the gap in $3$D object detection for autonomous driving},
In Proceedings of the IEEE Conference on Computer Vision and Pattern Recognition (pp. 8445-8453), 2019
 
\bibitem{118}
S. Shi, X. Wang, and H. Li,
\emph{PointRCNN: $3$D object proposal generation and detection from point cloud},
In Proceedings of the IEEE Conference on Computer Vision and Pattern Recognition (pp. 770-779), 2019.
 
\bibitem{119}
A.H. Lang, S. Vora, H. Caesar, L. Zhou, J. Yang, and O. Beijbom,
\emph{PointPillars: Fast encoders for object detection from point clouds},
In Proceedings of the IEEE Conference on Computer Vision and Pattern Recognition (pp. 12697-12705), 2019.
 
\bibitem{120}
B. Li, W. Ouyang, L. Sheng, X. Zeng, and X. Wang,
\emph{GS$3$D: An Efficient $3$D Object Detection Framework for Autonomous Driving},
In Proceedings of the IEEE Conference on Computer Vision and Pattern Recognition (pp. 1019-1028), 2019.
 
\bibitem{121}
X. Chen, K. Kundu, Y. Zhu, A.G. Berneshawi, H. Ma, S. Fidler, and R. Urtasun,
\emph{$3$D object proposals for accurate object class detection},
In Advances in Neural Information Processing Systems (pp. 424-432), 2015.
  
\bibitem{122}
J. Ku, A.D. Pon, and S.L. Waslander,
\emph{Monocular $3$D Object Detection Leveraging Accurate Proposals and Shape Reconstruction},
In Proceedings of the IEEE Conference on Computer Vision and Pattern Recognition (pp. 11867-11876), 2019.
  
\bibitem{123}
M. Liang, B. Yang, Y. Chen, R. Hu, and R. Urtasun,
\emph{Multi-Task Multi-Sensor Fusion for $3$D Object Detection},
In Proceedings of the IEEE Conference on Computer Vision and Pattern Recognition (pp. 7345-7353), 2019.
   
\bibitem{124}
J. Ku, M. Mozifian, J. Lee, A. Harakeh, and S.L. Waslander,
\emph{Joint $3$D proposal generation and object detection from view aggregation},
In 2018 IEEE/RSJ International Conference on Intelligent Robots and Systems (IROS) (pp. 1-8). IEEE.
 
\bibitem{125}
Z.Q. Zhao, P. Zheng, S.T. Xu, and X. Wu,
\emph{Object detection with deep learning: A review},
IEEE transactions on neural networks and learning systems, 2019.

\bibitem{126}
A. Geiger, P. Lenz, and R. Urtasun,
\emph{Are we ready for autonomous driving? The KITTI vision benchmark suite},
In 2012 IEEE Conference on Computer Vision and Pattern Recognition (pp. 3354-3361). IEEE.
 
\bibitem{127}
N. Silberman, D. Hoiem, P. Kohli, and R. Fergus,
\emph{Indoor segmentation and support inference from RGBD images},
In European Conference on Computer Vision (pp. 746-760). Springer, Berlin, Heidelberg.
 
\bibitem{128}
S. Song, S. Lichtenberg, and J. Xiao,
\emph{SUN RGB-D: A RGBD scene understanding benchmark suite},
In CVPR, 2015.

\bibitem{129}
C. Sahin and T-K. Kim,
\emph{Category-level $6$D Object Pose Recovery in Depth Images},
ECCVW on R$6$D, 2018.

\bibitem{130}
H. Wang, S. Sridhar, J. Huang, J. Valentin, S. Song, and L.J. Guibas,
\emph{Normalized Object Coordinate Space for Category-Level $6$D Object Pose and Size Estimation},
In Proceedings of the IEEE Conference on Computer Vision and Pattern Recognition, 2019 (pp. 2642-2651).

\bibitem{131}
R. Girshick, J. Donahue, T. Darrell, and J. Malik,
\emph{Rich feature hierarchies for accurate object detection and semantic segmentation},
In Proceedings of the IEEE conference on computer vision and pattern recognition (pp. 580-587), 2014

\bibitem{132}
R. Girshick,
\emph{Fast R-CNN},
In Proceedings of the IEEE international conference on computer vision (pp. 1440-1448), 2015.

\bibitem{133}
S. Ren, K. He, R. Girshick, and J. Sun,
\emph{Faster R-CNN: Towards real-time object detection with region proposal networks},
In Advances in neural information processing systems (pp. 91-99), 2015.

\bibitem{134}
J. Dai, Y. Li, K. He, and J. Sun,
\emph{R-FCN: Object detection via region-based fully convolutional networks},
In Advances in neural information processing systems (pp. 379-387), 2016.

\bibitem{135}
T.Y. Lin, P. Dollar, R. Girshick, K. He, B. Hariharan, and S. Belongie,
\emph{Feature pyramid networks for object detection},
In Proceedings of the IEEE conference on computer vision and pattern recognition (pp. 2117-2125), 2017.

\bibitem{136}
J. Redmon, S. Divvala, R. Girshick, and A. Farhadi,
\emph{You only look once: Unified, real-time object detection},
In Proceedings of the IEEE conference on computer vision and pattern recognition (pp. 779-788), 2016.

\bibitem{137}
W. Liu, D. Anguelov, D. Erhan, C. Szegedy, S. Reed, C.Y. Fu, and A.C. Berg,
\emph{SSD: Single shot multibox detector},
In European conference on computer vision (pp. 21-37). Springer, Cham, 2016.
 
\bibitem{138}
P. Krahenbuhl and V. Koltun,
\emph{Geodesic object proposals},
In European conference on computer vision (pp. 725-739). Springer, Cham, 2014.
 
\bibitem{139}
Z. Cai, Q. Fan, R.S. Feris, and N. Vasconcelos,
\emph{A unified multi-scale deep convolutional neural network for fast object detection},
In european conference on computer vision (pp. 354-370). Springer, Cham, 2016.

\bibitem{140}
J.R. Uijlings, K.E. Van De Sande, T. Gevers, and A.W. Smeulders,
\emph{Selective search for object recognition},
International journal of computer vision, 104(2), pp.154-171, 2013.

\bibitem{141}
A. Krizhevsky, I. Sutskever, and G.E. Hinton,
\emph{ImageNet classification with deep convolutional neural networks},
In Advances in neural information processing systems (pp. 1097-1105), 2012.
 
\bibitem{142}
X. Ren and D. Ramanan,
\emph{Histograms of sparse codes for object detection},
CVPR, 2013.
  
\bibitem{x17}
L. Breiman,
\emph{Random forests},
Mach. Learn. 45, 2001.

\bibitem{x18}
J. Gall and V. Lempitsky,
\emph{Class-specific Hough forests for object detection},
Decision forests for computer vision and medical image analysis, 2013.

\bibitem{x19}
F. Moosmann, B. Triggs, and F. Jurie,
\emph{Fast discriminative visual codebooks using randomized clustering forests},
Adv. Neural Inf. Proces. Syst. 2007, 985-992.

\bibitem{x20}
U. Bonde, T-K. Kim, and K. R. Ramakrishnan,
\emph{Randomised manifold forests for principal angle-based face recognition},
Asian Conference on Computer Vision, 2010.

\bibitem{x21}
C. Leistner, A. Saffari, J. Santner, and H. Bischof,
\emph{Semi-supervised random forests},
ICCV, 2009.

\bibitem{x22}
M. Sun, P. Kohli, and J. Shotton,
\emph{Conditional regression forests for human pose estimation},
CVPR, 2012.

\bibitem{x23}
J. Gall, A. Yao, N. Razavi, L. V. Gool, and V. Lempitsky,
\emph{Hough forests for object detection, tracking, and action recognition},
TPAMI, 2011.

\bibitem{x24}
A. Criminisi and J. Shotton
\emph{Decision forests for computer vision and medical image analysis},
Springer Science and Business Media, 2013.

\bibitem{143}
T. Birdal and S. Ilic,
\emph{Point pair features based object detection and pose estimation revisited},
International Conference on $3$D Vision. IEEE, pp. 527–535, 2015.

\bibitem{144}
C. Choi, A.J. Trevor, and H.I. Christensen,
\emph{RGB-D edge detection and edge-based registration},
IEEE/RSJ International Conference on Intelligent Robots and Systems. IEEE, pp. 1568–1575, 2013.

\bibitem{145}
C. Choi and H.I. Christensen,
\emph{RGB-D object pose estimation in unstructured environments},
Rob. Auton. Syst. 75, 595–613, 2016.

\bibitem{146}
J. Carreira and C. Sminchisescu,
\emph{CPMC: Automatic object segmentation using constrained parametric min-cuts},
IEEE Transactions on Pattern Analysis and Machine Intelligence, 34(7), pp.1312-1328, 2011.

\bibitem{147}
P. Arbelaez, J. Pont-Tuset, J.T. Barron, F. Marques, and J. Malik,
\emph{Multi-scale combinatorial grouping},
In Proceedings of the IEEE Conference on Computer Vision and Pattern Recognition, pages 328–335, 2014.

\bibitem{148}
C.R. Qi, H. Su, K. Mo, and L.J. Guibas,
\emph{PointNet: Deep learning on point sets for $3$D classification and segmentation},
In Proceedings of the IEEE Conference on Computer Vision and Pattern Recognition (pp. 652-660), 2017.

\bibitem{149}
C.R. Qi, L. Yi, H. Su, and L.J. Guibas,
\emph{PointNet++: Deep hierarchical feature learning on point sets in a metric space},
In Advances in neural information processing systems (pp. 5099-5108), 2017.

\bibitem{e150}
B. Li,
\emph{$3$D fully convolutional network for vehicle detection in point cloud},
In IEEE/RSJ International Conference on Intelligent Robots and Systems (IROS) (pp. 1513-1518). IEEE, 2017.

\bibitem{e151}
M. Liang, B. Yang, S. Wang, and R. Urtasun,
\emph{Deep continuous fusion for multi-sensor $3$D object detection},
In Proceedings of the European Conference on Computer Vision (ECCV) (pp. 641-656), 2018.

\bibitem{e152}
F. Chabot, M. Chaouch, J. Rabarisoa, C. Teuliere, and T. Chateau,
\emph{Deep MANTA: A coarse-to-fine many-task network for joint $2$D and $3$D vehicle analysis from monocular image},
In Proceedings of the IEEE Conference on Computer Vision and Pattern Recognition (pp. 2040-2049), 2017.

\bibitem{e154}
B. Xu and Z. Chen,
\emph{Multi-level fusion based $3$D object detection from monocular images},
In Proceedings of the IEEE Conference on Computer Vision and Pattern Recognition (pp. 2345-2353), 2018.

\bibitem{e156}
B. Yang, W. Luo, and R. Urtasun,
\emph{PIXOR: Real-time $3$D object detection from point clouds},
In Proceedings of the IEEE conference on Computer Vision and Pattern Recognition (pp. 7652-7660), 2018.

\bibitem{e157}
D. Xu, D. Anguelov, and A. Jain,
\emph{PointFusion: Deep sensor fusion for $3$D bounding box estimation},
In Proceedings of the IEEE Conference on Computer Vision and Pattern Recognition (pp. 244-253), 2018.

\bibitem{e158}
Y. Yan, Y. Mao, and B. Li,
\emph{SECOND: Sparsely embedded convolutional detection},
Sensors, 18(10), 2018.

\bibitem{e160}
Z. Ren and E.B. Sudderth,
\emph{Three-dimensional object detection and layout prediction using clouds of oriented gradients},
In Proceedings of the IEEE Conference on Computer Vision and Pattern Recognition (pp. 1525-1533), 2016.

\bibitem{e161}
B. Li, T. Zhang, and T. Xia,
\emph{Vehicle detection from $3$D lidar using fully convolutional network},
Robotics: Science and Systems, 2016.

\bibitem{e162}
Z. Wang and K. Jia,
\emph{Frustum ConvNet: Sliding Frustums to Aggregate Local Point-Wise Features for Amodal $3$D Object Detection},
IEEE IROS, 2019.

\bibitem{e163}
D. Maturana and S. Scherer,
\emph{VoxNet: A $3$D convolutional neural network for real-time object recognition},
In 2015 IEEE/RSJ International Conference on Intelligent Robots and Systems (IROS) (pp. 922-928). IEEE, 2015.

\bibitem{e164}
X. Du, M.H. Ang, S. Karaman, and D. Rus,
\emph{A general pipeline for $3$D detection of vehicles},
In 2018 IEEE International Conference on Robotics and Automation (ICRA) (pp. 3194-3200). IEEE, 2018.

\bibitem{e165a}
M. Z. Zia, M. Stark, and K. Schindler,
\emph{Are cars just $3$D boxes? Jointly estimating the $3$D shape of multiple objects},
In Proceedings of the IEEE Conference on Computer Vision and Pattern Recognition (pp. 3678-3685), 2014.

\bibitem{e165b}
S. Song and M. Chandraker,
\emph{Joint SFM and detection cues for monocular $3$D localization in road scenes},
In Proceedings of the IEEE Conference on Computer Vision and Pattern Recognition (pp. 3734-3742), 2015.

\bibitem{e166}
K. Simonyan and A. Zisserman,
\emph{Very deep convolutional networks for large-scale image recognition},
ICLR, 2015.

\bibitem{b167}
A. Kanezaki, Y. Matsushita, and Y. Nishida,
\emph{RotationNet: Joint object categorization and pose estimation using multiviews from unsupervised viewpoints},
In Proceedings of the IEEE Conference on Computer Vision and Pattern Recognition (pp. 5010-5019), 2018.

\bibitem{b168}
A. Kundu, Y. Li, and J.M. Rehg,
\emph{$3$D-RCNN: Instance-level $3$D object reconstruction via render-and-compare},
In Proceedings of the IEEE Conference on Computer Vision and Pattern Recognition (pp. 3559-3568), 2018.

\bibitem{b169}
Y. Hu, J. Hugonot, P. Fua, and M. Salzmann,
\emph{Segmentation-driven $6$D object pose estimation},
In Proceedings of the IEEE Conference on Computer Vision and Pattern Recognition (pp. 3385-3394), 2019.

\bibitem{b170}
S. Peng, Y. Liu, Q. Huang, X. Zhou, and H. Bao,
\emph{PVNet: Pixel-wise Voting Network for $6$DoF Pose Estimation},
In Proceedings of the IEEE Conference on Computer Vision and Pattern Recognition (pp. 4561-4570), 2019.

\bibitem{b171}
C. Wang, D. Xu, Y. Zhu, R. Martín-Martín, C. Lu, L. Fei-Fei, and S. Savarese,
\emph{DenseFusion: $6$D object pose estimation by iterative dense fusion},
In Proceedings of the IEEE Conference on Computer Vision and Pattern Recognition (pp. 3343-3352), 2019.

\bibitem{b172}
C. Li, J. Bai, and G.D. Hager,
\emph{A unified framework for multi-view multi-class object pose estimation},
In Proceedings of the European Conference on Computer Vision (ECCV) (pp. 254-269), 2018.

\bibitem{b173}
M. Oberweger, M. Rad, and V. Lepetit,
\emph{Making deep heatmaps robust to partial occlusions for $3$D object pose estimation},
In Proceedings of the European Conference on Computer Vision (ECCV) (pp. 119-134), 2018.

\bibitem{b174}
M. Sundermeyer, Z.C. Marton, M. Durner, M. Brucker, and R. Triebel,
\emph{Implicit $3$D orientation learning for $6$D object detection from RGB images},
In Proceedings of the European Conference on Computer Vision (ECCV) (pp. 699-715), 2018.

\bibitem{b175}
R. Brégier, F. Devernay, L. Leyrit, and J.L. Crowley,
\emph{Symmetry aware evaluation of $3$D object detection and pose estimation in scenes of many parts in bulk},
In Proceedings of the IEEE International Conference on Computer Vision (pp. 2209-2218), 2017.

\bibitem{b176}
S. Zakharov, I. Shugurov, and S. Ilic,
\emph{DPOD: $6$D Pose Object Detector and Refiner},
ICCV19.

\bibitem{b177}
Z. Li, G. Wang, and X. Ji,
\emph{CDPN: Coordinates-Based Disentangled Pose Network for Real-Time RGB-Based 6-DoF Object Pose Estimation},
ICCV19.

\bibitem{b178}
K. Park, T. Patten, and M. Vincze,
\emph{Pix2Pose: Pixel-Wise Coordinate Regression of Objects for $6$D Pose Estimation},
ICCV19.

\bibitem{b179}
A. Zeng, K.T. Yu, S. Song, D. Suo, E. Walker, A. Rodriguez, and J. Xiao,
\emph{Multi-view self-supervised deep learning for $6$D pose estimation in the Amazon Picking Challenge},
In IEEE International Conference on Robotics and Automation (ICRA) (pp. 1386-1383). IEEE, 2017.

\bibitem{b180}
M. Sundermeyer, Z.C. Marton, M. Durner, and R. Triebel,
\emph{Augmented Autoencoders: Implicit $3$D Orientation Learning for $6$D Object Detection},
International Journal of Computer Vision, pp.1-16, 2019.

\bibitem{b181}
J. Vidal, C.Y. Lin, X. Lladó, and R. Martí,
\emph{A Method for $6$D Pose Estimation of Free-Form Rigid Objects Using Point Pair Features on Range Data},
Sensors, 18(8), 2018.

\bibitem{b182}
P. Castro, A. Armagan, and T-K Kim,
\emph{Accurate $6$D Object Pose Estimation by Pose Conditioned Mesh Reconstruction},
arXiv preprint arXiv:1910.10653 (2019).

\bibitem{b183}
J. Sock, G. Garcia-Hernando, and T-K Kim,
\emph{Active $6$D Multi-Object Pose Estimation in Cluttered Scenarios with Deep Reinforcement Learning},
arXiv preprint arXiv:1910.08811, 2019.
 
\bibitem{b184}
A. Kendall, M. Grimes, and R. Cipolla,
\emph{PoseNet: A convolutional network for real-time 6-dof camera relocalization},
In The IEEE International Conference on Computer Vision (ICCV), December 2015.

\bibitem{b185}
O. Erkent, D. Shukla, and J. Piater.
\emph{Integration of probabilistic pose estimates from multiple views},
In ECCV, 2016.

\bibitem{b186}
N. Payet and S. Todorovic,
\emph{From contours to $3$D object detection and pose estimation},
In International Conference on Computer Vision (pp. 983-990). IEEE, 2011.

\bibitem{b187}
H. Tjaden, U. Schwanecke, and E. Schomer,
\emph{Real-time monocular pose estimation of $3$D objects using temporally consistent local color histograms},
In Proceedings of the IEEE International Conference on Computer Vision (pp. 124-132), 2017.

\bibitem{b188}
Y. Konishi, Y. Hanzawa, M. Kawade, and M. Hashimoto,
\emph{Fast $6$D pose estimation from a monocular image using hierarchical pose trees},
In European Conference on Computer Vision (pp. 398-413). Springer, Cham, 2016.

\bibitem{b189}
M. Ulrich, C. Wiedemann, and C. Steger,
\emph{Combining scale-space and similarity-based aspect graphs for fast $3$D object recognition},
IEEE transactions on pattern analysis and machine intelligence, 34(10), pp.1902-1914, 2011.

\bibitem{b190}
K. Park, T. Patten, J. Prankl, and M. Vincze,
\emph{Multi-Task Template Matching for Object Detection, Segmentation and Pose Estimation Using Depth Images},
In International Conference on Robotics and Automation (ICRA) (pp. 7207-7213). IEEE, 2019.

\bibitem{b191}
S. Zakharov, W. Kehl, B. Planche, A. Hutter, and S. Ilic,
\emph{$3$D object instance recognition and pose estimation using triplet loss with dynamic margin},
In IEEE/RSJ International Conference on Intelligent Robots and Systems (IROS) (pp. 552-559). IEEE, 2017.

\bibitem{b192}
E. Muñoz, Y. Konishi, V. Murino, and A. Del Bue,
\emph{Fast $6$D pose estimation for texture-less objects from a single RGB image},
In 2016 IEEE International Conference on Robotics and Automation (ICRA) (pp. 5623-5630). IEEE, 2016.

\bibitem{b193}
M. Li and K. Hashimoto,
\emph{Accurate object pose estimation using depth only},
Sensors, 18(4), p.1045, 2018.

\bibitem{b194}
C. Rennie, R. Shome, K.E. Bekris, and A.F. De Souza,
\emph{A dataset for improved RGBD-based object detection and pose estimation for warehouse pick-and-place},
Robotics and Automation Letters (2016).

\bibitem{b195}
A. Janoch, S. Karayev, Y. Jia, J.T. Barron, M. Fritz, K. Saenko, and T. Darrell,
\emph{A Category-Level $3$-D Object Dataset: Putting the Kinect to Work},
ICCV Workshop on Consumer Depth Cameras in Computer Vision 2011.

\bibitem{b196}
J. Xiao, A. Owens, and A. Torralba,
\emph{SUN$3$D: A Database of Big Spaces Reconstructed using SfM and Object Labels},
Proceedings of 14th IEEE International Conference on Computer Vision (ICCV2013).

\bibitem{b197}
C.R. Qi, O. Litany, K. He, and L.J. Guibas,
\emph{Deep Hough Voting for $3$D Object Detection in Point Clouds},
ICCV, 2019.

\bibitem{b198}
K. Lai, L. Bo, X. Ren, and D. Fox,
\emph{A large-scale hierarchical multi-view RGB-D object dataset},
In ICRA, 2011.

\bibitem{b199}
J. Lim, H. Pirsiavash, and A. Torralba,
\emph{Parsing IKEA objects: Fine pose estimation},
In ICCV, 2013.

\bibitem{b200}
K. Matzen and N. Snavely,
\emph{NYC$3$DCars: A dataset of $3$D vehicles in geographic context},
In ICCV, 2013.

\bibitem{b201}
M. Ozuysal, V. Lepetit, and P. Fua,
\emph{Pose estimation for category specific multiview object localization},
In CVPR, 2009.

\bibitem{b202}
M. Sun, G. Bradski, B.-X. Xu, and S. Savarese,
\emph{Depth-encoded hough voting for joint object detection and shape recovery},
In ECCV, 2010.

\bibitem{b203}
Y. Xiang and S. Savarese,
\emph{Estimating the aspect layout of object categories},
In CVPR, 2012.

\bibitem{b204}
Y. Xiang, R. Mottaghi, and S. Savarese,
\emph{Beyond PASCAL: A benchmark for $3$D object detection in the wild},
In IEEE Winter Conference on Applications of Computer Vision (pp. 75-82). IEEE, 2014.

\bibitem{b205}
B. Calli, A. Singh, J. Bruce, A. Walsman, K. Konolige, S. Srinivasa, P. Abbeel, and A.M. Dollar,
\emph{Yale-CMU-Berkeley dataset for robotic manipulation research},
In International Journal of Robotics Research, 2017.

\bibitem{b206}
B. Calli, A. Singh, A. Walsman, S. Srinivasa, P. Abbeel, and A. Dollar,
\emph{The YCB Object and Model Set: Towards Common Benchmarks for Manipulation Research},
In International Conference on Advanced Robotics, 2015.

\bibitem{b207}
C. Li, J. Bohren, E. Carlson, and G.D. Hager,
\emph{Hierarchical Semantic Parsing for Object Pose Estimation in Densely Cluttered Scenes},
In International Conference on Robotics Automation (ICRA), 2016.

\bibitem{b208}
V. Hegde and R. Zadeh,
\emph{Fusionnet: $3$D object classification using multiple data representations},
3D Deep Learning Workshop at NIPS 2016.

\bibitem{b209}
L. Liu, W. Ouyang, X. Wang, P. Fieguth, J. Chen, X. Liu, and M. Pietikäinen,
\emph{Deep learning for generic object detection: A survey},
Int. J. Comput. Vis. 1809, 2018, 1–58.

\bibitem{b210}
D. Holz, M. Nieuwenhuisen, D. Droeschel, J. Stuckler, A. Berner, J. Li, R. Klein, and S. Behnke,
\emph{Active Recognition and Manipulation for Mobile Robot Bin Picking},
book chapter in Gearing Up and Accelerating Cross-fertilization between Academic and Industrial Robotics Research in Europe, Springer, 2014.

\bibitem{b211}
M. Mohamad, D. Rappaport, and M. Greenspan,
\emph{Generalized $4$-Points Congruent Sets for $3$D Registration},
IEEE International Conference on 3D Vision, 2014.

\bibitem{b212}
M. Nieuwenhuisen, D. Droeschel, D. Holz, J. Stuckler, A. Berner, J. Li, R. Klein, and S. Behnke,
\emph{Mobile Bin Picking with an Anthropomorphic Service Robot},
International Conference on Robotics and Automation, 2013.

\bibitem{b213}
H. Deng, T. Birdal, and S. Ilic,
\emph{PPFNet: Global context aware local features for robust $3$D point matching},
In Proceedings of the IEEE Conference on Computer Vision and Pattern Recognition (pp. 195-205), 2018.

\bibitem{b214}
H. Deng, T. Birdal, and S. Ilic,
\emph{PPF-FoldNet: Unsupervised learning of rotation invariant $3$D local descriptors},
In Proceedings of the European Conference on Computer Vision (ECCV) (pp. 602-618), 2018.

\bibitem{b215}
T. Hodan, R. Kouskouridas, T-K Kim, F. Tombari, K. Bekris, B. Drost, T. Groueix, K. Walas, V. Lepetit, A. Leonardis, C. Steger, F. Michel, C. Sahin, C. Rother, and J. Matas,
\emph{A Summary of the $4$th International Workshop on Recovering $6$D Object Pose},
in Computer Vision – ECCV 2018 Workshops, pp. 589–600, Springer, Cham, Sept. 2018.

\bibitem{b216}
C. Sahin, G. Garcia-Hernando, J. Sock, and T-K Kim,
\emph{Instance- and Category-level $6$D Object Pose Estimation},
book chapter in RGB-D Image Analysis and Processing (pp. 243-265). Springer, 2019.


\end{thebibliography}
\end{document}